\documentclass[preprint,12pt,authoryear]{elsarticle}

\newcommand{\gain}[1]{%
    \mbox{\hspace{1pt}\raisebox{-0.5ex}{%
        \scriptsize\textbf{(+#1)}%
    }}%
}

\usepackage{float}
\usepackage[utf8]{inputenc}
\usepackage[T1]{fontenc}
\usepackage{xcolor}
\usepackage{caption}
\usepackage[justification=centering]{caption}
\usepackage{tabularx}
\usepackage{array}
\usepackage{booktabs} 
\usepackage{lscape}
\usepackage{orcidlink}

\usepackage{longtable}   
\usepackage{adjustbox}
\usepackage{multirow}
\usepackage{subcaption}
\usepackage{pdflscape}
\usepackage{amssymb}
\usepackage{amsmath}
\usepackage{graphicx}
\usepackage{hyperref}

\usepackage{makecell}

\newcolumntype{Y}{>{\centering\arraybackslash}X}

\journal{Nuclear Physics B}

\begin{document}

\begin{frontmatter}

\title{HMMF-Net: A Lightweight Hierarchical Multi-Frame Multimodal Fusion Network for Accurate Fish Feeding Intensity Assessment}

\author{Phi-Hung Hoang}
\ead{hunghpde180523@fpt.edu.vn}

\author{Van-Manh Tran}
\ead{manhtvde180090@fpt.edu.vn}

\author{Nam-Thuan Trinh}
\ead{thuantnde180305@fpt.edu.vn}

\author{Thi-Thu-Hong Phan\corref{cor}\orcidlink{0000-0001-6880-3721}}
\ead{hongptt11@fe.edu.vn}

\cortext[cor]{Corresponding author}

\hypersetup{pdfauthor={Thi-Thu-Hong Phan}}

\affiliation{organization={AIT laboratory, Faculty of Artificial Intelligence, \\FPT University},
            addressline={Da Nang},
            postcode={550000}, 
            country={Vietnam}}

\begin{abstract}

Automated fish feeding intensity assessment (FFIA) is important for precision aquaculture and feed cost optimization. However, unimodal methods are vulnerable to acoustic noise and visual disturbances, while existing multimodal frameworks often rely on computationally intensive 3D convolutions or Transformers, limiting edge deployment.
To address these limitations, we propose HMMF-Net, a lightweight multimodal architecture that combines early multi-frame visual encoding and multi-axis acoustic modeling through a hierarchical bidirectional messenger cross-modal fusion (BMCF) mechanism. The visual branch captures temporal variations in feeding behavior by stacking temporally distant frames as a multi-channel input to an adapted 2D EfficientNetB0 backbone, while the acoustic branch employs adaptive multi-axis residual units (AMARU) to extract temporal, frequency, and local time--frequency representations from log mel-spectrograms. BMCF progressively exchanges information between the two modalities across intermediate representation levels through compact messenger tokens, reducing the computational overhead associated with dense cross-modal attention.
Evaluated on the AV-FFIA dataset using five-fold cross-validation, HMMF-Net achieves $96.77 \pm 0.27\%$ accuracy and $96.88 \pm 0.26\%$ F1-score on the held-out test folds, with only 3.987M parameters and 1.002 GFLOPs. Compared with the best-performing visual configuration (95.49\% accuracy), HMMF-Net reduces the classification error by 28.38\%. On-device deployment on an NVIDIA Jetson Orin Nano using TensorRT with FP32 precision maintains comparable accuracy with an inference time of 11.54 ms per 2-second sample and a model size of 18.81~MB. These results demonstrate that HMMF-Net effectively combines multimodal representation and computational efficiency, supporting accurate and practical FFIA on resource-constrained edge platforms.

\end{abstract}

\begin{keyword}
Fish feeding intensity assessment; Multimodal fusion; Multi-frame learning; Cross-modal interaction; Edge deployment
\end{keyword}

\end{frontmatter}

\section{Introduction}

Aquaculture plays an increasingly important role in meeting the growing global demand for aquatic foods. However, feed management remains one of the major cost factors in aquaculture, accounting for approximately 30\%--70\% of total production costs~\citep{Verdal2018}. Traditional feeding practices based on farmer experience, manual observation, or predetermined schedules may result in overfeeding or underfeeding, leading to feed waste, reduced water quality, and suboptimal fish growth~\citep{Zhou2018,Li2020}. Therefore, automated methods based on visual and acoustic information have been increasingly investigated for fish feeding intensity assessment (FFIA) to support more accurate feeding decisions and real-time feeding management in aquaculture~\citep{Xiao2025}.

Research on automated fish feeding intensity assessment has primarily explored unimodal approaches, including acoustic-based and visual-based methods. Acoustic approaches have been investigated under turbid, aerated, or poorly illuminated water conditions by capturing feeding-related sounds, such as food capture, chewing, and water-surface agitation. Existing studies have transformed acoustic signals into time-frequency representations, such as mel-spectrograms and MFCCs, and applied CNNs, lightweight architectures, and attention mechanisms to characterize feeding intensity~\citep{Cui2022,Du2023,Huang2025,Li2026}. Nevertheless, acoustic-only methods remain sensitive to background noise from water pumps, aerators, and other mechanical equipment, which may mask feeding-related sounds and reduce assessment reliability. In parallel, visual-based approaches assess feeding intensity through observable fish activities and water-surface disturbances. Early visual studies mainly applied 2D CNNs to static images with attention mechanisms or preprocessing techniques to reduce environmental interference~\citep{Zhou2019,Zhang2022,Wu2024,Zhang2025c}. However, static images provide limited temporal information. Subsequent studies therefore employed video sequences with 3D CNNs, optical flow, structured state space models, and Vision Transformers (ViTs) to capture motion and temporal dynamics~\citep{Dong2024,Liu2024,Wang2026a,Wang2026b}. These approaches can introduce considerable computational and memory requirements, while visual systems remain sensitive to water turbidity, illumination variations, and fish occlusion.

The complementary characteristics of acoustic and visual signals have motivated multimodal approaches that integrate information from both modalities~\citep{Cui2025}. Recent studies have explored various fusion mechanisms, including low-rank bilinear pooling, audio-guided cross-attention, dynamic interactive modules, and hybrid Transformer architectures~\citep{Li2025,Wang2025,Sha2026,Yang2024}. Other approaches have further incorporated environmental information, such as dissolved oxygen and water-wave signals, through progressive interaction mechanisms or cross-modal knowledge distillation~\citep{Gu2025,Zhang2026a,Zhang2025a}. Despite these advances, three challenges remain for practical multimodal FFIA. First, acoustic and visual signals capture different aspects of feeding behavior and are influenced by distinct environmental conditions, making their effective integration essential for reliable feeding assessment. Second, a single visual frame provides a limited temporal observation of feeding behavior, whereas capturing temporal dynamics through full video sequences can introduce substantial computational and memory costs. Third, multimodal fusion requires effective information exchange between heterogeneous visual and acoustic representations, while conventional cross-modal interactions can become computationally demanding when applied to high-dimensional feature maps. These challenges raise a fundamental question for real-time FFIA: how can information from both modalities, useful temporal cues, and efficient cross-modal interaction be jointly achieved under the computational constraints of edge platforms~\citep{Cui2025,Li2025,Du2026}?

To address these challenges, this study proposes HMMF-Net, a lightweight multimodal framework for fish feeding intensity assessment under resource-constrained aquaculture conditions. The framework is guided by a simple design principle: preserve informative temporal and cross-modal cues while minimizing computationally intensive processing. Specifically, the visual stream uses sparse temporal observations to capture changes in feeding behavior without requiring full-sequence modeling, while the acoustic stream provides additional information from feeding-related sounds. These modality-specific representations are progressively integrated through an efficient cross-modal interaction mechanism that limits the computational burden associated with high-dimensional feature fusion. This design provides a practical balance between multimodal representation capability and computational efficiency for real-time edge deployment.

The main contributions of this study are as follows:

\begin{itemize}
\item We propose a lightweight multimodal learning framework, HMMF-Net, that employs sparse multi-frame visual encoding with an adapted 2D EfficientNetB0 backbone to capture different temporal aspects of feeding behavior while avoiding the computational overhead of processing the full visual sequence.

\item We introduce an adaptive multi-axis residual unit (AMARU) for the acoustic branch, which jointly models temporal, frequency, and local time--frequency dependencies in Mel-spectrogram representations to extract discriminative features for distinguishing different levels of fish feeding intensity.

\item We design a hierarchical bidirectional messenger cross-modal fusion (BMCF) module that progressively exchanges information between acoustic and visual representations through a compact, low-dimensional messenger space, enabling input-dependent cross-modal interaction while avoiding dense fusion over high-dimensional feature maps.

\item We comprehensively evaluate HMMF-Net on the AV-FFIA dataset and further validate its practical feasibility by deploying it on an NVIDIA Jetson Orin Nano using TensorRT with FP32 precision.
\end{itemize}

The rest of the paper is organized as follows. Section~\ref{sec:related_works} reviews related work on fish feeding intensity assessment, while Section~\ref{sec:methodology} presents the proposed methodology and framework architecture. Section~\ref{sec:experiments} describes the experimental setup and implementation details, and Section~\ref{sec:results_discussion} reports the experimental results, ablation studies, and on-device deployment analysis. Finally, Section~\ref{sec:conclusion} concludes the paper and discusses future research directions.

\section{Related works}\label{sec:related_works}

Research on automated FFIA has explored different sensing modalities and representation learning strategies, with recent studies increasingly moving from unimodal analysis toward multimodal learning. Existing approaches can be broadly categorized into acoustic-based, visual-based, and multimodal methods. These approaches differ substantially in how they represent temporal feeding dynamics, extract modality-specific information, and exchange information across modalities. The design choices also lead to different computational requirements, which are particularly relevant for real-time deployment in aquaculture environments.

\subsection{Acoustic-based approaches}

Acoustic-based FFIA methods typically transform underwater recordings into time-frequency representations, such as mel-spectrograms or MFCCs, and then employ convolutional or attention-based networks to characterize feeding-related acoustic patterns.~\cite{Cui2022} established the AFFIA3K dataset and demonstrated the feasibility of CNN-based feeding intensity classification from mel-spectrograms. Subsequent studies have explored lightweight architectures and feature enhancement mechanisms.~\cite{Du2023} introduced MobileNetV3-SBSC to reduce the computational cost of mel-spectrogram classification, while~\cite{Huang2025} enhanced VGG16 with spatial and channel-wise attention to improve the extraction of discriminative acoustic features. More recently,~\cite{Li2026} proposed FTPG-CNN6, which combines MFCC representations with frequency-time attention to improve robustness to mechanical interference and facilitate deployment on embedded devices.

These studies demonstrate the effectiveness of acoustic information for FFIA, but acoustic representations remain susceptible to mechanical interference from aerators, water pumps, and other equipment. For edge-oriented multimodal systems, an acoustic encoder therefore needs to balance discriminative capability and computational efficiency.

\subsection{Visual-based approaches}

Visual-based FFIA has evolved from static-image classification toward increasingly explicit temporal modeling. Early approaches mainly relied on 2D Convolutional Neural Networks (CNNs) to characterize fish activity and water-surface changes from individual images.~\cite{Zhou2019} employed LeNet-5 with data augmentation, while~\cite{Zhang2022} introduced MobileNetV2-SENet to emphasize informative visual features. To address interference from reflections, ripples, and background regions,~\cite{Wu2024} used class activation maps to generate pseudo-labels for feeding splashes.~\cite{Zhang2025c} further incorporated water-surface glare removal with EfficientNetB0-FPSANet to focus on feeding-related regions.

A major development in visual FFIA has been the incorporation of temporal information.~\cite{Feng2022} proposed a lightweight 3D ResNet-GloRe for video-based representation learning, while~\cite{Dong2024} combined binarized images, optical flow, and statistical texture features in a three-stream architecture.~\cite{Wang2026a} adopted a dual-stream RGB and optical-flow representation with structured state space modeling to capture spatiotemporal feeding patterns. Transformer-based approaches have also been investigated.~\cite{Liu2024} proposed CFFI-ViT by incorporating residual modules into a modified Vision Transformer, whereas~\cite{Wang2026b} introduced FeedingMonitor based on a Transformer-in-Transformer architecture with blur interpolation and dynamic token sparsification.

The progression from static images to video and sequence-based representations highlights the importance of temporal information for distinguishing feeding intensity. However, 3D convolutions, optical-flow extraction, multiple streams, and Transformer-based modeling can increase computational and memory requirements, creating a trade-off between temporal representation capability and computational efficiency for edge-oriented FFIA systems.

\subsection{Multimodal approaches}

The complementary characteristics of acoustic and visual information have motivated multimodal FFIA approaches that integrate heterogeneous sensing modalities within a unified framework.~\cite{Cui2025} established the AV-FFIA benchmark and proposed U-FFIA, a unified mixed-modality architecture incorporating modality dropout, spectral pooling (SimPFs), and knowledge distillation to support audio, visual, and audio-visual inputs with reduced computational requirements. Subsequent studies have explored different mechanisms for cross-modal interaction.~\cite{Li2025} proposed ACAF-Net, which combines bilateral filtering of mel-spectrograms with Low-Rank Bilinear Pooling and environment-aware adaptive attention.~\cite{Wang2025} introduced a hierarchical gated cross-modal network using audio-guided attention, a hybrid BiLSTM-Transformer architecture, and learnable gating.~\cite{Sha2026} developed a dynamic audio-visual fusion framework to adaptively integrate heterogeneous representations, while~\cite{Yang2024} proposed a U-shaped bilinear fusion architecture with a TAP module for audio-visual fish behavior recognition.

Recent studies have further extended multimodal FFIA by incorporating environmental signals.~\cite{Gu2025} developed MMFINet to jointly exploit video, audio, and dissolved oxygen information using lightweight separable convolutions and cross-modal attention.~\cite{Zhang2026a} proposed a progressive multimodal interaction network integrating image, audio, and water-wave information through auxiliary-modality reinforcement and adaptive evidential reasoning. Focusing on model efficiency,~\cite{Zhang2025a} introduced a multimodal knowledge distillation framework that combines video and dissolved oxygen information with semi-supervised domain adaptation, reducing the student model to 2.49M parameters. More recently,~\cite{Du2026} proposed an uncertainty-aware heterogeneous multi-level knowledge distillation framework that combines MoViNet-A0 and EfficientAT with reliability-aware gating, bidirectional residual interaction, and multi-level knowledge transfer to obtain a compact audio-visual model.

Despite these advances, multimodal FFIA methods involve different trade-offs between fusion capability and computational complexity. Simple feature concatenation is computationally efficient but provides limited explicit interaction, whereas bilinear pooling, cross-attention, and multi-level interaction can capture richer cross-modal dependencies at a higher computational cost. These trade-offs are particularly relevant for FFIA systems targeting resource-constrained edge platforms.

\subsection{Research gap and positioning of HMMF-Net}

The reviewed studies highlight four challenges for practical FFIA. First, capturing temporal variations in feeding behavior remains challenging without incurring the computational cost of explicit video or sequence modeling. Second, acoustic modeling requires complementary temporal, frequency, and joint time--frequency representations while maintaining a lightweight architecture. Third, multimodal architectures require efficient and input-dependent interaction between heterogeneous representations without dense fusion over high-dimensional feature maps. Finally, the practical feasibility of these designs on resource-constrained edge devices remains an important consideration.

HMMF-Net addresses these challenges through a unified design that combines sparse multi-frame visual encoding, lightweight multi-axis acoustic modeling, and efficient hierarchical cross-modal interaction. This positioning enables the framework to jointly consider temporal representation, modality complementarity, and computational efficiency within a single FFIA architecture. Its practical feasibility is further validated through deployment on an NVIDIA Jetson Orin Nano.

\section{Proposed framework}\label{sec:methodology}

\subsection{Overview of the proposed framework}

\begin{figure}[!t]
    \centering
    \includegraphics[width=\linewidth]{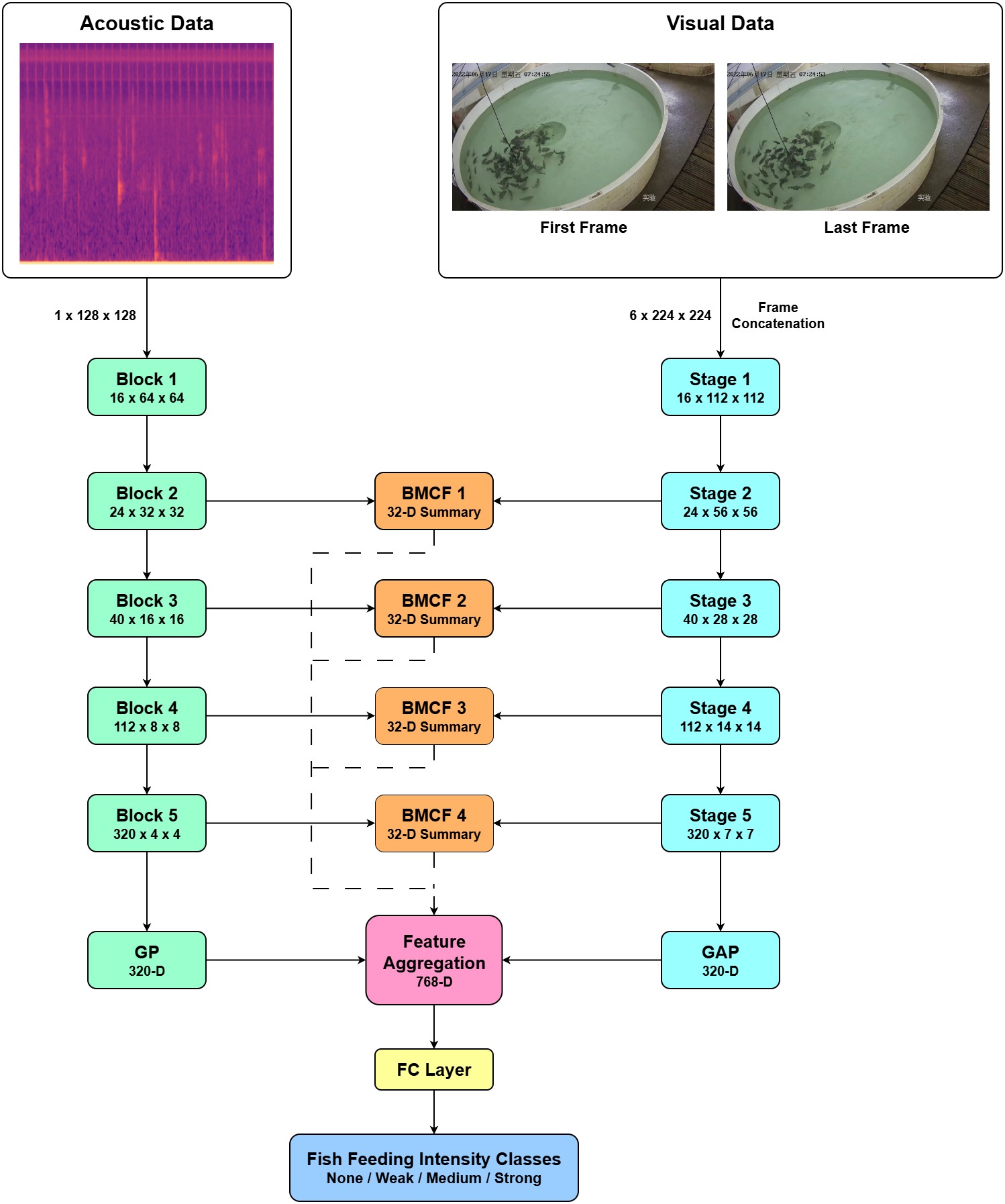}
    \caption{Overall architecture of the proposed multimodal HMMF-Net framework.}
    \label{fig:pipeline}
\end{figure}

In this study, we propose a lightweight multimodal framework for accurate fish feeding intensity assessment (FFIA) that integrates complementary visual and acoustic information while maintaining low computational cost for edge deployment. The overall architecture is illustrated in Figure~\ref{fig:pipeline}.

The framework comprises four main components that enable efficient multimodal representation learning and hierarchical cross-modal interaction:

\vspace{0.1cm}

\textbf{1. Early multi-frame visual encoding and visual branch}

\vspace{0.1cm}

To capture complementary visual cues from different stages of the feeding process without the computational overhead of 3D convolutions, the first and last frames of each video clip are jointly encoded at the input level. Their RGB channels are combined to form a six-channel representation, which is subsequently processed by an adapted EfficientNetB0 to extract hierarchical visual features. This early multi-frame design preserves the efficiency of conventional 2D convolutional processing while incorporating information from two temporally separated observations.

\vspace{0.1cm}

\textbf{2. AMARU-Net acoustic branch}

\vspace{0.1cm}

The acoustic branch processes Log mel-spectrograms from synchronized underwater recordings using the proposed AMARU-Net. Its adaptive multi-axis residual units (AMARUs) extract complementary acoustic patterns along the temporal, frequency, and joint time--frequency dimensions. The hierarchical structure provides intermediate feature representations aligned with the visual branch for subsequent multimodal interaction.

\vspace{0.1cm}

\textbf{3. Hierarchical bidirectional messenger cross-modal fusion (BMCF)}

\vspace{0.1cm}

BMCF facilitates progressive information exchange between the visual and acoustic branches. At multiple intermediate stages, each BMCF module forms compact messenger representations that exchange complementary information in both directions, avoiding dense cross-attention over the full feature maps. The resulting cross-modal context is then reintegrated into the modality-specific features through gated residual connections. By applying BMCF across multiple representation levels, the framework enables hierarchical multimodal interaction while preserving the distinctive characteristics of each branch.

\vspace{0.1cm}

\textbf{4. Multimodal feature aggregation and prediction}

\vspace{0.1cm}

At the final stage, the modality-specific embeddings and cross-modal features from the hierarchical fusion stages are aggregated into a unified multimodal representation. This representation combines modality-specific and complementary cross-modal information and is then fed to the prediction layer to classify the feeding intensity into four levels: \textit{None}, \textit{Weak}, \textit{Medium}, and \textit{Strong}.

The proposed framework is described in detail in the following subsections.

\subsection{Architecture of HMMF-Net}

HMMF-Net consists of parallel visual and acoustic branches interconnected by hierarchical bidirectional messenger cross-modal fusion (BMCF) modules. The visual branch processes the early multi-frame representation, while the acoustic branch operates on mel-spectrograms. Cross-modal interaction is progressively performed at multiple intermediate levels through the BMCF modules, after which the refined modality-specific features are aggregated for feeding intensity prediction. The detailed design of each component is presented in the following subsections.

\subsubsection{Early multi-frame visual encoding}

To incorporate information from temporally separated observations beyond conventional single-frame representations while avoiding the high computational and memory requirements of multi-frame 3D convolutional models, an early multi-frame visual encoding strategy is adopted. Specifically, the first and last frames of each video clip are selected to capture visual information at two temporally distant moments. These two RGB frames are then concatenated along the channel dimension to form a unified six-channel representation that can be processed by a standard 2D convolutional network.

Given a video clip consisting of $T$ frames, the visual sequence is denoted as
\begin{equation}
    \mathcal{V} = \{I_0, I_1, \ldots, I_{T-1}\},
\end{equation}
where $I_t \in \mathbb{R}^{3 \times H \times W}$ represents the RGB frame at time step $t$. The first frame $I_0$ and the last frame $I_{T-1}$ are concatenated along the channel dimension:
\begin{equation}
    \mathbf{X}_v = \operatorname{Concat}_{c}(I_0, I_{T-1}),
\end{equation}
yielding
\begin{equation}
    \mathbf{X}_v \in \mathbb{R}^{6 \times H \times W}.
\end{equation}

This formulation integrates visual observations from two temporally distant positions into a single input tensor, providing complementary information without explicitly processing the entire frame sequence. Because conventional 2D convolutional networks are designed for three-channel RGB inputs, the first convolutional layer of the backbone is adapted by expanding its input channels from 3 to 6, while the remaining network architecture is kept unchanged. The resulting six-channel representation is then fed into the visual branch for hierarchical feature extraction.

\subsubsection{Visual branch}

The visual branch adopts EfficientNetB0~\citep{Tan2020} as the backbone due to its favorable trade-off between representation capacity and computational efficiency. EfficientNetB0 is built upon mobile inverted bottleneck convolution (MBConv) blocks that combine pointwise expansion, depthwise convolution, squeeze-and-excitation, and residual connections where applicable. The first convolutional layer is adapted to accept the six-channel multi-frame input, while the remaining architecture is retained.

\begin{table}[H]
\centering
\caption{Architecture of the EfficientNetB0 visual branch.}
\label{tab:visual_branch}
\renewcommand{\arraystretch}{1.2}
\setlength{\tabcolsep}{6pt}

\scalebox{0.758}{
\begin{tabular}{|l|c|c|}
\hline
\textbf{Stage} & \textbf{Operator} & \textbf{Output size} \\
\hline

Stem
& $3 \times 3$ convolution, $6 \rightarrow 32$, stride 2
& $32 \times 112 \times 112$ \\
\hline

Stage 1
& MBConv1, $3 \times 3$, $\times 1$
& $16 \times 112 \times 112$ \\
\hline

Stage 2
& MBConv6, $3 \times 3$, $\times 2$
& $24 \times 56 \times 56$ \\
\hline

Stage 3
& MBConv6, $5 \times 5$, $\times 2$
& $40 \times 28 \times 28$ \\
\hline

\multirow{2}{*}{Stage 4}
& MBConv6, $3 \times 3$, $\times 3$
& $80 \times 14 \times 14$ \\
\cline{2-3}
& MBConv6, $5 \times 5$, $\times 3$
& $112 \times 14 \times 14$ \\
\hline

\multirow{2}{*}{Stage 5}
& MBConv6, $5 \times 5$, $\times 4$
& $192 \times 7 \times 7$ \\
\cline{2-3}
& MBConv6, $3 \times 3$, $\times 1$
& $320 \times 7 \times 7$ \\
\hline

Global pooling
& Global average pooling
& $320$-D \\
\hline

\end{tabular}
}

\vspace{1mm}

\begin{minipage}{\textwidth}
\scriptsize
\textit{Note:} MBConv1 and MBConv6 denote MBConv blocks with expansion ratios of 1 and 6, respectively. For the proposed fusion hierarchy, consecutive MBConv groups operating at the same spatial resolution are grouped into a single stage. Feature-map dimensions are reported as channels $\times$ height $\times$ width.
\end{minipage}

\end{table}

As summarized in Table~\ref{tab:visual_branch}, the visual encoder follows a five-stage hierarchy in which the spatial resolution is progressively reduced while the channel dimension increases. The outputs of Stages 2--5, with 24, 40, 112, and 320 channels, respectively, serve as the four hierarchical representation levels for cross-modal interaction. The corresponding acoustic blocks are designed with matching channel dimensions to facilitate fusion via the BMCF modules. Differences in spatial resolution between the two modalities are handled within the BMCF modules.

After the final cross-modal interaction, global average pooling is applied to the $320 \times 7 \times 7$ feature map to obtain the visual embedding:
\begin{equation}
\mathbf{z}_v = \operatorname{GAP}(\mathbf{V}^{(5)}),
\qquad
\mathbf{z}_v \in \mathbb{R}^{320},
\end{equation}
where $\mathbf{V}^{(5)}$ denotes the Stage-5 visual feature map after BMCF. The resulting vector $\mathbf{z}_v$ is subsequently used for multimodal aggregation and prediction.

\subsubsection{Acoustic branch (AMARU-Net)}
\label{sec:amaru_net}

The acoustic branch, termed AMARU-Net, is designed to extract compact and informative spectro-temporal representations from mel-spectrograms while maintaining low computational complexity. As summarized in Table~\ref{tab:acoustic_branch}, AMARU-Net follows a five-block hierarchical architecture. Each block consists of two consecutive adaptive multi-axis residual units (AMARUs), followed by a $2 \times 2$ average pooling operation with a stride of 2. As the network becomes deeper, the temporal and frequency resolutions are progressively reduced, while the number of feature channels increases. This hierarchical design enables the network to gradually transform the input spectrogram into increasingly compact acoustic representations.

The output channel dimensions of Blocks 2--5 are set to 24, 40, 112, and 320, respectively, matching the channel dimensions of Stages 2--5 in the visual branch. This correspondence provides channel-compatible hierarchical representations for cross-modal interaction through the BMCF modules. The complete architecture of AMARU-Net is summarized in Table~\ref{tab:acoustic_branch}.

\paragraph{Adaptive multi-axis residual unit}

\begin{figure}[H]
    \centering
    \includegraphics[width=\linewidth]{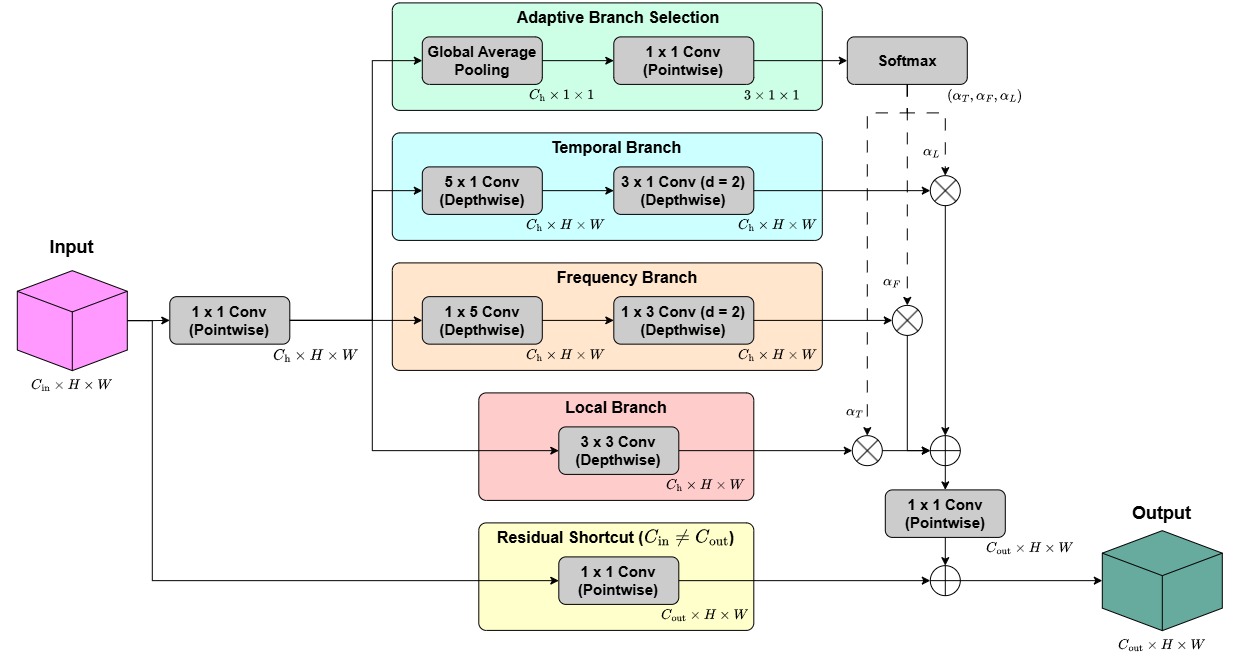}
    \caption{Architecture of the proposed adaptive multi-axis residual unit (AMARU).}
    \label{fig:amaru}
\end{figure}

The AMARU serves as the basic feature extraction unit of the acoustic branch and is illustrated in Figure~\ref{fig:amaru}. Given an input feature map
$\mathbf{X} \in \mathbb{R}^{C_{\mathrm{in}} \times H \times W}$, where $H$ and $W$ denote the temporal and frequency dimensions, respectively, a $1 \times 1$ pointwise convolution first projects the input into a hidden feature space with $C_h$ channels. The resulting feature is then processed by three parallel depthwise convolutional branches: a temporal branch using $5 \times 1$ and dilated $3 \times 1$ kernels, a frequency branch using $1 \times 5$ and dilated $1 \times 3$ kernels, and a local branch using a $3 \times 3$ kernel. The second convolution in the temporal and frequency branches uses a dilation rate of 2 to enlarge the corresponding receptive fields without increasing the number of kernel parameters.

\begin{table}[H]
\centering
\caption{Architecture of the AMARU-Net acoustic branch.}
\label{tab:acoustic_branch}
\renewcommand{\arraystretch}{1.2}
\setlength{\tabcolsep}{5pt}

\scalebox{0.75}{
\begin{tabular}{|l|c|c|c|c|}
\hline
\textbf{Block} &
\textbf{Operator} &
\textbf{Repeats} &
\textbf{Channel configuration} &
\textbf{Output size} \\
\hline

\multirow{3}{*}{Block 1}
& \multirow{2}{*}{AMARU}
& \multirow{2}{*}{2}
& $1 \rightarrow 16 \rightarrow 16$
& \multirow{2}{*}{$16 \times 128 \times 128$} \\
\cline{4-4}
&
&
& $16 \rightarrow 16 \rightarrow 16$
& \\
\cline{2-5}
& $2 \times 2$ average pooling (stride 2)
& 1
& --
& $16 \times 64 \times 64$ \\
\hline

\multirow{3}{*}{Block 2}
& \multirow{2}{*}{AMARU}
& \multirow{2}{*}{2}
& $16 \rightarrow 24 \rightarrow 24$
& \multirow{2}{*}{$24 \times 64 \times 64$} \\
\cline{4-4}
&
&
& $24 \rightarrow 24 \rightarrow 24$
& \\
\cline{2-5}
& $2 \times 2$ average pooling (stride 2)
& 1
& --
& $24 \times 32 \times 32$ \\
\hline

\multirow{3}{*}{Block 3}
& \multirow{2}{*}{AMARU}
& \multirow{2}{*}{2}
& $24 \rightarrow 40 \rightarrow 40$
& \multirow{2}{*}{$40 \times 32 \times 32$} \\
\cline{4-4}
&
&
& $40 \rightarrow 40 \rightarrow 40$
& \\
\cline{2-5}
& $2 \times 2$ average pooling (stride 2)
& 1
& --
& $40 \times 16 \times 16$ \\
\hline

\multirow{3}{*}{Block 4}
& \multirow{2}{*}{AMARU}
& \multirow{2}{*}{2}
& $40 \rightarrow 64 \rightarrow 112$
& \multirow{2}{*}{$112 \times 16 \times 16$} \\
\cline{4-4}
&
&
& $112 \rightarrow 64 \rightarrow 112$
& \\
\cline{2-5}
& $2 \times 2$ average pooling (stride 2)
& 1
& --
& $112 \times 8 \times 8$ \\
\hline

\multirow{3}{*}{Block 5}
& \multirow{2}{*}{AMARU}
& \multirow{2}{*}{2}
& $112 \rightarrow 128 \rightarrow 320$
& \multirow{2}{*}{$320 \times 8 \times 8$} \\
\cline{4-4}
&
&
& $320 \rightarrow 128 \rightarrow 320$
& \\
\cline{2-5}
& $2 \times 2$ average pooling (stride 2)
& 1
& --
& $320 \times 4 \times 4$ \\
\hline

\begin{tabular}[c]{@{}c@{}}
Global \\
pooling
\end{tabular}
&
\begin{tabular}[c]{@{}c@{}}
Frequency mean \\
$\rightarrow$ temporal max + mean pooling
\end{tabular}
&
--
&
$320 \rightarrow 320$
&
$320$-D \\
\hline

\end{tabular}
}

\vspace{1mm}

\begin{minipage}{\textwidth}
\scriptsize
\textit{Note:} AMARU denotes the adaptive multi-axis residual unit. Channel configurations are reported as input $\rightarrow$ hidden $\rightarrow$ output channels for each AMARU. Feature-map dimensions are reported as channels $\times$ height ($H$) $\times$ width ($W$), where $H$ and $W$ correspond to the temporal and frequency dimensions, respectively.
\end{minipage}

\end{table}

To adaptively combine the complementary branch representations, global average pooling followed by a $1 \times 1$ pointwise transformation and softmax normalization generates three input-dependent weights, $\alpha_T$, $\alpha_F$, and $\alpha_L$. The branch outputs are combined as

\begin{equation}
\mathbf{F}_M =
\alpha_T \mathbf{F}_T +
\alpha_F \mathbf{F}_F +
\alpha_L \mathbf{F}_L,
\qquad
\alpha_T + \alpha_F + \alpha_L = 1,
\label{eq:amaru_fusion}
\end{equation}

\noindent where $\mathbf{F}_T$, $\mathbf{F}_F$, and $\mathbf{F}_L$ denote the temporal, frequency, and local branch outputs, respectively. The fused representation is then projected to $C_{\mathrm{out}}$ channels using a $1 \times 1$ pointwise convolution and combined with a residual shortcut. An identity mapping is used when $C_{\mathrm{in}}=C_{\mathrm{out}}$; otherwise, a $1 \times 1$ convolution aligns the channel dimensions.

Multiple AMARUs are stacked hierarchically to construct AMARU-Net, enabling progressive extraction of compact acoustic representations at different levels.

\paragraph{Acoustic representation}

After the final acoustic block, the resulting feature map is denoted by
$\mathbf{A}^{(5)} \in \mathbb{R}^{320 \times 4 \times 4}$. To obtain a compact acoustic embedding, mean pooling is first applied along the frequency dimension $W$, producing a temporal representation
$\overline{\mathbf{A}} \in \mathbb{R}^{320 \times 4}$. The remaining temporal responses are then summarized using both maximum and mean pooling, followed by element-wise addition:

\begin{equation}
\mathbf{z}_a =
\operatorname{Max}_{H}(\overline{\mathbf{A}})
+
\operatorname{Mean}_{H}(\overline{\mathbf{A}}),
\qquad
\mathbf{z}_a \in \mathbb{R}^{320}.
\label{eq:acoustic_embedding}
\end{equation}

The resulting $\mathbf{z}_a$ serves as the final acoustic representation and is subsequently combined with the visual representation through the multimodal aggregation and prediction stage.

\subsubsection{Bidirectional messenger cross-modal fusion (BMCF)}

\begin{landscape}

\begin{figure}[H]
    \centering
    \includegraphics[width=\linewidth]{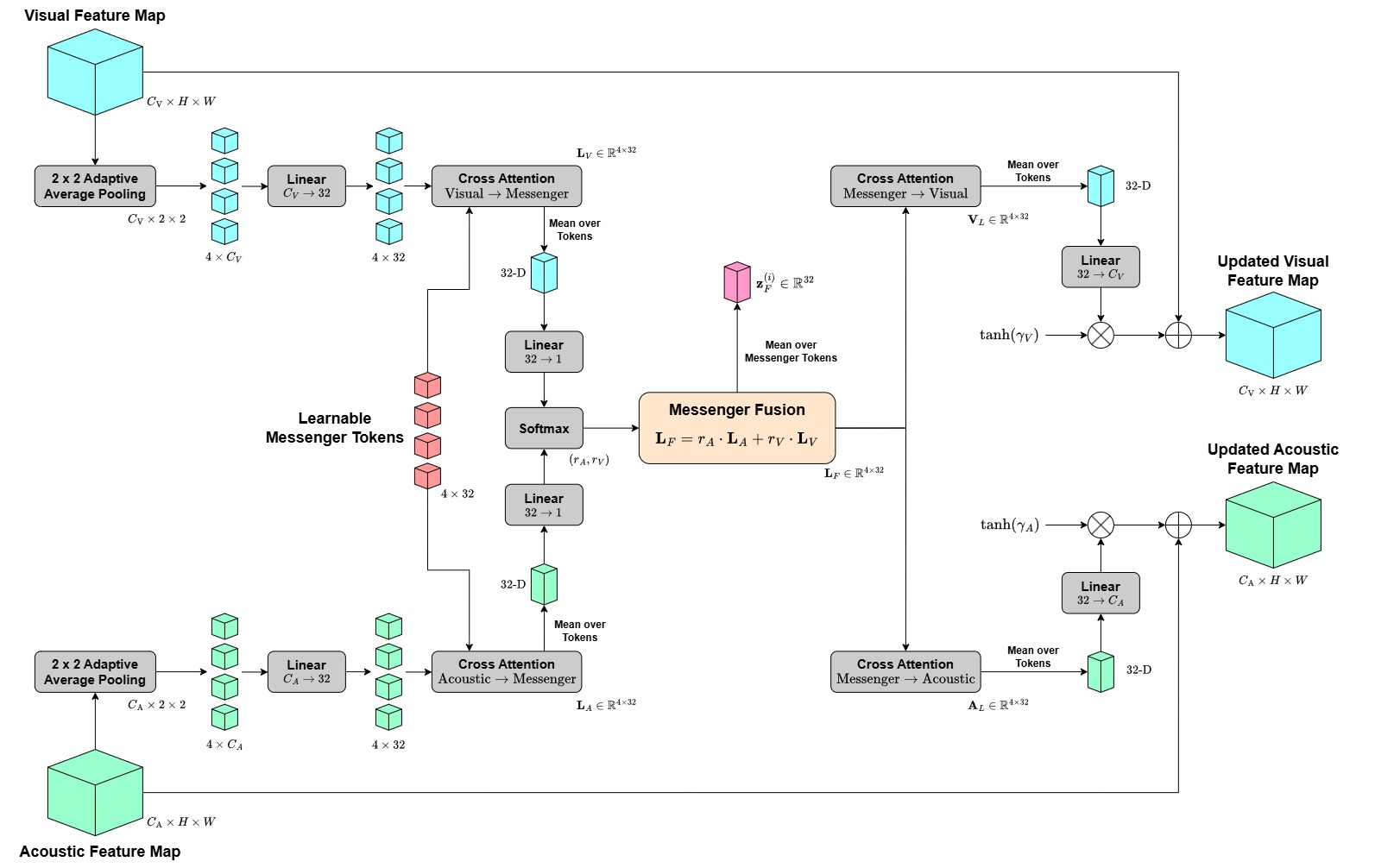}
    \caption{Architecture of the proposed bidirectional messenger cross-modal fusion (BMCF) module.}
    \label{fig:BMCF}
\end{figure}

\end{landscape}

A bidirectional messenger cross-modal fusion (BMCF) module is introduced at multiple hierarchical representation levels to facilitate efficient information exchange between the visual and acoustic branches. BMCF compresses each modality into a small set of tokens and performs cross-modal interaction within a compact shared messenger space. This design avoids dense cross-modal attention over the original feature maps while enabling bidirectional information flow between the two modality-specific representations. The overall structure of the BMCF module is illustrated in Figure~\ref{fig:BMCF}.

Given a visual feature map $\mathbf{V} \in \mathbb{R}^{C_V \times H_V \times W_V}$ and an acoustic feature map $\mathbf{A} \in \mathbb{R}^{C_A \times H_A \times W_A}$, adaptive average pooling (AAP) reduces each feature map to a spatial resolution of $2 \times 2$. The pooled representations are then reshaped into four tokens and independently projected into a shared messenger dimension of $d=32$, yielding

\begin{equation}
\mathbf{V}_t
=
\mathcal{P}_V
\left(
\operatorname{Flatten}
\left(
\operatorname{AAP}_{2\times2}(\mathbf{V})
\right)
\right),
\qquad
\mathbf{V}_t \in \mathbb{R}^{4 \times 32},
\end{equation}

\begin{equation}
\mathbf{A}_t
=
\mathcal{P}_A
\left(
\operatorname{Flatten}
\left(
\operatorname{AAP}_{2\times2}(\mathbf{A})
\right)
\right),
\qquad
\mathbf{A}_t \in \mathbb{R}^{4 \times 32},
\end{equation}

\noindent where $\mathcal{P}_V$ and $\mathcal{P}_A$ denote the corresponding linear projections. Mapping both modalities into a common messenger dimensionality enables cross-modal interaction regardless of their original spatial resolutions.

A set of four learnable messenger tokens, $\mathbf{L} \in \mathbb{R}^{4 \times 32}$, is employed as a compact shared representation through which information from the two modalities is exchanged. In the first interaction direction, the messenger tokens serve as queries, while the modality-specific tokens provide the keys and values. Four-head multi-head attention (MHA) is then applied to obtain modality-conditioned messenger representations:

\begin{equation}
\mathbf{M}_V
=
\operatorname{MHA}
\left(
\mathbf{L},
\mathbf{V}_t,
\mathbf{V}_t
\right),
\qquad
\mathbf{M}_A
=
\operatorname{MHA}
\left(
\mathbf{L},
\mathbf{A}_t,
\mathbf{A}_t
\right),
\end{equation}

The attention outputs are combined with the original learnable messenger tokens through residual connections and layer normalization (LN):

\begin{equation}
\mathbf{L}_V
=
\operatorname{LN}
\left(
\mathbf{L}+\mathbf{M}_V
\right),
\qquad
\mathbf{L}_A
=
\operatorname{LN}
\left(
\mathbf{L}+\mathbf{M}_A
\right),
\end{equation}

\noindent where
$\mathbf{L}_V,\mathbf{L}_A \in \mathbb{R}^{4 \times 32}$
represent the visual- and acoustic-conditioned messenger representations, respectively.

Since the relative contribution of the visual and acoustic modalities may vary across samples, BMCF further introduces an input-dependent modality weighting mechanism. Each modality-conditioned messenger representation is first averaged over the token dimension and mapped to a scalar score:

\begin{equation}
s_V
=
f_V
\left(
\operatorname{Mean}(\mathbf{L}_V)
\right),
\qquad
s_A
=
f_A
\left(
\operatorname{Mean}(\mathbf{L}_A)
\right),
\end{equation}

\noindent where $f_V(\cdot)$ and $f_A(\cdot)$ denote learnable linear transformations from the 32-dimensional messenger space to a scalar. The two scores are subsequently normalized using a softmax function:

\begin{equation}
(r_A,r_V)
=
\operatorname{Softmax}(s_A,s_V),
\qquad
r_A+r_V=1,
\end{equation}

The resulting modality weights are used to adaptively combine the two modality-conditioned messenger representations:

\begin{equation}
\mathbf{L}_F
=
r_A \cdot \mathbf{L}_A
+
r_V \cdot \mathbf{L}_V,
\qquad
\mathbf{L}_F \in \mathbb{R}^{4 \times 32},
\end{equation}

This formulation allows the relative contributions of acoustic and visual information to the shared messenger representation to be dynamically adjusted according to the current input, without relying on fixed fusion weights.

The fused messenger representation is subsequently propagated back to both modality-specific branches. In this stage, the visual and acoustic tokens serve as queries, while $\mathbf{L}_F$ provides the keys and values:

\begin{equation}
\mathbf{V}_L
=
\operatorname{MHA}
\left(
\mathbf{V}_t,
\mathbf{L}_F,
\mathbf{L}_F
\right),
\qquad
\mathbf{A}_L
=
\operatorname{MHA}
\left(
\mathbf{A}_t,
\mathbf{L}_F,
\mathbf{L}_F
\right),
\end{equation}

\noindent where
$\mathbf{V}_L,\mathbf{A}_L \in \mathbb{R}^{4 \times 32}$.
Residual connections with the original modality tokens are then applied, followed by layer normalization and token-wise averaging to obtain compact cross-modal context vectors:

\begin{equation}
\mathbf{c}_V
=
\operatorname{Mean}
\left[
\operatorname{LN}
\left(
\mathbf{V}_t+\mathbf{V}_L
\right)
\right],
\qquad
\mathbf{c}_V \in \mathbb{R}^{32},
\end{equation}

\begin{equation}
\mathbf{c}_A
=
\operatorname{Mean}
\left[
\operatorname{LN}
\left(
\mathbf{A}_t+\mathbf{A}_L
\right)
\right],
\qquad
\mathbf{c}_A \in \mathbb{R}^{32},
\end{equation}

The resulting context vectors are projected back to their respective channel dimensions to produce channel-wise cross-modal residual contributions,
$\Delta\mathbf{V} \in \mathbb{R}^{C_V}$
and
$\Delta\mathbf{A} \in \mathbb{R}^{C_A}$.
These vectors are broadcast across the corresponding spatial dimensions and incorporated into the original feature maps through learnable gated residual connections:

\begin{equation}
\widetilde{\mathbf{V}}
=
\mathbf{V}
+
\tanh(\gamma_V)
\cdot
\Delta\mathbf{V},
\qquad
\widetilde{\mathbf{V}}
\in
\mathbb{R}^{C_V \times H_V \times W_V},
\end{equation}

\begin{equation}
\widetilde{\mathbf{A}}
=
\mathbf{A}
+
\tanh(\gamma_A)
\cdot
\Delta\mathbf{A},
\qquad
\widetilde{\mathbf{A}}
\in
\mathbb{R}^{C_A \times H_A \times W_A},
\end{equation}

\noindent where $\gamma_V$ and $\gamma_A$ are learnable scalar parameters initialized to zero. At initialization, $\tanh(\gamma_V)=\tanh(\gamma_A)=0$, so the cross-modal residual contributions 

\noindent are initially inactive and the original modality-specific representations are preserved. Similar zero-initialized $\tanh$ gates have been employed to regulate residual cross-modal contributions~\citep{Alayrac2022}. During training, the learnable gates gradually adjust the strength of the cross-modal residual contributions while preserving the original modality-specific pathways.
 
In addition to refining the modality-specific feature maps, each BMCF module produces a compact representation of the fused cross-modal information. The fused messenger tokens are averaged and normalized to obtain

\begin{equation}
\mathbf{z}_F^{(i)}
=
\operatorname{LN}
\left(
\operatorname{Mean}
\left(
\mathbf{L}_F^{(i)}
\right)
\right),
\qquad
\mathbf{z}_F^{(i)}
\in
\mathbb{R}^{32},
\end{equation}

\noindent where $i$ denotes the corresponding fusion level. Four BMCF modules are applied to the channel-matched visual and acoustic representation levels with channel dimensions of 24, 40, 112, and 320, respectively. The resulting hierarchical cross-modal summaries are subsequently combined with the final visual and acoustic representations for multimodal feature aggregation and feeding intensity prediction.

\subsubsection{Multimodal feature aggregation and prediction}

The four hierarchical BMCF modules generate cross-modal summaries
$\mathbf{z}_F^{(i)} \in \mathbb{R}^{32}$,
where $i \in \{1,2,3,4\}$ denotes the corresponding fusion level. These summaries are concatenated to integrate complementary cross-modal information captured at different representation depths:

\begin{equation}
\mathbf{z}_F
=
\operatorname{Concat}
\left(
\mathbf{z}_F^{(1)},
\mathbf{z}_F^{(2)},
\mathbf{z}_F^{(3)},
\mathbf{z}_F^{(4)}
\right),
\qquad
\mathbf{z}_F \in \mathbb{R}^{128},
\end{equation}

A linear transformation followed by layer normalization is then applied to learn interactions across the hierarchical fusion levels and consolidate them into a unified cross-modal representation:

\begin{equation}
\mathbf{z}_c
=
\operatorname{LN}
\left(
\mathbf{W}_f \mathbf{z}_F + \mathbf{b}_f
\right),
\qquad
\mathbf{z}_c \in \mathbb{R}^{128},
\end{equation}

The resulting cross-modal representation is combined with the final acoustic embedding
$\mathbf{z}_a \in \mathbb{R}^{320}$
and visual embedding
$\mathbf{z}_v \in \mathbb{R}^{320}$
to construct a unified multimodal representation:

\begin{equation}
\mathbf{z}_m
=
\operatorname{Concat}
\left(
\mathbf{z}_a,
\mathbf{z}_v,
\mathbf{z}_c
\right),
\qquad
\mathbf{z}_m \in \mathbb{R}^{768},
\end{equation}

This aggregation preserves modality-specific information from the acoustic and visual branches while incorporating complementary cross-modal information learned across multiple hierarchical levels. The resulting multimodal representation is finally mapped through a fully connected layer to obtain the prediction logits:

\begin{equation}
\mathbf{o}
=
\mathbf{W}_c \mathbf{z}_m + \mathbf{b}_c,
\qquad
\mathbf{o} \in \mathbb{R}^{4},
\end{equation}

\noindent where the four output dimensions correspond to the \textit{None}, \textit{Weak}, \textit{Medium}, and \textit{Strong} feeding intensity levels. The predicted category is determined by the class corresponding to the highest output logit.

\section{Experiments}\label{sec:experiments}

\subsection{Data description}

The AV-FFIA (Audio-Visual Fish Feeding Intensity Assessment) dataset, originally collected and annotated by~\cite{Cui2025}, is a multimodal dataset developed for fish feeding intensity assessment using synchronized acoustic and visual information. The dataset was acquired in a controlled aquaculture environment in Yantai, Shandong Province, China, using \textit{Oplegnathus punctatus}. During feeding sessions, video was recorded at a resolution of $1920 \times 1080$ pixels and 25 frames per second, while underwater acoustic signals were sampled at 256 kHz. The recorded streams were aligned and segmented into 2-second audio-video clips. Each clip was assigned to one of four feeding intensity classes: ``Strong'', ``Medium'', ``Weak'', and ``None'', according to the criteria defined in Table~\ref{tab:annotation_description}. Further details regarding the data acquisition setup and feeding procedures are provided in the original study~\citep{Cui2025}.

\begin{table}[H]
\centering
\caption{Descriptions of fish feeding intensities (FFI).}
\label{tab:annotation_description}
\scalebox{0.75}{
\begin{tabular}{ll}
\toprule
\textbf{FFI} & \textbf{Description} \\
\midrule
Strong & Fish move freely between food items and consume all the available food \\
Medium & Fish move to take the food, but return to their original positions \\
Weak & Fish eat only pellets that fall directly in front of them but do not move to take food \\
None & Fish do not respond to food \\
\bottomrule
\end{tabular}
}
\end{table}

\begin{figure}[H]
    \centering

    \begin{subfigure}[b]{0.243\textwidth}
        \centering
        \includegraphics[width=\textwidth]{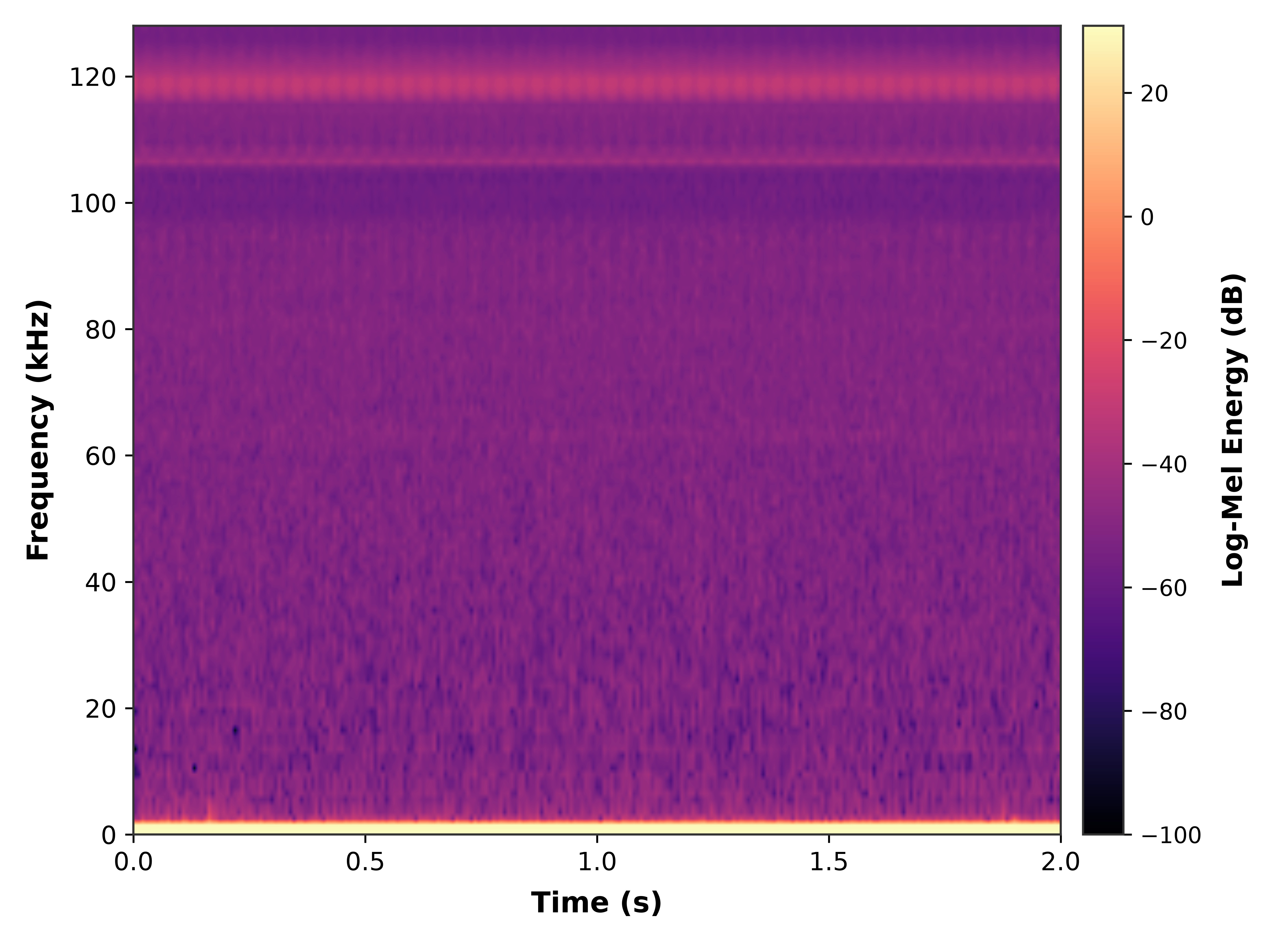}
        \caption{None}
        \label{fig:mel_none}
    \end{subfigure}
    \hfill
    \begin{subfigure}[b]{0.243\textwidth}
        \centering
        \includegraphics[width=\textwidth]{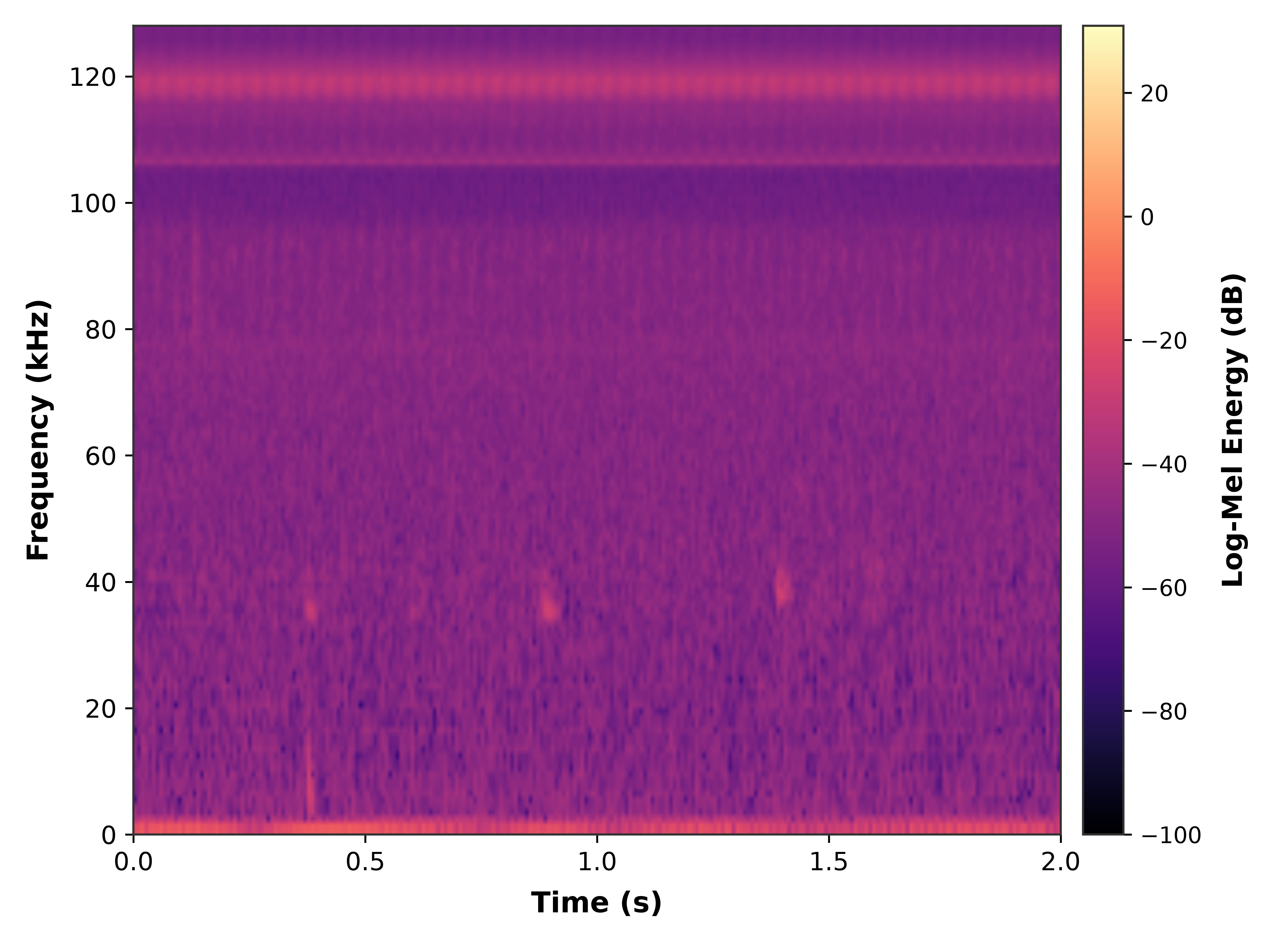}
        \caption{Weak}
        \label{fig:mel_weak}
    \end{subfigure}
    \hfill
    \begin{subfigure}[b]{0.243\textwidth}
        \centering
        \includegraphics[width=\textwidth]{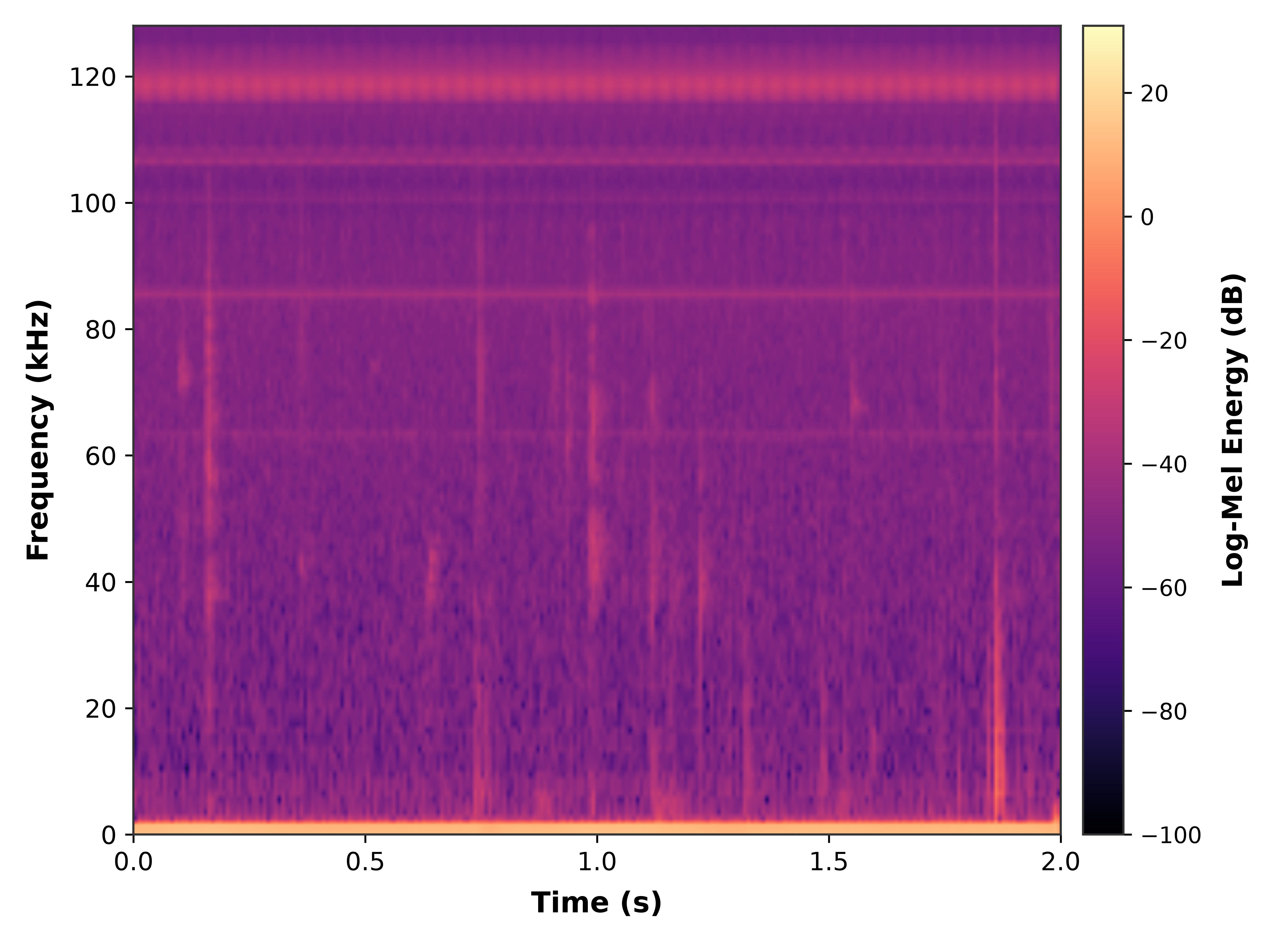}
        \caption{Medium}
        \label{fig:mel_medium}
    \end{subfigure}
    \hfill
    \begin{subfigure}[b]{0.243\textwidth}
        \centering
        \includegraphics[width=\textwidth]{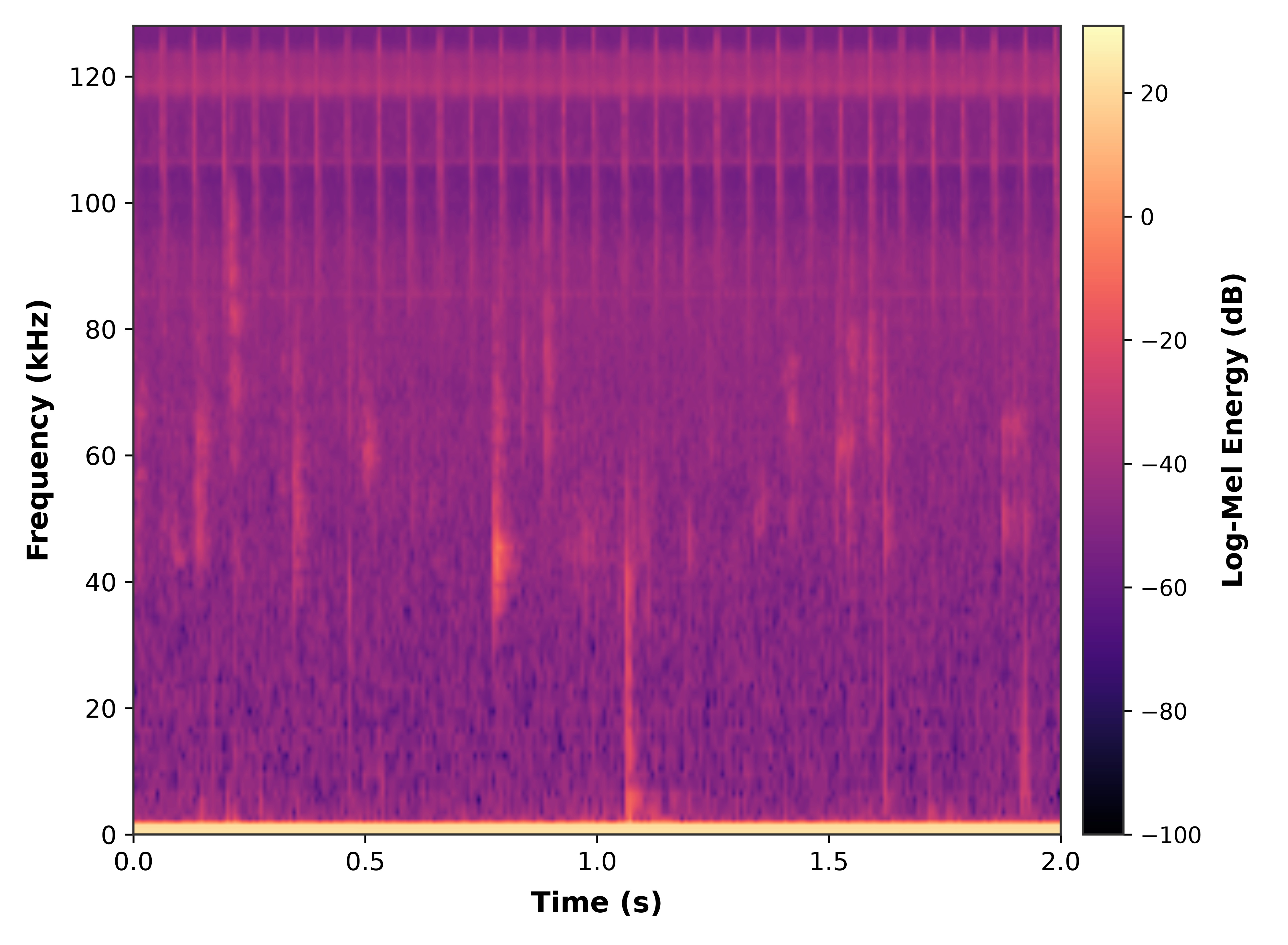}
        \caption{Strong}
        \label{fig:mel_strong}
    \end{subfigure}

    \caption{Representative mel-spectrograms of the four fish feeding intensity classes: (a) None, (b) Weak, (c) Medium, and (d) Strong.}
    \label{fig:mel_spectrogram_classes}
\end{figure}

\begin{figure}[H]
    \centering

    \begin{subfigure}[b]{0.243\textwidth}
        \centering
        \includegraphics[width=\textwidth]{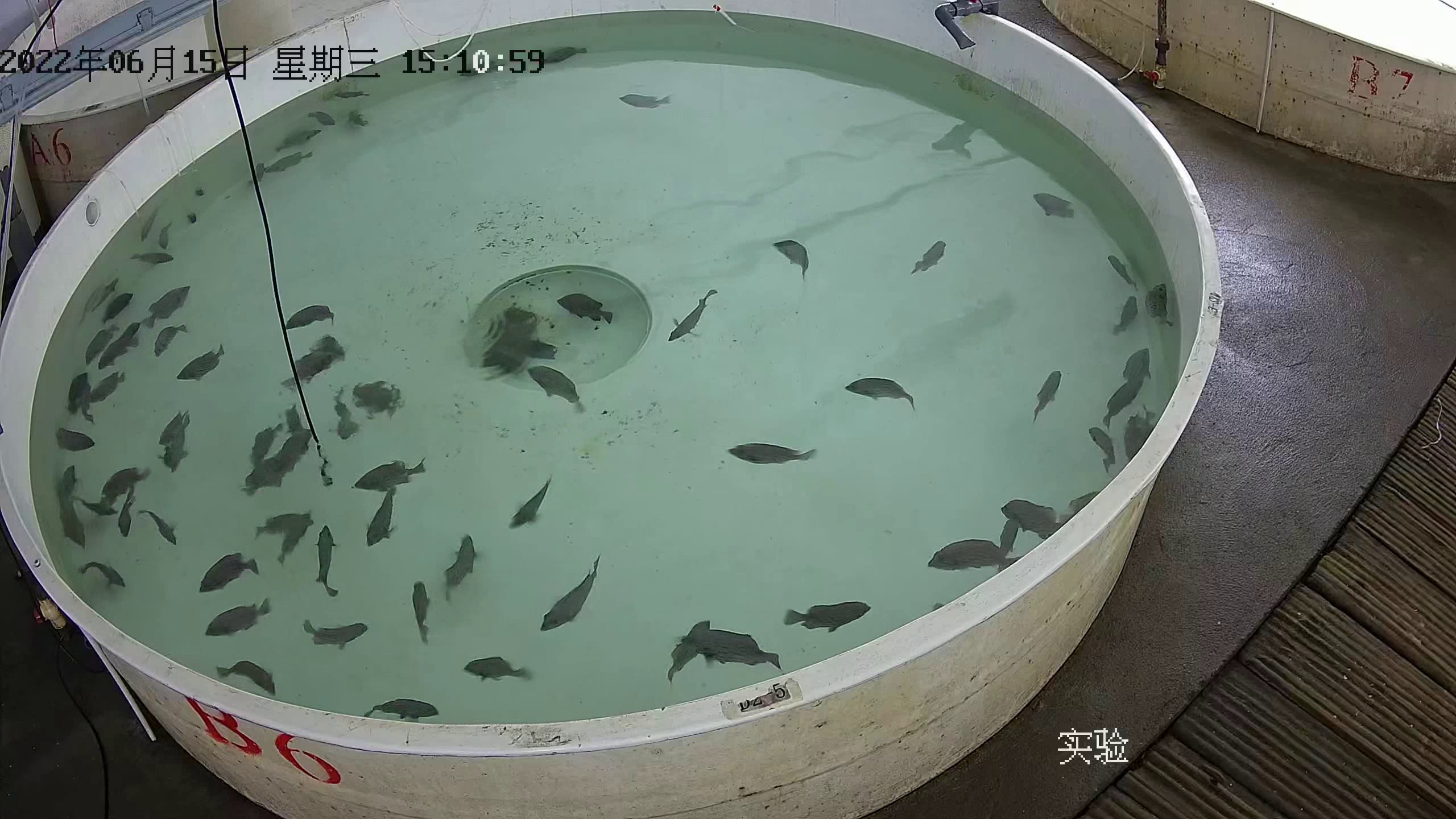}
        \caption{None}
        \label{fig:visual_none}
    \end{subfigure}
    \hfill
    \begin{subfigure}[b]{0.243\textwidth}
        \centering
        \includegraphics[width=\textwidth]{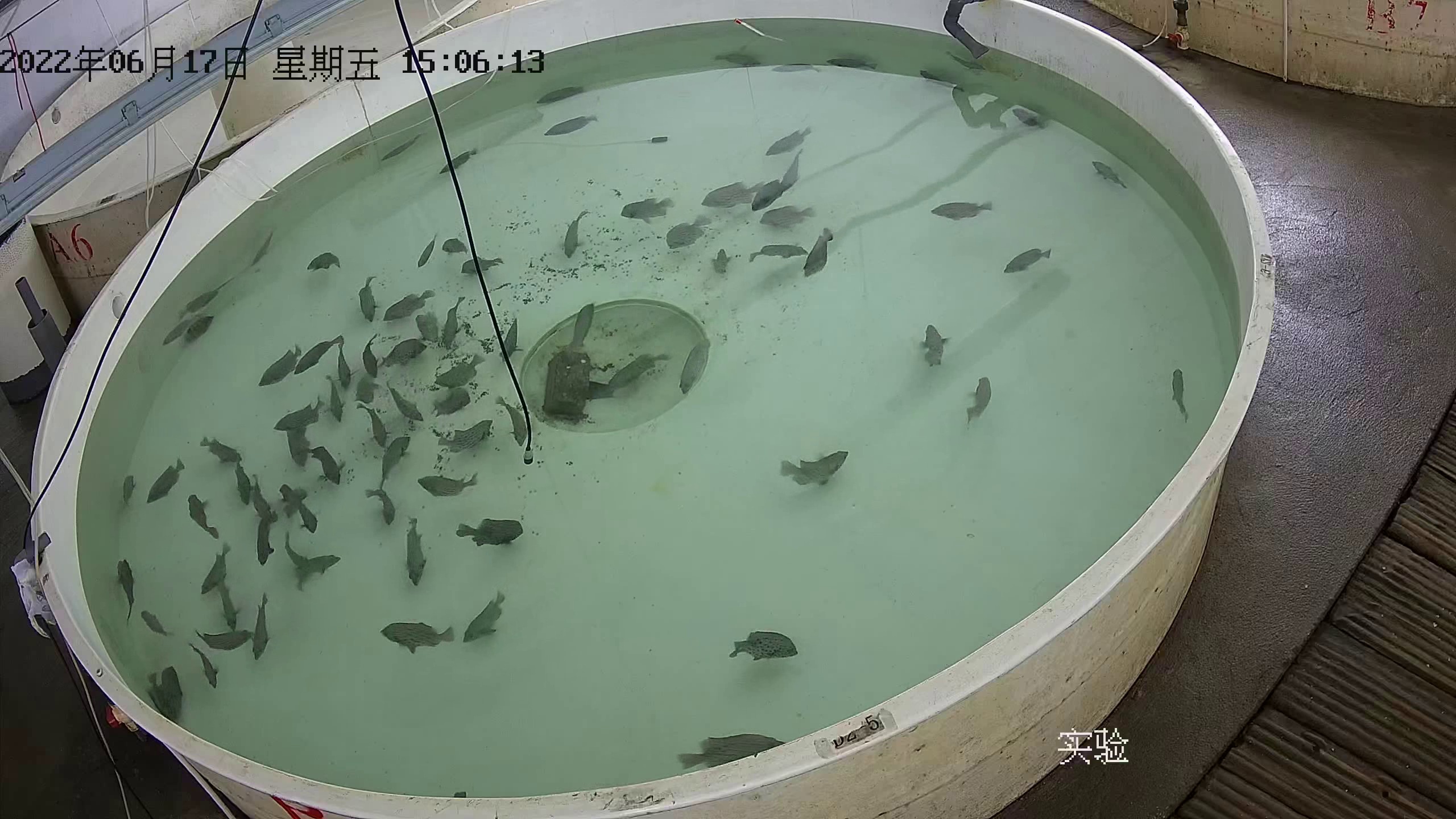}
        \caption{Weak}
        \label{fig:visual_weak}
    \end{subfigure}
    \hfill
    \begin{subfigure}[b]{0.243\textwidth}
        \centering
        \includegraphics[width=\textwidth]{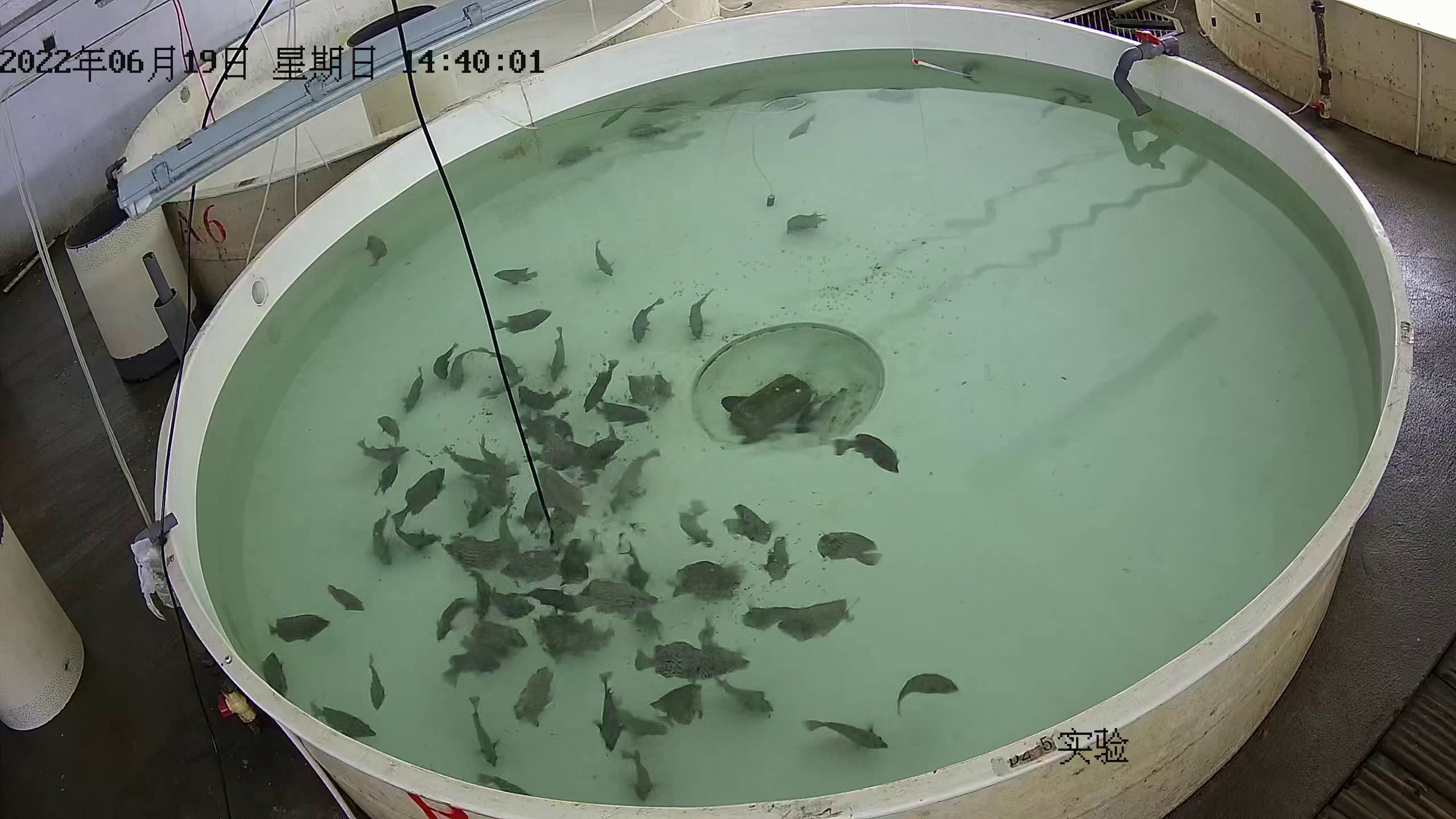}
        \caption{Medium}
        \label{fig:visual_medium}
    \end{subfigure}
    \hfill
    \begin{subfigure}[b]{0.243\textwidth}
        \centering
        \includegraphics[width=\textwidth]{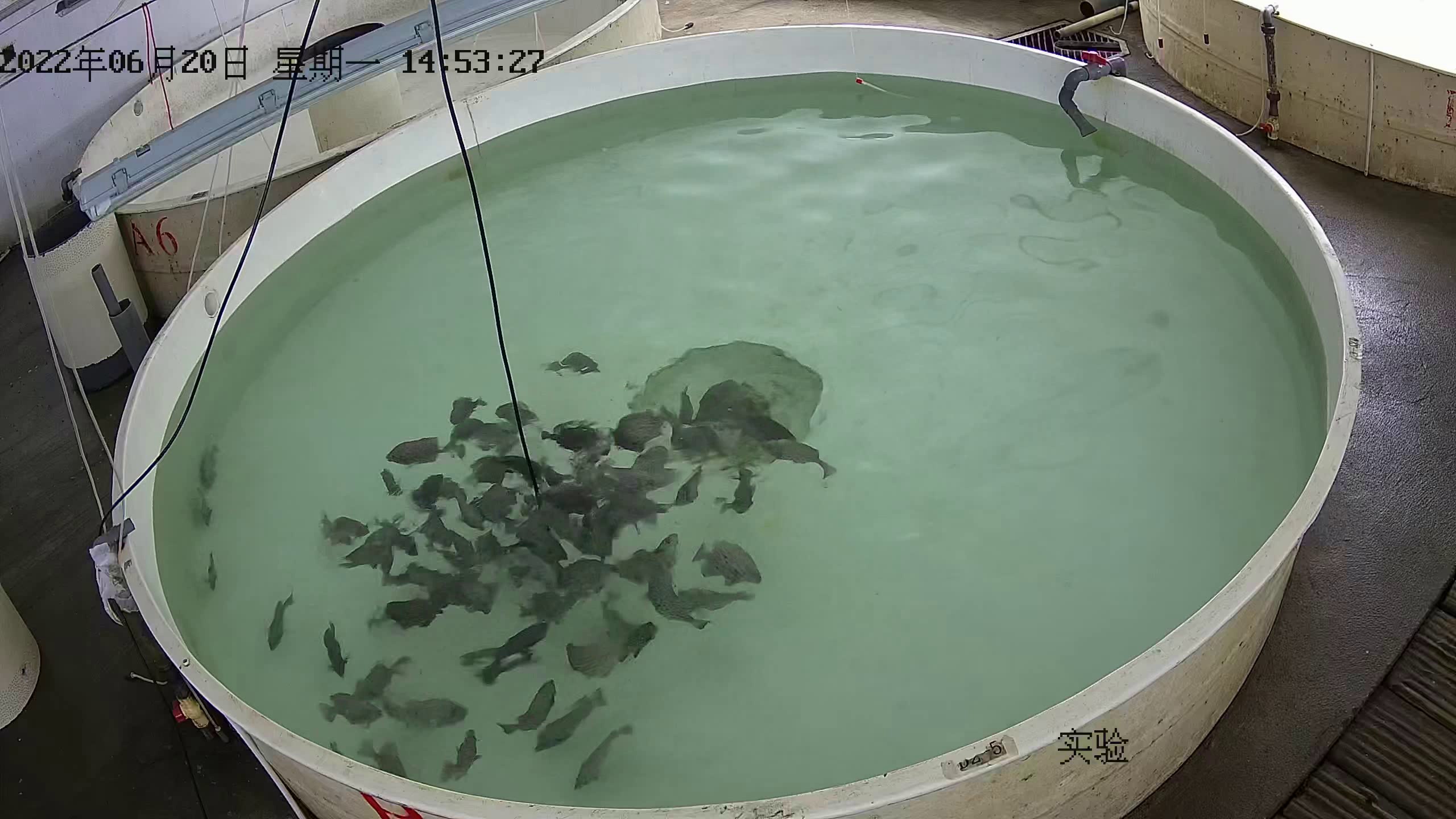}
        \caption{Strong}
        \label{fig:visual_strong}
    \end{subfigure}

    \caption{Representative visual samples of the four fish feeding intensity classes: (a) None, (b) Weak, (c) Medium, and (d) Strong.}
    \label{fig:visual_classes}
\end{figure}

In this study, we reuse the complete publicly available AV-FFIA dataset without removing or filtering samples. The dataset contains 27,067 paired 2-second audio-video clips distributed across four feeding intensity classes: ``None'' (4,871 samples), ``Weak'' (7,979 samples), ``Medium'' (6,640 samples), and ``Strong'' (7,577 samples). Representative mel-spectrograms and visual frames from the four feeding intensity classes are shown in Figures~\ref{fig:mel_spectrogram_classes} and~\ref{fig:visual_classes}, respectively.

\subsection{Experimental setting}

All experiments were conducted on an NVIDIA RTX PRO 6000 Blackwell GPU. Five-fold cross-validation was used to evaluate model performance. In each fold, one subset served as the held-out test set, while the remaining data were split into training and validation sets at an 80:20 ratio. The experimental results are reported as the mean $\pm$ standard deviation across the five folds. The preprocessing, data augmentation, and training configurations used in the experiments are summarized in Table~\ref{tab:configurations}.

\begin{table}[]
\centering
\caption{Preprocessing, augmentation, and training configurations used in the experiments.}
\label{tab:configurations}
\footnotesize
\renewcommand{\arraystretch}{1.35}
\setlength{\tabcolsep}{4pt}

\resizebox{0.95\textwidth}{!}{
\begin{tabular}{
|>{\raggedright\arraybackslash}p{3.3cm}
|>{\raggedright\arraybackslash}p{8.0cm}
|>{\raggedright\arraybackslash}p{5.0cm}|
}
\hline

\textbf{Experiment} &
\textbf{Preprocessing \& augmentation} &
\textbf{Training configurations} \\
\hline

\textbf{Acoustic} &
\textit{Preprocessing:} \newline
Downsampling to 64 kHz \newline
STFT (Hann window = 2048, hop = 1024) \newline
Mel filter banks (128 bins) \newline
\textit{Augmentation:} \newline
SpecAugment \newline
Frequency masking (max width = 8, stripes = 2) \newline
Time masking (max width = 64, stripes = 2)
&
Epochs: 400 \newline
Batch size: 256 \newline
Optimizer: Adam \newline
Learning rate: $1 \times 10^{-3}$ \newline
Loss: Cross-entropy
\\
\hline

\textbf{Visual 2D} \newline
\textbf{(single-frame)} &
\textit{Preprocessing:} \newline
Resize to $224 \times 224$ \newline
\textit{Augmentation:} \newline
Color jitter (brightness = 0.15) \newline
Random horizontal flip ($p=0.5$) \newline
Random rotation ($\pm20^\circ$) \newline
Random affine (translation = 0.2)
&
ImageNet pre-training \newline
Epochs: 250 \newline
Batch size: 256 \newline
Optimizer: Adam \newline
Learning rate: $1 \times 10^{-4}$ \newline
Loss: Cross-entropy \newline
Patience: 40
\\
\hline

\textbf{Visual 2D} \newline
\textbf{(multi-frame)} &
\textit{Preprocessing:} \newline
Resize to $224 \times 224$ \newline
\textit{Augmentation:} \newline
Color jitter (brightness = 0.15) \newline
Random horizontal flip ($p=0.5$) \newline
Random rotation ($\pm20^\circ$) \newline
Random affine (translation = 0.2)
&
ImageNet pre-training \newline
Epochs: 250 \newline
Batch size: 256 \newline
Optimizer: Adam \newline
Learning rate: $1 \times 10^{-4}$ \newline
Loss: Cross-entropy \newline
Patience: 40
\\
\hline

\textbf{Visual 3D} \newline
\textbf{(multi-frame)} &
\textit{Preprocessing:} \newline
20-frame sequence extraction \newline
S3D, I3D, X3D-M: resize to $224 \times 224$ \newline
X3D-XS: resize to $160 \times 160$ \newline
\textit{Augmentation:} \newline
Color jitter (brightness = 0.15) \newline
Random horizontal flip ($p=0.5$) \newline
Random rotation ($\pm20^\circ$) \newline
Random affine (translation = 0.2)
&
ImageNet pre-training \newline
Epochs: 40 \newline
Batch size: 16 \newline
Optimizer: Adam \newline
Learning rate: $1 \times 10^{-4}$ \newline
Loss: Cross-entropy
\\
\hline

\textbf{Multimodal} &
\textit{Visual preprocessing:} \newline
Resize to $224 \times 224$ \newline
\textit{Visual augmentation:} \newline
Color jitter (brightness = 0.15) \newline
Random horizontal flip ($p=0.5$) \newline
Random rotation ($\pm20^\circ$) \newline
Random affine (translation = 0.2) \newline
\textit{Acoustic preprocessing:} \newline
Downsampling to 64 kHz \newline
STFT (Hann window = 2048, hop = 1024) \newline
Mel filter banks (128 bins) \newline
\textit{Acoustic augmentation:} \newline
SpecAugment \newline
Frequency masking (max width = 8, stripes = 2) \newline
Time masking (max width = 64, stripes = 2)
&
Epochs: 400 \newline
Batch size: 256 \newline
Optimizer: Adam \newline
Learning rate: $1 \times 10^{-4}$ \newline
Loss: Cross-entropy \newline
Patience: 50
\\
\hline

\end{tabular}
}

\end{table}

\section{Results and discussion}\label{sec:results_discussion}

This section evaluates the proposed framework from unimodal to multimodal settings. We first examine the visual and acoustic modalities in terms of predictive performance and computational characteristics, followed by an analysis of the proposed multimodal framework and alternative fusion strategies. Finally, the proposed approach is compared with existing methods and further evaluated for system implementation and deployment.

\subsection{Results on the visual modality}

The visual modality is first evaluated using single-frame models, including the effect of frame position on performance. We then investigate the proposed multi-frame representation across different frame combinations and its computational overhead, followed by conventional 3D CNNs to assess the contribution of explicit spatiotemporal modeling.

\subsubsection{Single-frame and the effect of frame positions}

Table~\ref{tab:visual_single_frame_validation} summarizes the performance of the evaluated single-frame models using the last frame (index 49). Swin-Tiny achieves the highest accuracy of $93.33 \pm 0.18\%$ and F1-score of $93.57 \pm 0.19\%$, followed by EfficientNetB0 and ConvNeXt-Tiny with accuracies of $92.62 \pm 0.69\%$ and $92.53 \pm 0.51\%$, respectively. In contrast, ResNet18 and ViT-Base/16 show lower performance under the same setting. These results indicate that single-frame models can capture informative visual cues of feeding activity, but their effectiveness varies across architectures.

\begin{table}[H]
\centering
\caption{Classification performance (\%, mean $\pm$ std) of single-frame models on the validation set using the last frame (index 49).}
\label{tab:visual_single_frame_validation}
\footnotesize
\renewcommand{\arraystretch}{1.2}
\setlength{\tabcolsep}{6pt}

\resizebox{0.85\textwidth}{!}{
\begin{tabular}{|l|c|c|c|c|}
\hline
\textbf{Model} & \textbf{Accuracy} & \textbf{Precision} & \textbf{Recall} & \textbf{F1-score} \\
\hline
MobileNetV2~\citep{Sandler2019} & $91.89 \pm 0.30$ & $92.18 \pm 0.26$ & $92.26 \pm 0.32$ & $92.21 \pm 0.28$ \\
\hline
EfficientNetB0~\citep{Tan2020} & $92.62 \pm 0.69$ & $92.90 \pm 0.64$ & $92.96 \pm 0.68$ & $92.92 \pm 0.66$ \\
\hline
DenseNet121~\citep{Huang2018} & $92.05 \pm 0.54$ & $92.33 \pm 0.55$ & $92.32 \pm 0.50$ & $92.31 \pm 0.51$ \\
\hline
ResNet18~\citep{He2015} & $90.35 \pm 0.66$ & $90.81 \pm 0.71$ & $90.71 \pm 0.56$ & $90.73 \pm 0.63$ \\
\hline
ConvNeXt-Tiny~\citep{Liu2022} & $92.53 \pm 0.51$ & $92.81 \pm 0.42$ & $92.85 \pm 0.58$ & $92.81 \pm 0.50$ \\
\hline
Swin-Tiny~\citep{Liu2021} & $\mathbf{93.33 \pm 0.18}$ & $\mathbf{93.59 \pm 0.17}$ & $\mathbf{93.59 \pm 0.23}$ & $\mathbf{93.57 \pm 0.19}$ \\
\hline
ViT-Base/16~\citep{Dosovitskiy2021} & $90.22 \pm 0.54$ & $90.61 \pm 0.51$ & $90.52 \pm 0.61$ & $90.51 \pm 0.62$ \\
\hline
\end{tabular}
}
\end{table}

\begin{figure}[H]
    \centering
    \includegraphics[width=\textwidth]{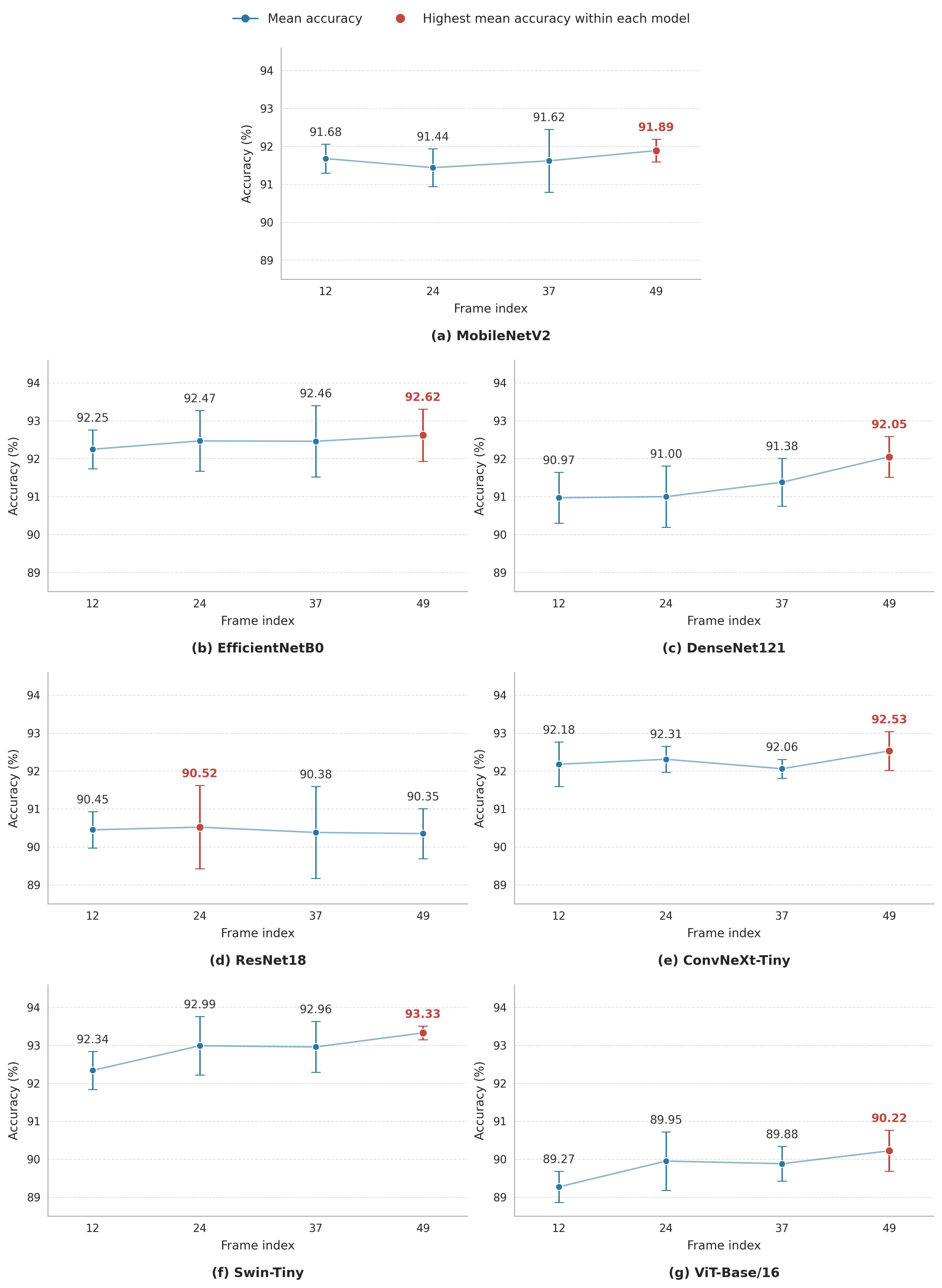}
    \caption{Classification accuracy of single-frame models on the validation set at different frame positions.}
    \label{fig:single_frame_accuracy}
\end{figure}

However, these results are obtained using only the last frame (index 49), raising the question of whether single-frame performance varies with the temporal position of the selected frame. Figure~\ref{fig:single_frame_accuracy} therefore examines classification accuracy across different frame positions within the 50-frame video clips.
For most architectures, the last frame (index 49) achieves the highest mean accuracy. From frame 12 to frame 49, accuracy increases by 0.21, 1.08, 0.37, 0.35, 0.99, and 0.95 percentage points (pp) for MobileNetV2, DenseNet121, EfficientNetB0, ConvNeXt-Tiny, Swin-Tiny, and ViT-Base/16, respectively. ResNet18 is the only exception, achieving its highest accuracy at frame 24, although its variation across frame positions remains relatively small.

The higher performance of later frames may be associated with the progressive development of visible feeding responses within each clip. As feeding

\noindent activity progresses, surface disturbances such as ripples and splashes may become more pronounced, providing stronger spatial cues for distinguishing feeding intensity levels. Earlier frames may contain less developed or more transient visual responses, making adjacent intensity levels more difficult to distinguish. Nevertheless, the variation across frame positions highlights the limitation of relying on a single temporal observation, as one frame may not fully capture the evolution of feeding activity. This motivates the incorporation of multiple temporal positions in the subsequent multi-frame evaluation.

\subsubsection{Multi-frame and the effect of frame selection}

To exploit complementary visual information across temporal positions, the proposed early multi-frame fusion strategy concatenates selected frames along the channel dimension before feature extraction. Different frame selection policies are investigated using the seven 2D architectures, with the last frame (index 49) used as the single-frame baseline. Table~\ref{tab:ablation_frame_policy} reports the predictive performance on the validation set, together with the corresponding parameter count and computational cost.

The two-frame configuration $(0,49)$ consistently improves predictive performance over the corresponding single-frame baseline across all seven models. Accuracy increases by 2.65, 2.87, 2.13, 3.29, 2.82, 1.94, and 2.64 percentage points for MobileNetV2, EfficientNetB0, DenseNet121, ResNet18, ConvNeXt-Tiny, Swin-Tiny, and ViT-Base/16, respectively. Similar improvements are observed in precision, recall, and F1-score, indicating that the gain is consistent across the evaluation metrics. The improvement may be associated with complementary visual information captured at temporally distant positions, where the first and last frames correspond to different stages of the feeding activity. Combining these observations at the input level therefore provides additional temporal context while retaining the 2D feature extraction architecture.

Increasing the number of concatenated frames, however, does not consistently improve performance. The $(0,49)$ configuration achieves the highest accuracy for six of the seven models, with DenseNet121 being the only exception, for which $(12,24,49)$ performs best. In most cases, the three- and four-frame configurations yield lower accuracy than $(0,49)$. This result suggests that selecting informative temporal positions may be more important than simply increasing the number of frames. The lower performance of denser frame selections may partly reflect redundancy among temporally adjacent frames. In addition, because the proposed early fusion concatenates frames directly along the channel dimension, temporal order and motion are not explicitly modeled, which may limit the benefit of adding further frames.

The performance gains are achieved with a relatively small increase in computational cost for most CNN- and Transformer-based models. For example, EfficientNetB0 increases from 4.013M to 4.014M parameters and from 0.828 to 0.849 GFLOPs, while MobileNetV2 increases by only 0.001M parameters and 0.022 GFLOPs under the $(0,49)$ configuration. Similar trends are observed for ConvNeXt-Tiny and Swin-Tiny. ViT-Base/16 exhibits a larger parameter increase, from 85.802M to 86.392M, together with a moderate increase in GFLOPs. These results indicate that early multi-frame fusion can incorporate complementary information from multiple temporal positions with limited additional computational cost for most CNN-based architectures.

\begin{table}[H]
\centering
\caption{Evaluation of different frame integration configurations in terms of classification performance (\%, mean $\pm$ std) and computational cost on the validation set}
\label{tab:ablation_frame_policy}
\footnotesize
\renewcommand{\arraystretch}{1.2}
\setlength{\tabcolsep}{6pt}

\resizebox{\textwidth}{!}{
\begin{tabular}{|l|l|c|c|c|c|c|c|}
\hline
\textbf{Model} & \textbf{Frame policy} & \textbf{Accuracy} & \textbf{Precision} & \textbf{Recall} & \textbf{F1-score} & \textbf{Params (M)} & \textbf{GFLOPs} \\
\hline

\multirow{6}{*}{MobileNetV2}
 & Single frame 49 (baseline) & $91.89 \pm 0.30$ & $92.18 \pm 0.26$ & $92.26 \pm 0.32$ & $92.21 \pm 0.28$ & 2.229 & 0.652 \\
\cline{2-8}
 & Multi-frame (0, 49) & ${\mathbf{94.54 \pm 0.29}}\gain{2.65}$ & ${\mathbf{94.66 \pm 0.32}}\gain{2.48}$ & ${\mathbf{94.75 \pm 0.25}}\gain{2.49}$ & ${\mathbf{94.70 \pm 0.29}}\gain{2.49}$ & $2.230\gain{0.001}$ & $0.674\gain{0.022}$ \\
 & Multi-frame (12, 49) & $93.62 \pm 0.49$ & $93.80 \pm 0.47$ & $93.89 \pm 0.48$ & $93.83 \pm 0.48$ & 2.230 & 0.674 \\
 & Multi-frame (24, 49) & $93.21 \pm 0.40$ & $93.38 \pm 0.41$ & $93.44 \pm 0.42$ & $93.41 \pm 0.42$ & 2.230 & 0.674 \\
 & Multi-frame (12, 24, 49) & $93.70 \pm 0.40$ & $93.85 \pm 0.38$ & $93.97 \pm 0.38$ & $93.89 \pm 0.38$ & 2.231 & 0.696 \\
 & Multi-frame (12, 24, 37, 49) & $93.54 \pm 0.50$ & $93.68 \pm 0.51$ & $93.81 \pm 0.50$ & $93.74 \pm 0.50$ & 2.232 & 0.717 \\
\hline

\multirow{6}{*}{EfficientNetB0}
 & Single frame 49 (baseline) & $92.62 \pm 0.69$ & $92.90 \pm 0.64$ & $92.96 \pm 0.68$ & $92.92 \pm 0.66$ & 4.013 & 0.828 \\
\cline{2-8}
 & Multi-frame (0, 49) & ${\mathbf{95.49 \pm 0.35}}\gain{2.87}$ & ${\mathbf{95.60 \pm 0.32}}\gain{2.70}$ & ${\mathbf{95.64 \pm 0.35}}\gain{2.68}$ & ${\mathbf{95.61 \pm 0.34}}\gain{2.69}$ & $4.014\gain{0.001}$ & $0.849\gain{0.021}$ \\
 & Multi-frame (12, 49) & $94.23 \pm 0.37$ & $94.41 \pm 0.37$ & $94.48 \pm 0.37$ & $94.44 \pm 0.36$ & 4.014 & 0.849 \\
 & Multi-frame (24, 49) & $93.85 \pm 0.56$ & $94.02 \pm 0.51$ & $94.07 \pm 0.61$ & $94.03 \pm 0.56$ & 4.014 & 0.849 \\
 & Multi-frame (12, 24, 49) & $94.19 \pm 0.48$ & $94.33 \pm 0.49$ & $94.35 \pm 0.46$ & $94.33 \pm 0.47$ & 4.014 & 0.871 \\
 & Multi-frame (12, 24, 37, 49) & $94.16 \pm 0.30$ & $94.36 \pm 0.26$ & $94.33 \pm 0.35$ & $94.34 \pm 0.31$ & 4.015 & 0.893 \\
\hline

\multirow{6}{*}{DenseNet121}
 & Single frame 49 (baseline) & $92.05 \pm 0.54$ & $92.33 \pm 0.55$ & $92.32 \pm 0.50$ & $92.31 \pm 0.51$ & 6.958 & 5.792 \\
\cline{2-8}
 & Multi-frame (0, 49) & ${94.18 \pm 0.66}\gain{2.13}$ & ${94.34 \pm 0.64}\gain{2.01}$ & ${94.32 \pm 0.69}\gain{2.00}$ & ${94.32 \pm 0.67}\gain{2.01}$ & $6.967\gain{0.009}$ & $6.028\gain{0.236}$ \\
 & Multi-frame (12, 49) & $92.84 \pm 0.78$ & $93.05 \pm 0.72$ & $93.11 \pm 0.75$ & $93.07 \pm 0.73$ & 6.967 & 6.028 \\
 & Multi-frame (24, 49) & $92.01 \pm 0.71$ & $92.32 \pm 0.68$ & $92.34 \pm 0.64$ & $92.28 \pm 0.66$ & 6.967 & 6.028 \\
 & Multi-frame (12, 24, 49) & $\mathbf{94.39 \pm 0.46}$ & $\mathbf{94.53 \pm 0.45}$ & $\mathbf{94.58 \pm 0.45}$ & $\mathbf{94.55 \pm 0.45}$ & 6.977 & 6.264 \\
 & Multi-frame (12, 24, 37, 49) & $92.67 \pm 0.44$ & $92.97 \pm 0.42$ & $92.88 \pm 0.41$ & $92.90 \pm 0.41$ & 6.986 & 6.500 \\
\hline

\multirow{6}{*}{ResNet18}
 & Single frame 49 (baseline) & $90.35 \pm 0.66$ & $90.81 \pm 0.71$ & $90.71 \pm 0.56$ & $90.73 \pm 0.63$ & 11.179 & 3.647 \\
\cline{2-8}
 & Multi-frame (0, 49) & ${\mathbf{93.64 \pm 0.82}}\gain{3.29}$ & ${\mathbf{93.80 \pm 0.78}}\gain{2.99}$ & ${\mathbf{93.85 \pm 0.88}}\gain{3.14}$ & ${\mathbf{93.81 \pm 0.84}}\gain{3.08}$ & $11.188\gain{0.009}$ & $3.883\gain{0.236}$ \\
 & Multi-frame (12, 49) & $93.27 \pm 0.54$ & $93.44 \pm 0.53$ & $93.57 \pm 0.53$ & $93.48 \pm 0.53$ & 11.188 & 3.883 \\
 & Multi-frame (24, 49) & $92.37 \pm 0.85$ & $92.59 \pm 0.81$ & $92.60 \pm 0.84$ & $92.57 \pm 0.83$ & 11.188 & 3.883 \\
 & Multi-frame (12, 24, 49) & $92.60 \pm 0.39$ & $92.79 \pm 0.45$ & $92.84 \pm 0.35$ & $92.79 \pm 0.39$ & 11.197 & 4.119 \\
 & Multi-frame (12, 24, 37, 49) & $92.55 \pm 0.77$ & $92.77 \pm 0.70$ & $92.87 \pm 0.73$ & $92.81 \pm 0.71$ & 11.207 & 4.355 \\
\hline

\multirow{6}{*}{ConvNeXt-Tiny}
 & Single frame 49 (baseline) & $92.53 \pm 0.51$ & $92.81 \pm 0.42$ & $92.85 \pm 0.58$ & $92.81 \pm 0.50$ & 27.823 & 8.927 \\
\cline{2-8}
 & Multi-frame (0, 49) & ${\mathbf{95.35 \pm 0.37}}\gain{2.82}$ & ${\mathbf{95.48 \pm 0.35}}\gain{2.67}$ & ${\mathbf{95.49 \pm 0.34}}\gain{2.64}$ & ${\mathbf{95.47 \pm 0.35}}\gain{2.66}$ & $27.828\gain{0.005}$ & $8.956\gain{0.029}$ \\
 & Multi-frame (12, 49) & $93.98 \pm 0.37$ & $94.17 \pm 0.40$ & $94.17 \pm 0.36$ & $94.15 \pm 0.36$ & 27.828 & 8.956 \\
 & Multi-frame (24, 49) & $93.49 \pm 0.52$ & $93.65 \pm 0.53$ & $93.78 \pm 0.50$ & $93.70 \pm 0.52$ & 27.828 & 8.956 \\
 & Multi-frame (12, 24, 49) & $93.83 \pm 0.59$ & $94.02 \pm 0.55$ & $94.02 \pm 0.56$ & $94.00 \pm 0.56$ & 27.832 & 8.985 \\
 & Multi-frame (12, 24, 37, 49) & $93.49 \pm 0.70$ & $93.71 \pm 0.66$ & $93.81 \pm 0.74$ & $93.73 \pm 0.69$ & 27.837 & 9.013 \\
\hline

\multirow{6}{*}{Swin-Tiny}
 & Single frame 49 (baseline) & $93.33 \pm 0.18$ & $93.59 \pm 0.17$ & $93.59 \pm 0.23$ & $93.57 \pm 0.19$ & 27.522 & 5.955 \\
\cline{2-8}
 & Multi-frame (0, 49) & ${\mathbf{95.27 \pm 0.47}}\gain{1.94}$ & ${\mathbf{95.40 \pm 0.46}}\gain{1.81}$ & ${\mathbf{95.44 \pm 0.44}}\gain{1.85}$ & ${\mathbf{95.42 \pm 0.44}}\gain{1.85}$ & $27.527\gain{0.005}$ & $5.984\gain{0.029}$ \\
 & Multi-frame (12, 49) & $94.20 \pm 0.81$ & $94.41 \pm 0.75$ & $94.41 \pm 0.83$ & $94.40 \pm 0.78$ & 27.527 & 5.984 \\
 & Multi-frame (24, 49) & $94.17 \pm 0.46$ & $94.34 \pm 0.45$ & $94.39 \pm 0.45$ & $94.36 \pm 0.44$ & 27.527 & 5.984 \\
 & Multi-frame (12, 24, 49) & $94.39 \pm 0.46$ & $94.53 \pm 0.45$ & $94.58 \pm 0.45$ & $94.55 \pm 0.45$ & 27.532 & 6.012 \\
 & Multi-frame (12, 24, 37, 49) & $94.35 \pm 0.42$ & $94.47 \pm 0.45$ & $94.64 \pm 0.39$ & $94.55 \pm 0.42$ & 27.536 & 6.041 \\
\hline

\multirow{6}{*}{ViT-Base/16}
 & Single frame 49 (baseline) & $90.22 \pm 0.54$ & $90.61 \pm 0.51$ & $90.52 \pm 0.61$ & $90.51 \pm 0.62$ & 85.802 & 22.571 \\
\cline{2-8}
 & Multi-frame (0, 49) & ${\mathbf{92.86 \pm 0.90}}\gain{2.64}$ & ${\mathbf{93.03 \pm 0.83}}\gain{2.42}$ & ${\mathbf{93.19 \pm 0.87}}\gain{2.67}$ & ${\mathbf{93.08 \pm 0.85}}\gain{2.57}$ & $86.392\gain{0.590}$ & $22.802\gain{0.231}$ \\
 & Multi-frame (12, 49) & $92.33 \pm 0.79$ & $92.56 \pm 0.74$ & $92.63 \pm 0.75$ & $92.57 \pm 0.76$ & 86.392 & 22.802 \\
 & Multi-frame (24, 49) & $91.19 \pm 0.82$ & $91.45 \pm 0.68$ & $91.54 \pm 0.81$ & $91.47 \pm 0.77$ & 86.392 & 22.802 \\
 & Multi-frame (12, 24, 49) & $91.68 \pm 1.03$ & $91.98 \pm 0.99$ & $91.85 \pm 1.01$ & $91.89 \pm 1.01$ & 86.981 & 23.033 \\
 & Multi-frame (12, 24, 37, 49) & $91.63 \pm 0.42$ & $91.85 \pm 0.36$ & $91.98 \pm 0.48$ & $91.90 \pm 0.41$ & 87.574 & 23.265 \\
\hline
\end{tabular}
}
\par\vspace{0.1cm}
\noindent
\begin{minipage}{\textwidth}
\tiny
\textit{Note:} Red values in parentheses indicate improvements in percentage points over the corresponding single-frame baseline.
\end{minipage}
\end{table}

\subsubsection{Comparison with 3D architectures}

The effectiveness of explicit spatiotemporal modeling is further examined using four representative 3D CNN architectures with 20 input frames sampled from each 2-s video clip. As shown in Table~\ref{tab:visual_3d_cnn_validation}, X3D-M achieves the highest accuracy among the evaluated 3D models at $93.49 \pm 0.66\%$, with an F1-score of $93.72 \pm 0.65\%$, followed by X3D-XS at $93.14 \pm 0.48\%$. X3D-M slightly outperforms the best single-frame model, Swin-Tiny, by 0.16 pp. However, its accuracy remains 2.00 pp below the proposed EfficientNetB0-based $(0,49)$ configuration, which achieves $95.49 \pm 0.35\%$. These results indicate that incorporating multiple temporal observations through explicit 3D modeling does not necessarily provide greater performance gains than the proposed early fusion of selected 2D frames.

\begin{table}[H]
\centering
\caption{Classification performance (\%, mean $\pm$ std) and computational cost of 3D CNN models using 20 frames on the validation set.}
\label{tab:visual_3d_cnn_validation}
\footnotesize
\renewcommand{\arraystretch}{1.2}
\setlength{\tabcolsep}{6pt}

\resizebox{\textwidth}{!}{
\begin{tabular}{|l|c|c|c|c|c|c|}
\hline
\textbf{Model} & \textbf{Accuracy} & \textbf{Precision} & \textbf{Recall} & \textbf{F1-score} & \textbf{Params (M)} & \textbf{GFLOPs} \\
\hline
S3D~\citep{Xie2018} & $92.26 \pm 0.76$ & $92.68 \pm 0.75$ & $92.45 \pm 0.78$ & $92.46 \pm 0.80$ & 7.914 & 45.231 \\
\hline
I3D~\citep{Carreira2018} & $92.91 \pm 0.78$ & $93.13 \pm 0.69$ & $93.24 \pm 0.79$ & $93.16 \pm 0.74$ & 27.232 & 143.341 \\
\hline
X3D-M~\citep{Feichtenhofer2020} & $93.49 \pm 0.66$ & $93.67 \pm 0.66$ & $93.82 \pm 0.66$ & $93.72 \pm 0.65$ & 2.983 & 12.655 \\
\hline
X3D-XS~\citep{Feichtenhofer2020} & $93.14 \pm 0.48$ & $93.27 \pm 0.47$ & $93.44 \pm 0.42$ & $93.34 \pm 0.45$ & 2.983 & 6.482 \\
\hline
\end{tabular}
}
\end{table}

The results also show that increasing the computational capacity of the 3D models does not lead to a proportional performance improvement. X3D-XS and X3D-M have the same parameter count of 2.983M, while increasing GFLOPs from 6.482 to 12.655 yields only a 0.35 pp improvement in accuracy. The lower performance of dense 3D processing may partly reflect redundancy among neighboring frames, particularly because 20 frames are sampled from 2-s clips. In contrast, the proposed $(0,49)$ configuration uses only two temporally distant observations, which may provide more complementary information with fewer redundant frames.

A substantial difference is observed in computational cost. X3D-XS and X3D-M require 6.482 and 12.655 GFLOPs, respectively, while S3D and I3D require 45.231 and 143.341 GFLOPs. By comparison, the proposed two-frame EfficientNetB0 configuration requires only 0.849 GFLOPs, corresponding to approximately 7.6$\times$ and 14.9$\times$ lower computation than X3D-XS and X3D-M, respectively. The results suggest that explicit 3D spatiotemporal modeling can provide competitive visual performance, but the proposed early multi-frame fusion achieves higher accuracy with substantially lower computational cost for incorporating temporal information in the present task.

\subsection{Results on the audio modality}

Following the visual modality, the acoustic modality is evaluated using different deep learning models in terms of predictive performance and computational cost. The effect of audio sampling rate is then investigated using selected models to assess its impact on model performance.

\subsubsection{Acoustic model performance}

Table~\ref{tab:acoustic_validation} 
summarizes the classification performance and computational cost of the evaluated acoustic models on the validation set. Among the conventional acoustic architectures, the PANNs-based models achieve higher predictive performance than MobileNetV1, MobileNetV2, EfficientNetB0, and ResNet22. PANNs CNN6 obtains the highest accuracy of $82.64 \pm 0.45\%$ and precision of $83.32 \pm 0.60\%$, while PANNs CNN14 achieves the highest recall of $83.19 \pm 0.59\%$ and F1-score of $83.14 \pm 0.54\%$. PANNs CNN10 also achieves competitive performance with an accuracy of $81.18 \pm 0.32\%$. This performance advantage may be related to the design of PANNs for audio representation learning, which may provide more suitable feature extraction for the time--frequency patterns in the Mel-spectrogram inputs.

\begin{table}[H]
\centering
\caption{Classification performance (\%, mean $\pm$ std) and computational cost of acoustic models on the validation set.}
\label{tab:acoustic_validation}
\footnotesize
\renewcommand{\arraystretch}{1.2}
\setlength{\tabcolsep}{6pt}

\resizebox{\textwidth}{!}{
\begin{tabular}{|l|c|c|c|c|c|c|}
\hline
\textbf{Model} & \textbf{Accuracy} & \textbf{Precision} & \textbf{Recall} & \textbf{F1-score} & \textbf{Params (M)} & \textbf{GFLOPs} \\
\hline
Proposed AMARU-Net & $78.87 \pm 0.42$ & $79.62 \pm 0.14$ & $79.56 \pm 0.30$ & $79.44 \pm 0.29$ & \textbf{0.238} & \textbf{0.193} \\
\hline
MobileNetV1~\citep{Howard2017} & $79.13 \pm 0.91$ & $79.60 \pm 0.90$ & $79.76 \pm 0.89$ & $79.62 \pm 0.88$ & 4.260 & 0.397 \\
\hline
MobileNetV2~\citep{Sandler2019} & $79.03 \pm 0.45$ & $79.61 \pm 0.42$ & $79.77 \pm 0.39$ & $79.66 \pm 0.41$ & 3.539 & 0.195 \\
\hline
EfficientNetB0~\citep{Tan2020} & $77.09 \pm 0.97$ & $77.63 \pm 0.83$ & $77.62 \pm 0.87$ & $77.53 \pm 0.77$ & 4.012 & 0.266 \\
\hline
ResNet22~\citep{Kong2020} & $80.73 \pm 0.64$ & $81.44 \pm 0.69$ & $81.23 \pm 0.65$ & $81.25 \pm 0.67$ & 62.603 & 7.209 \\
\hline
PANNs CNN14~\citep{Kong2020} & $82.53 \pm 0.48$ & $83.22 \pm 0.41$ & $\mathbf{83.19 \pm 0.59}$ & $\mathbf{83.14 \pm 0.54}$ & 79.682 & 10.328 \\
\hline
PANNs CNN10~\citep{Kong2020} & $81.18 \pm 0.32$ & $81.88 \pm 0.54$ & $81.85 \pm 0.36$ & $81.79 \pm 0.47$ & 4.951 & 6.695 \\
\hline
PANNs CNN6~\citep{Kong2020} & $\mathbf{82.64 \pm 0.45}$ & $\mathbf{83.32 \pm 0.60}$ & $83.14 \pm 0.57$ & $83.12 \pm 0.64$ & 4.569 & 5.102 \\
\hline
\end{tabular}
}
\end{table}

However, model size does not necessarily translate into better performance. PANNs CNN14 contains 79.682M parameters and requires 10.328 GFLOPs, yet its accuracy of $82.53 \pm 0.48\%$ is slightly lower than that of CNN6, which uses only 4.569M parameters and 5.102 GFLOPs. Similarly, ResNet22 contains 62.603M parameters and requires 7.209 GFLOPs but achieves an accuracy of $80.73 \pm 0.64\%$. These results indicate that larger model capacity does not necessarily translate into better performance for the present acoustic task. Among the conventional models, CNN6 therefore provides the strongest standalone acoustic performance with substantially lower computational cost than the larger PANNs and ResNet22 models.

Building on this accuracy and efficiency analysis, the proposed AMARU-Net is designed with a different priority: reducing the computational burden of the acoustic branch for multimodal integration and deployment. Although its accuracy of $78.87 \pm 0.42\%$ is lower than that of the conventional acoustic models, it requires only 0.238M parameters and 0.193 GFLOPs. This corresponds to a substantially more compact acoustic branch, which is important for the proposed multimodal framework and its deployment-oriented design. Therefore, CNN6 provides the strongest standalone acoustic performance, whereas AMARU-Net is selected as the acoustic branch because of its substantially lower computational cost and compact architecture. The subsequent multimodal experiments investigate whether this lightweight acoustic branch can provide complementary information to the visual modality and improve the overall feeding intensity assessment.

\subsubsection{Effect of audio sampling rate on acoustic performance}

To investigate the effect of audio sampling rate on acoustic representation learning, AMARU-Net and the three PANNs-based models were evaluated at 48 kHz, 64 kHz, and 128 kHz. Table \ref{tab:acoustic_sampling_rate} reports the classification performance on the validation set under five-fold cross-validation.

A consistent improvement is observed when increasing the sampling rate from 48 kHz to 64 kHz across all four acoustic models. AMARU-Net improves from $76.47 \pm 0.97\%$ to $78.87 \pm 0.42\%$ in accuracy, while PANNs CNN6, CNN10, and CNN14 improve by 2.59, 2.76, and 2.26 percentage points, respectively. The improvements are also consistently reflected in precision, recall, and F1-score. This pattern indicates that increasing the sampling rate to 64 kHz provides additional acoustic information that is useful for distinguishing feeding intensity levels.

However, increasing the sampling rate further to 128 kHz does not provide additional predictive benefits. AMARU-Net decreases from $78.87 \pm 0.42\%$ to $78.07 \pm 0.80\%$, while CNN10 and CNN14 decrease slightly by 0.05 and 0.32 percentage points, respectively. CNN6 remains nearly unchanged, with accuracy decreasing from $82.64 \pm 0.45\%$ to $82.61 \pm 0.48\%$. Similar patterns are observed for precision, recall, and F1-score. Thus, although 128 kHz retains a wider frequency range, the additional high-frequency information does not lead to improved discrimination of feeding intensity levels. With the STFT configuration and the number of Mel filter banks kept fixed, the wider frequency range is also represented within the same feature dimensionality, which may limit the benefit of the additional spectral content.

Based on these results, 64 kHz is selected for the subsequent acoustic experiments. It provides the best performance for all four acoustic models and yields a clear improvement over 48 kHz, whereas increasing the sampling rate to 128 kHz offers no consistent performance gain. The 64 kHz setting therefore provides a favorable balance between acoustic information and computational efficiency.

\begin{table}[H]
\centering
\caption{Classification performance (\%, mean $\pm$ std) on the validation set of acoustic models at different audio sampling rates.}
\label{tab:acoustic_sampling_rate}
\footnotesize
\renewcommand{\arraystretch}{1.2}
\setlength{\tabcolsep}{6pt}

\resizebox{0.85\textwidth}{!}{
\begin{tabular}{|l|c|c|c|c|c|}
\hline
\textbf{Model} & \textbf{Sampling rate (kHz)} & \textbf{Accuracy} & \textbf{Precision} & \textbf{Recall} & \textbf{F1-score} \\
\hline
\multirow{3}{*}{AMARU-Net}
& 48 & $76.47 \pm 0.97$ & $76.90 \pm 1.15$ & $77.19 \pm 0.68$ & $76.93 \pm 0.92$ \\
\cline{2-6}
& \textbf{64} & $\mathbf{78.87 \pm 0.42}$ & $\mathbf{79.62 \pm 0.14}$ & $\mathbf{79.56 \pm 0.30}$ & $\mathbf{79.44 \pm 0.29}$ \\
\cline{2-6}
& 128 & $78.07 \pm 0.80$ & $78.87 \pm 0.71$ & $78.71 \pm 0.83$ & $78.64 \pm 0.84$ \\
\hline
\multirow{3}{*}{PANNs CNN6}
& 48 & $80.05 \pm 0.65$ & $80.61 \pm 0.81$ & $80.75 \pm 0.53$ & $80.56 \pm 0.71$ \\
\cline{2-6}
& \textbf{64} & $\mathbf{82.64 \pm 0.45}$ & $\mathbf{83.32 \pm 0.60}$ & $\mathbf{83.14 \pm 0.57}$ & $\mathbf{83.12 \pm 0.64}$ \\
\cline{2-6}
& 128 & $82.61 \pm 0.48$ & $83.32 \pm 0.51$ & $83.17 \pm 0.46$ & $83.21 \pm 0.46$ \\
\hline
\multirow{3}{*}{PANNs CNN10}
& 48 & $78.42 \pm 0.57$ & $79.33 \pm 0.84$ & $79.20 \pm 0.53$ & $79.19 \pm 0.66$ \\
\cline{2-6}
& \textbf{64} & $\mathbf{81.18 \pm 0.32}$ & $\mathbf{81.88 \pm 0.54}$ & $\mathbf{81.85 \pm 0.36}$ & $\mathbf{81.79 \pm 0.47}$ \\
\cline{2-6}
& 128 & $81.13 \pm 0.48$ & $81.79 \pm 0.47$ & $81.75 \pm 0.34$ & $81.67 \pm 0.35$ \\
\hline
\multirow{3}{*}{PANNs CNN14}
& 48 & $80.27 \pm 0.64$ & $80.95 \pm 0.70$ & $80.80 \pm 0.66$ & $80.77 \pm 0.69$ \\
\cline{2-6}
& \textbf{64} & $\mathbf{82.53 \pm 0.48}$ & $\mathbf{83.22 \pm 0.41}$ & $\mathbf{83.19 \pm 0.59}$ & $\mathbf{83.14 \pm 0.54}$ \\
\cline{2-6}
& 128 & $82.21 \pm 0.58$ & $83.02 \pm 0.67$ & $82.97 \pm 0.81$ & $82.93 \pm 0.73$ \\
\hline
\end{tabular}
}
\end{table}

\subsection{Results on the multimodal framework}

\subsubsection{Performance of the proposed HMMF-Net}~\label{sec:performance_proposed}

\begin{figure}[H]
    \centering
    \includegraphics[width=\linewidth]{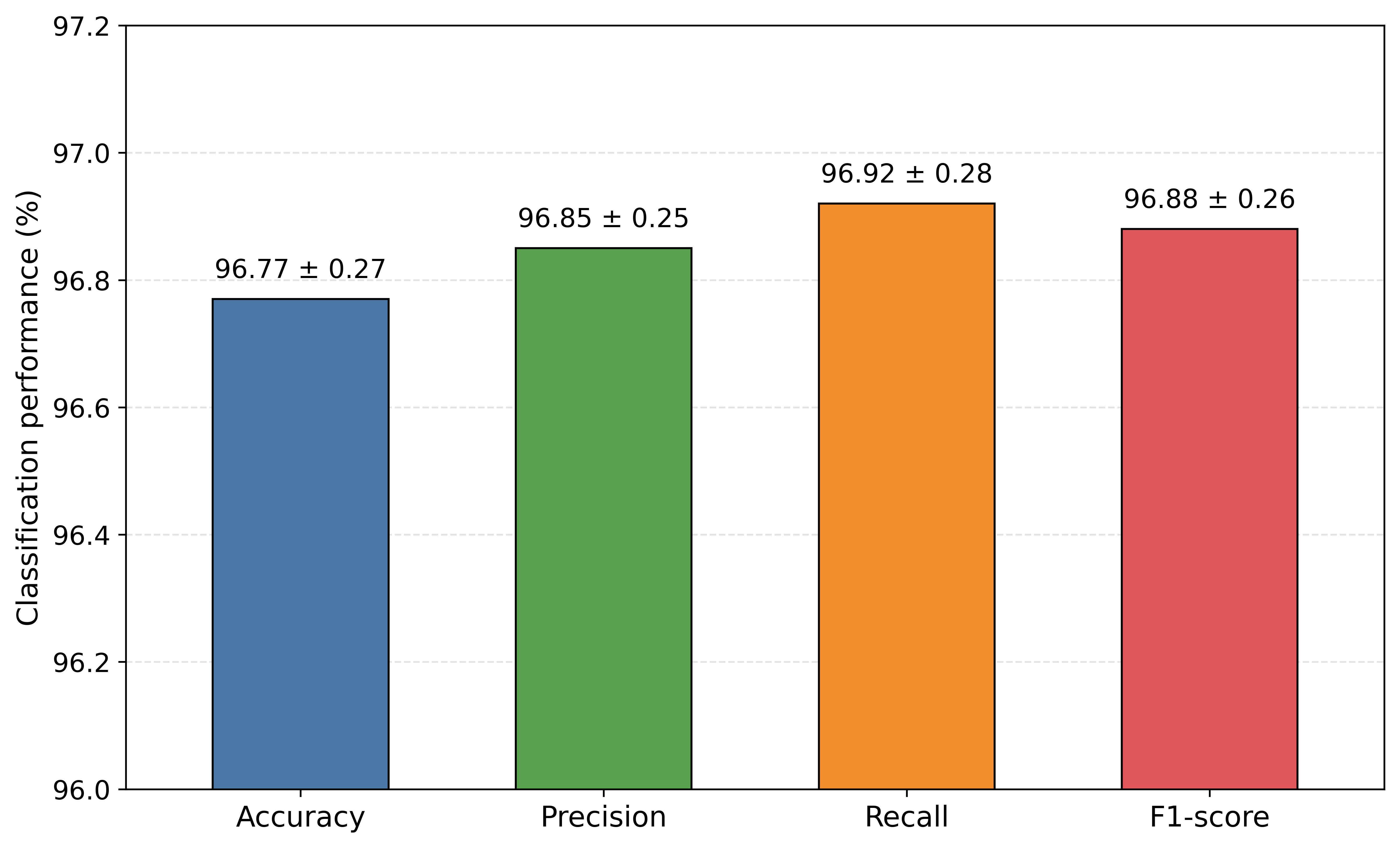}
  \caption{Classification performance (\%, mean $\pm$ std) of the proposed HMMF-Net on the held-out test folds.}
    \label{fig:proposed_results}
\end{figure}

Figure~\ref{fig:proposed_results} presents the classification performance of the proposed HMMF-Net on the held-out test folds. HMMF-Net achieves an accuracy of $96.77 \pm 0.27\%$, with precision, recall, and F1-score of $96.85 \pm 0.25\%$, $96.92 \pm 0.28\%$, and $96.88 \pm 0.26\%$, respectively. The four metrics remain closely aligned, with a maximum difference of only 0.15 percentage points, indicating consistent predictive performance across the evaluation criteria. The standard deviations also remain small, ranging from 0.25 to 0.28 percentage points, indicating limited variation across the held-out test folds.

Compared with the best performing multi-frame visual configuration, EfficientNetB0 with frames $(0,49)$, which achieves an accuracy of $95.49\%$, the proposed HMMF-Net improves the accuracy to $96.77\%$, corresponding to an absolute gain of 1.28 percentage points. More importantly, the classification error is reduced from 4.51\% to 3.23\%, corresponding to a 28.38\% relative reduction in the remaining classification error. This finding suggests that the acoustic modality provides complementary information beyond the temporal cues captured by multi-frame visual input, allowing the multimodal framework to recover a substantial portion of the residual errors of the visual-only configuration.

This improvement can be attributed to the complementary characteristics of the two sensing modalities. The visual branch employs early multi-frame encoding to capture temporally distributed visual cues within each feeding clip, whereas the AMARU-Net acoustic branch extracts compact spectro-temporal representations from feeding-related acoustic signals. These modality-specific representations are progressively integrated through the hierarchical BMCF modules, which enable bidirectional information exchange through learnable messenger tokens. Such integration is particularly relevant in aquaculture environments, where visual observations may be affected by water-surface agitation, splashes, bubbles, and fish occlusions, while acoustic observations may be influenced by background mechanical noise from aerators and water pumps. By combining these partially complementary observations, HMMF-Net can form a more robust representation of feeding activity under the evaluated conditions.

\subsubsection{Ablation study of hierarchical BMCF integration}
\label{sec:ablation_bmcf}

To validate the hierarchical design of BMCF, an ablation study was conducted by progressively adding BMCF modules from the deepest to the shallowest representation levels. The ablation starts with a feature-concatenation baseline without BMCF, followed by configurations that add BMCF at Stages 5, 4, 3, 2, and 1 in sequence. The configuration with BMCF at Stages 2--5 is adopted as the proposed setting, while the final configuration, which additionally incorporates Stage 1, examines whether cross-modal interaction at the shallowest representation level provides further benefit. Table~\ref{tab:ablation_fusion_stages} summarizes the classification performance, parameter count, and computational cost of the evaluated configurations on the validation set under five-fold cross-validation.

\begin{table}[H]
\caption{Ablation study of progressive BMCF integration from deep to shallow hierarchical representation levels on the validation set.}
\label{tab:ablation_fusion_stages}
\centering
\renewcommand{\arraystretch}{1.3}
\footnotesize
\resizebox{\linewidth}{!}{%
\begin{tabular}{|l|c|c|c|c|c|c|c|c|c|c|c|}
\hline
\multirow{2}{*}{\textbf{Configuration}} & \multicolumn{5}{c|}{\textbf{Hierarchical fusion stages (BMCF)}} & \multirow{2}{*}{\textbf{Accuracy (\%)}} & \multirow{2}{*}{\textbf{Precision (\%)}} & \multirow{2}{*}{\textbf{Recall (\%)}} & \multirow{2}{*}{\textbf{F1-score (\%)}} & \multirow{2}{*}{\textbf{Params (M)}} & \multirow{2}{*}{\textbf{GFLOPs}} \\ \cline{2-6}
 & \textbf{Stage 1} & \textbf{Stage 2} & \textbf{Stage 3} & \textbf{Stage 4} & \textbf{Stage 5} & & & & & & \\ \hline
Feature concat (baseline) &                  &                  &                  &                  &                  & 95.88 $\pm$ 0.22 & 95.98 $\pm$ 0.21 & 96.05 $\pm$ 0.18 & 96.01 $\pm$ 0.19 & 3.835 & 1.002 \\ \hline
+ Stage 5                 &                  &                  &                  &                  & \checkmark       & 96.01 $\pm$ 0.24 & 96.11 $\pm$ 0.21 & 96.20 $\pm$ 0.26 & 96.15 $\pm$ 0.24 & 3.895 & 1.002 \\ \hline
+ Stages 4--5             &                  &                  &                  & \checkmark       & \checkmark       & 96.20 $\pm$ 0.33 & 96.30 $\pm$ 0.33 & 96.37 $\pm$ 0.34 & 96.33 $\pm$ 0.33 & 3.927 & 1.002 \\ \hline
+ Stages 3--5             &                  &                  & \checkmark       & \checkmark       & \checkmark       & 96.40 $\pm$ 0.28 & 96.50 $\pm$ 0.26 & 96.55 $\pm$ 0.30 & 96.53 $\pm$ 0.28 & 3.950 & 1.002 \\ \hline
\textbf{+ Stages 2--5 (proposed)}  &                  & \checkmark       & \checkmark       & \checkmark       & \checkmark       & \textbf{96.81 $\pm$ 0.21} & \textbf{96.89 $\pm$ 0.19} & \textbf{96.95 $\pm$ 0.20} & \textbf{96.92 $\pm$ 0.20} & 3.987 & 1.002 \\ \hline
+ Stages 1--5       & \checkmark       & \checkmark       & \checkmark       & \checkmark       & \checkmark       & 96.57 $\pm$ 0.22 & 96.66 $\pm$ 0.22 & 96.72 $\pm$ 0.21 & 96.69 $\pm$ 0.21 & 3.990 & 1.002 \\ \hline
\end{tabular}%
}
\end{table}

The feature concatenation baseline combines the final visual and acoustic representations without intermediate cross-modal interaction, achieving an accuracy of $95.88 \pm 0.22\%$ and an F1-score of $96.01 \pm 0.19\%$. Adding BMCF progressively from Stage 5 to Stage 3 increases the accuracy to $96.01 \pm 0.24\%$, $96.20 \pm 0.33\%$, and $96.40 \pm 0.28\%$, respectively, with corresponding F1-scores of $96.15 \pm 0.24\%$, $96.33 \pm 0.33\%$, and $96.53 \pm 0.28\%$. These consistent gains indicate that intermediate cross-modal interaction contributes additional information beyond final-stage feature concatenation.

Adding Stage 2 produces the largest incremental gain and yields the proposed four-stage configuration (+ Stages 2--5), which achieves the highest performance with an accuracy of $96.81 \pm 0.21\%$, precision of $96.89 \pm 0.19\%$, recall of $96.95 \pm 0.20\%$, and F1-score of $96.92 \pm 0.20\%$. Compared with the feature-concatenation baseline, these results correspond to improvements of 0.93, 0.91, 0.90, and 0.91 percentage points, respectively. In particular, adding Stage 2 improves accuracy by 0.41 percentage points over the Stage 3--5 configuration, suggesting that this representation level provides particularly useful information for cross-modal interaction.

In contrast, adding Stage 1 does not provide further benefit. The five-stage configuration (+ Stages 1--5) reduces the accuracy and F1-score to $96.57 \pm 0.22\%$ and $96.69 \pm 0.21\%$, respectively, corresponding to decreases of 0.24 and 0.23 percentage points from the proposed configuration. This reduction may be associated with the less discriminative nature of very early representations, where high-resolution feature maps can contain more low-level and modality-specific details. Cross-modal interaction at this level may therefore provide less useful information for subsequent representation learning. Importantly, the reduction is not primarily attributable to computational cost, as adding Stage 1 increases the parameter count by only 0.003M.

From the computational perspective, the proposed four-stage configuration increases the parameter count from 3.835M to 3.987M, corresponding to an increase of 0.152M parameters or approximately 3.96\%. GFLOPs remain unchanged at 1.002 across all configurations. This low overhead is enabled by the BMCF design, which compresses modality-specific feature maps into four 32-dimensional messenger tokens before cross-modal attention. Overall, the results support BMCF integration at Stages 2--5, which provides the strongest predictive performance while maintaining a compact computational profile.

\subsubsection{Confusion matrix and class-wise analysis}
\label{sec:confusion_analysis}

Table~\ref{tab:confusion_class_wise} presents the confusion matrices and class-wise metrics of AMARU-Net, EfficientNetB0, and the proposed HMMF-Net on the validation set under five-fold cross-validation. The results reveal distinct class-level error patterns across the acoustic, visual, and multimodal configurations.

AMARU-Net exhibits substantial confusion between adjacent feeding intensities, particularly involving the Medium class. On average, 186.20 Medium samples are misclassified as Weak and 239.20 as Strong, while 171.80 Weak samples and 174.60 Strong samples are classified as Medium. This pattern is reflected in the low Medium-class recall and F1-score of $59.10\%$ and $61.24\%$, respectively. In contrast, EfficientNetB0 substantially reduces these cross-class errors, with Medium-to-Weak and Medium-to-Strong confusions decreasing to 39.80 and 52.60 samples, respectively. Its Medium-class F1-score consequently increases to $91.89\%$. These results indicate that the visual modality provides more discriminative information for distinguishing neighboring feeding intensities than the acoustic modality alone.

The proposed HMMF-Net further reduces the remaining class ambiguity. Medium-to-Weak and Medium-to-Strong confusions decrease to 26.20 and 

\begin{landscape}

\begin{table}[H]
\centering
\caption{Detailed confusion matrices and class-wise performance evaluation of representative unimodal baselines and the proposed multimodal framework on the validation set.}
\label{tab:confusion_class_wise}

\footnotesize
\renewcommand{\arraystretch}{1.25}
\setlength{\tabcolsep}{5pt}
\captionsetup[subtable]{font=footnotesize,skip=3pt}


\begin{subtable}[t]{0.485\linewidth}
\centering
\caption{AMARU-Net (acoustic): confusion matrix}
\label{tab:cm_cnn6}

\resizebox{\linewidth}{!}{%
\begin{tabular}{|l|c|c|c|c|}
\hline
\multirow{2}{*}{\textbf{True label}}
& \multicolumn{4}{c|}{\textbf{Predicted label}} \\
\cline{2-5}
& \textbf{None}
& \textbf{Weak}
& \textbf{Medium}
& \textbf{Strong} \\
\hline
\textbf{None}
& \textbf{731.40 $\pm$ 5.03}
& 26.00 $\pm$ 6.56
& 9.20 $\pm$ 3.70
& 12.40 $\pm$ 5.98 \\
\hline
\textbf{Weak}
& 18.40 $\pm$ 3.51
& \textbf{1062.00 $\pm$ 39.15}
& 171.80 $\pm$ 41.95
& 24.80 $\pm$ 3.35 \\
\hline
\textbf{Medium}
& 9.00 $\pm$ 2.00
& 186.20 $\pm$ 42.54
& \textbf{627.60 $\pm$ 71.92}
& 239.20 $\pm$ 37.51 \\
\hline
\textbf{Strong}
& 19.60 $\pm$ 5.32
& 19.20 $\pm$ 6.98
& 174.60 $\pm$ 46.62
& \textbf{998.60 $\pm$ 40.99} \\
\hline
\end{tabular}%
}

\end{subtable}
\hfill
\begin{subtable}[t]{0.485\linewidth}
\centering
\caption{AMARU-Net (acoustic): class-wise metrics}
\label{tab:metrics_cnn6}

\resizebox{\linewidth}{!}{%
\begin{tabular}{|l|c|c|c|c|}
\hline
\textbf{Class}
& \makecell{\textbf{Precision}\\\textbf{(\%)}}
& \makecell{\textbf{Recall}\\\textbf{(\%)}}
& \makecell{\textbf{F1-score}\\\textbf{(\%)}}
& \textbf{Support} \\
\hline
None
& 93.96 $\pm$ 0.82
& 93.89 $\pm$ 0.65
& 93.93 $\pm$ 0.70
& 779 \\
\hline
Weak
& 82.24 $\pm$ 2.48
& 83.16 $\pm$ 3.07
& 82.63 $\pm$ 0.79
& 1277 \\
\hline
Medium
& 64.26 $\pm$ 3.28
& 59.10 $\pm$ 6.77
& 61.24 $\pm$ 2.26
& 1062 \\
\hline
Strong
& 78.43 $\pm$ 2.13
& 82.39 $\pm$ 3.38
& 80.29 $\pm$ 0.76
& 1212 \\
\hline
\end{tabular}%
}

\end{subtable}

\vspace{0.8em}


\begin{subtable}[t]{0.485\linewidth}
\centering
\caption{EfficientNetB0 (early 2-frame visual): confusion matrix}
\label{tab:cm_eff}

\resizebox{\linewidth}{!}{%
\begin{tabular}{|l|c|c|c|c|}
\hline
\multirow{2}{*}{\textbf{True label}}
& \multicolumn{4}{c|}{\textbf{Predicted label}} \\
\cline{2-5}
& \textbf{None}
& \textbf{Weak}
& \textbf{Medium}
& \textbf{Strong} \\
\hline
\textbf{None}
& \textbf{769.80 $\pm$ 4.87}
& 0.20 $\pm$ 0.45
& 0.80 $\pm$ 0.45
& 8.20 $\pm$ 4.87 \\
\hline
\textbf{Weak}
& 0.40 $\pm$ 0.55
& \textbf{1235.80 $\pm$ 5.81}
& 40.80 $\pm$ 5.45
& 0.00 $\pm$ 0.00 \\
\hline
\textbf{Medium}
& 0.40 $\pm$ 0.55
& 39.80 $\pm$ 5.36
& \textbf{969.20 $\pm$ 17.57}
& 52.60 $\pm$ 18.99 \\
\hline
\textbf{Strong}
& 15.00 $\pm$ 1.41
& 0.60 $\pm$ 0.55
& 36.60 $\pm$ 8.08
& \textbf{1159.80 $\pm$ 7.92} \\
\hline
\end{tabular}%
}

\end{subtable}
\hfill
\begin{subtable}[t]{0.485\linewidth}
\centering
\caption{EfficientNetB0 (early 2-frame visual): class-wise metrics}
\label{tab:metrics_eff}

\resizebox{\linewidth}{!}{%
\begin{tabular}{|l|c|c|c|c|}
\hline
\textbf{Class}
& \makecell{\textbf{Precision}\\\textbf{(\%)}}
& \makecell{\textbf{Recall}\\\textbf{(\%)}}
& \makecell{\textbf{F1-score}\\\textbf{(\%)}}
& \textbf{Support} \\
\hline
None
& 97.99 $\pm$ 0.20
& 98.82 $\pm$ 0.62
& 98.40 $\pm$ 0.24
& 779 \\
\hline
Weak
& 96.82 $\pm$ 0.35
& 96.77 $\pm$ 0.45
& 96.80 $\pm$ 0.26
& 1277 \\
\hline
Medium
& 92.54 $\pm$ 0.34
& 91.26 $\pm$ 1.65
& 91.89 $\pm$ 0.81
& 1062 \\
\hline
Strong
& 95.04 $\pm$ 1.30
& 95.69 $\pm$ 0.65
& 95.36 $\pm$ 0.35
& 1212 \\
\hline
\end{tabular}%
}

\end{subtable}

\vspace{0.8em}


\begin{subtable}[t]{0.485\linewidth}
\centering
\caption{Proposed multimodal framework: confusion matrix}
\label{tab:cm_proposed}

\resizebox{\linewidth}{!}{%
\begin{tabular}{|l|c|c|c|c|}
\hline
\multirow{2}{*}{\textbf{True label}}
& \multicolumn{4}{c|}{\textbf{Predicted label}} \\
\cline{2-5}
& \textbf{None}
& \textbf{Weak}
& \textbf{Medium}
& \textbf{Strong} \\
\hline
\textbf{None}
& \textbf{773.40 $\pm$ 2.51}
& 0.00 $\pm$ 0.00
& 0.40 $\pm$ 0.55
& 5.20 $\pm$ 2.05 \\
\hline
\textbf{Weak}
& 0.40 $\pm$ 0.55
& \textbf{1246.00 $\pm$ 7.14}
& 30.40 $\pm$ 6.80
& 0.20 $\pm$ 0.45 \\
\hline
\textbf{Medium}
& 0.00 $\pm$ 0.00
& 26.20 $\pm$ 1.10
& \textbf{1003.20 $\pm$ 5.36}
& 32.60 $\pm$ 5.68 \\
\hline
\textbf{Strong}
& 9.60 $\pm$ 2.07
& 0.40 $\pm$ 0.55
& 32.60 $\pm$ 6.50
& \textbf{1169.40 $\pm$ 6.15} \\
\hline
\end{tabular}%
}

\end{subtable}
\hfill
\begin{subtable}[t]{0.485\linewidth}
\centering
\caption{Proposed multimodal framework: class-wise metrics}
\label{tab:metrics_proposed}

\resizebox{\linewidth}{!}{%
\begin{tabular}{|l|c|c|c|c|}
\hline
\textbf{Class}
& \makecell{\textbf{Precision}\\\textbf{(\%)}}
& \makecell{\textbf{Recall}\\\textbf{(\%)}}
& \makecell{\textbf{F1-score}\\\textbf{(\%)}}
& \textbf{Support} \\
\hline
None
& 98.72 $\pm$ 0.25
& 99.28 $\pm$ 0.32
& 99.00 $\pm$ 0.21
& 779 \\
\hline
Weak
& 97.91 $\pm$ 0.11
& 97.57 $\pm$ 0.56
& 97.74 $\pm$ 0.28
& 1277 \\
\hline
Medium
& \textbf{94.07 $\pm$ 0.97}
& \textbf{94.46 $\pm$ 0.50}
& \textbf{94.26 $\pm$ 0.42}
& 1062 \\
\hline
Strong
& 96.85 $\pm$ 0.46
& 96.49 $\pm$ 0.51
& 96.67 $\pm$ 0.35
& 1212 \\
\hline
\end{tabular}%
}

\end{subtable}

\end{table}

\end{landscape}

\noindent 32.60 samples, respectively, while Weak-to-Medium and Strong-to-Medium errors are reduced to 30.40 and 32.60 samples. Accordingly, the Medium-class F1-score increases from $91.89\%$ with EfficientNetB0 to $94.26\%$, with precision and recall reaching $94.07\%$ and $94.46\%$, respectively. The improvement is particularly relevant because Medium remains the most challenging feeding state among the evaluated classes.

\begin{figure}[H]
\centering

\begin{subfigure}[t]{0.95\linewidth}
    \centering
    \includegraphics[width=0.48\linewidth]{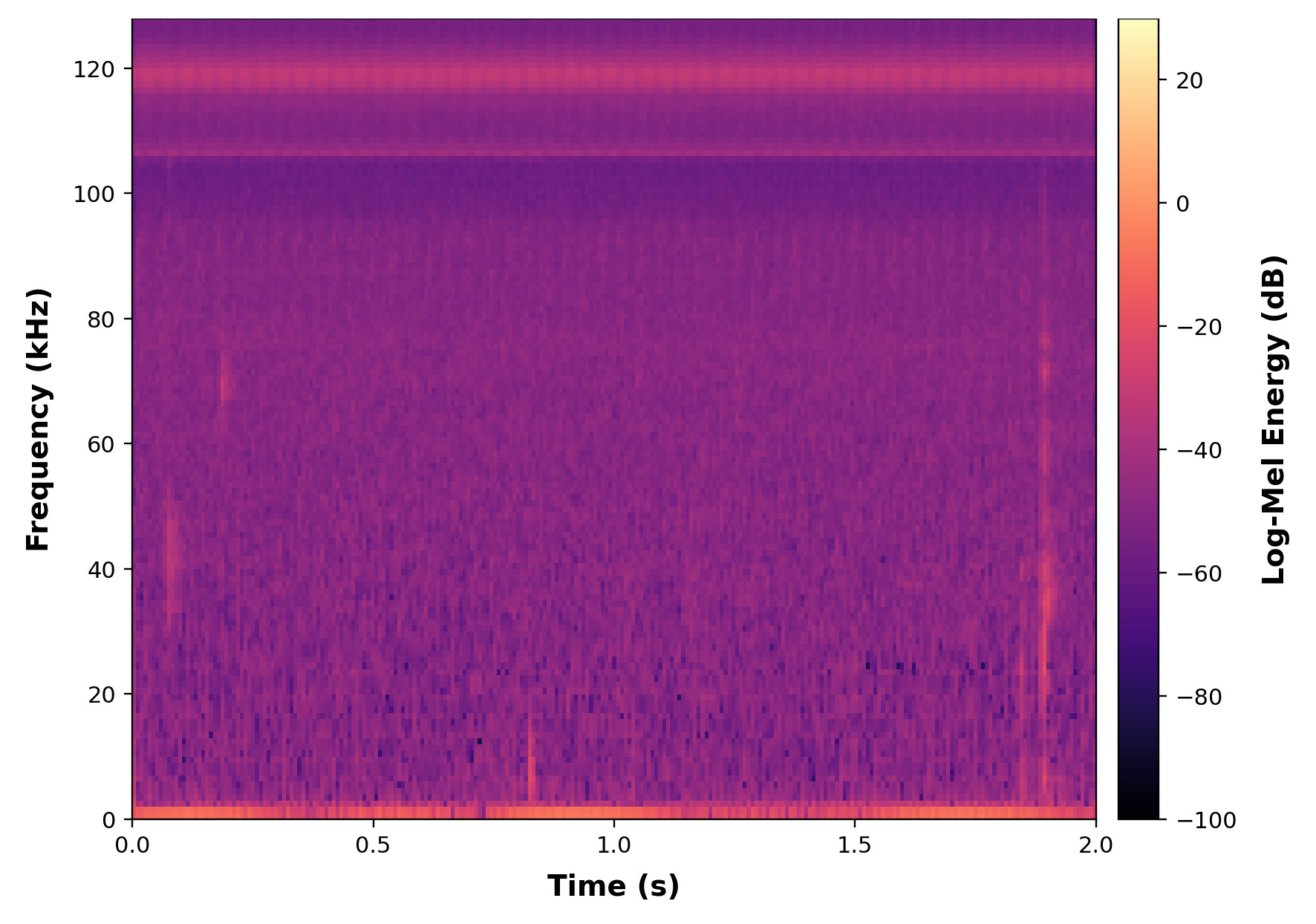}
    \caption{Mel-spectrogram input to AMARU-Net: prediction = Strong, ground truth = Strong.}
    \label{fig:case_mel}
\end{subfigure}

\vspace{0.8em}

\begin{subfigure}[t]{0.95\linewidth}
    \centering
    \begin{minipage}{\linewidth}
        \centering
        \begin{subfigure}[t]{0.48\linewidth}
            \centering
            \includegraphics[width=\linewidth]{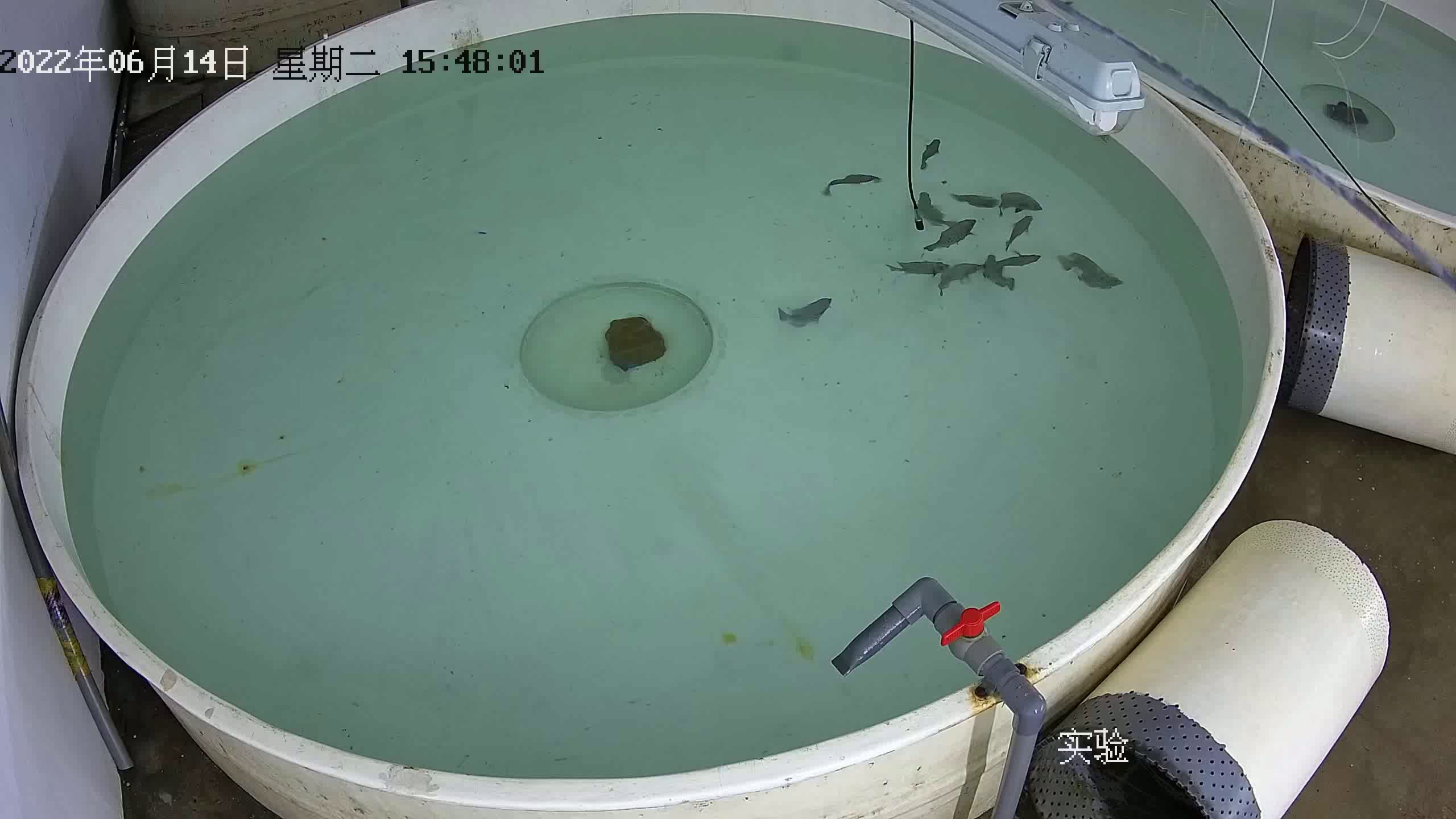}
            \caption{First frame}
            \label{fig:case_frame_first}
        \end{subfigure}
        \hfill
        \begin{subfigure}[t]{0.48\linewidth}
            \centering
            \includegraphics[width=\linewidth]{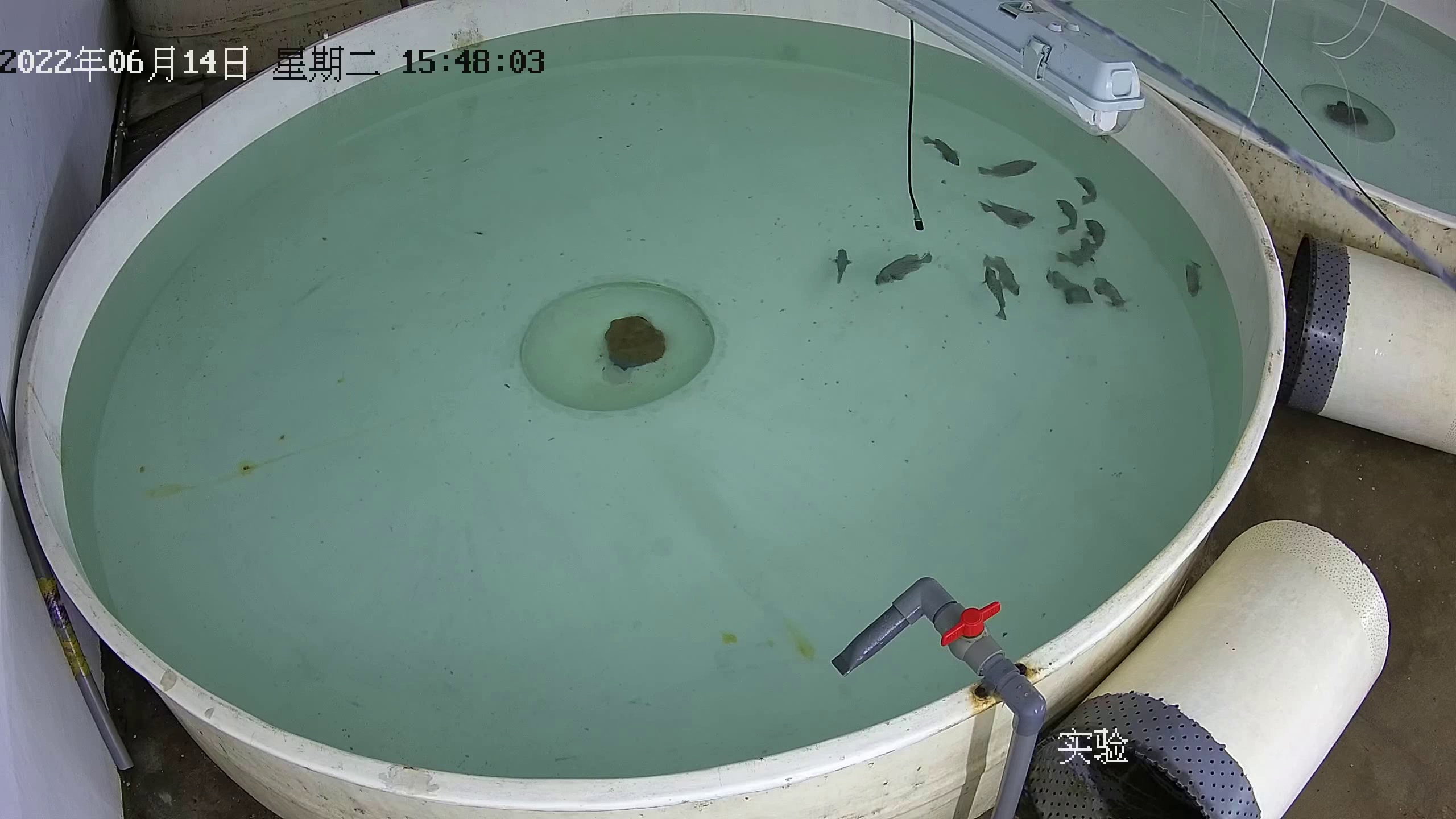}
            \caption{Last frame}
            \label{fig:case_frame_last}
        \end{subfigure}
    \end{minipage}

    \caption{ First and last frames input to EfficientNetB0: prediction = Medium, ground truth = Strong}
    \label{fig:case_visual}
\end{subfigure}

\caption{A representative held-out test sample: the acoustic model predicts Strong correctly, whereas the visual model predicts Medium.}
\label{fig:complementary_modalities}
\end{figure}

For the other classes, HMMF-Net maintains strong class-level discrimination, achieving F1-scores of $99.00\%$, $97.74\%$, and $96.67\%$ for None, Weak, and Strong, respectively. No None samples are misclassified as Weak, and no Medium samples are misclassified as None. A consistent trend is also observed in balanced accuracy, with HMMF-Net achieving $96.95 \pm 0.20\%$, compared with $95.64 \pm 0.35\%$ for EfficientNetB0. The remaining errors are concentrated mainly between adjacent feeding intensities, particularly at the Weak--Medium and Medium--Strong boundaries. The further reduction of these confusions with HMMF-Net indicates that combining acoustic and visual information provides complementary cues for distinguishing neighboring feeding states. Figure~\ref{fig:complementary_modalities}  illustrates how the two modalities complement each other using a representative held-out test sample, where the acoustic model predicts correctly, the visual model misclassifies the sample, and HMMF-Net produces the correct feeding state through multimodal fusion.

\subsubsection{Learning curves and training stability analysis}

\begin{figure}[H]
\centering

\begin{subfigure}{\linewidth}
    \centering
    \includegraphics[width=0.95\linewidth]{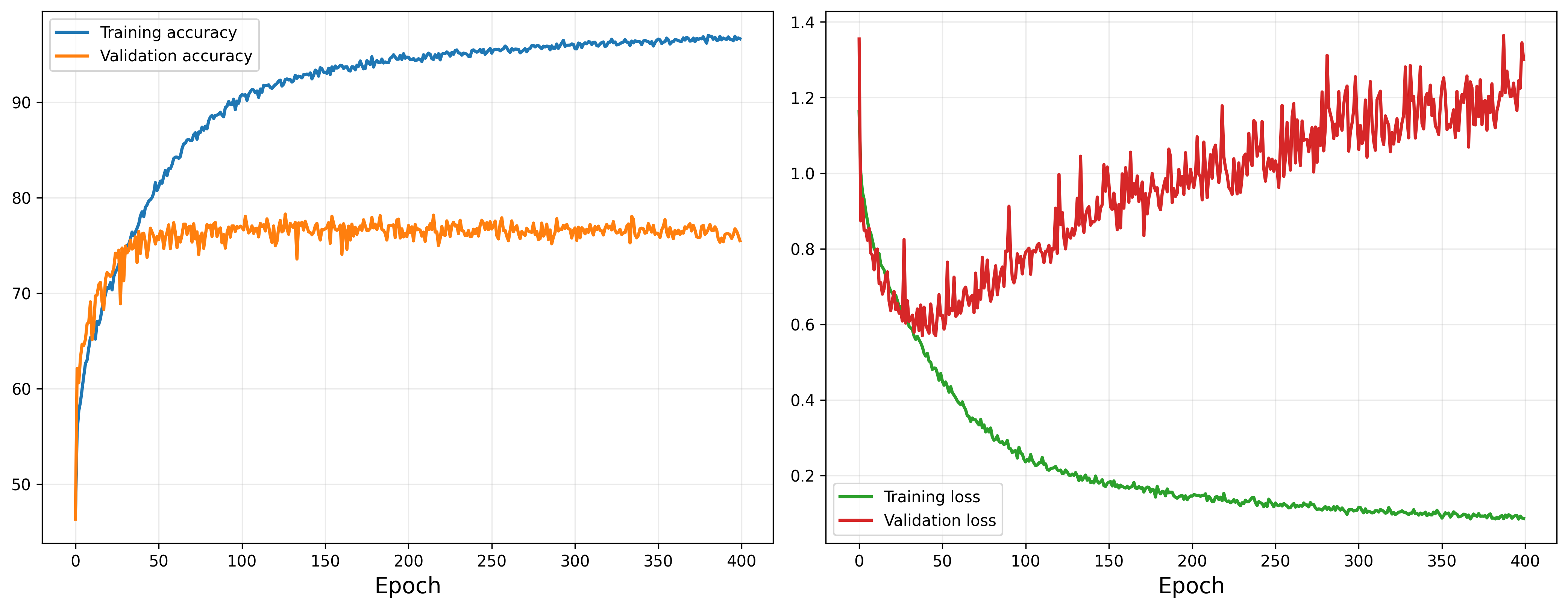}
    \caption{AMARU-Net (acoustic)}
    \label{fig:lc_cnn6}
\end{subfigure}

\vspace{0.2cm}

\begin{subfigure}{\linewidth}
    \centering
    \includegraphics[width=0.95\linewidth]{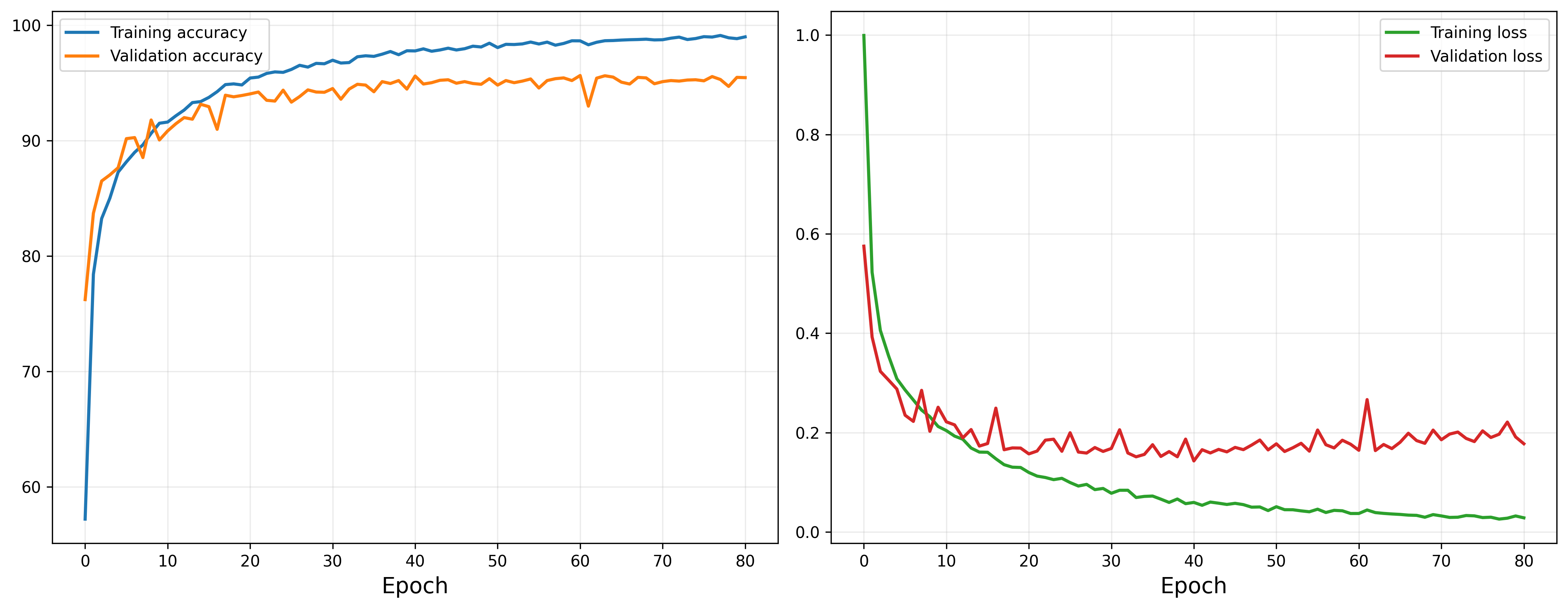}
    \caption{EfficientNetB0 (early 2-frame visual)}
    \label{fig:lc_eff}
\end{subfigure}

\vspace{0.2cm}

\begin{subfigure}{\linewidth}
    \centering
    \includegraphics[width=0.95\linewidth]{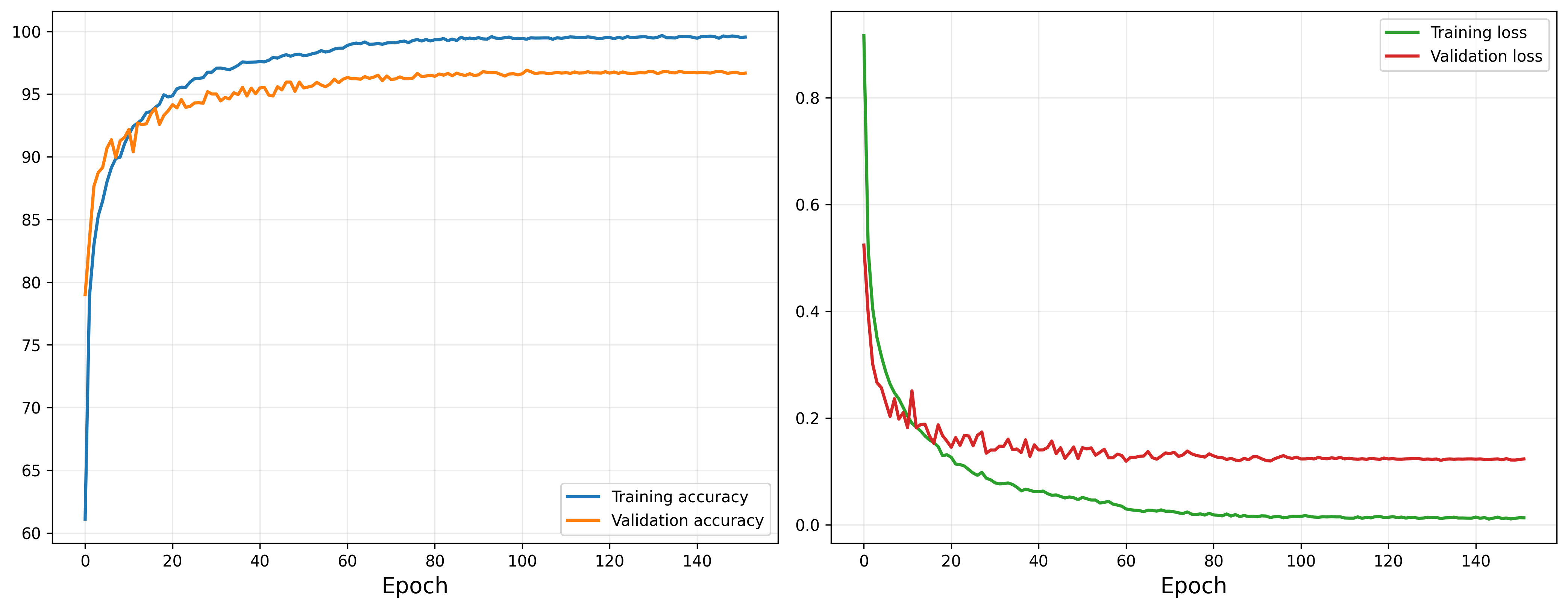}
    \caption{Proposed multimodal framework}
    \label{fig:lc_multimodal}
\end{subfigure}

\caption{Training and validation accuracy and loss curves of representative unimodal baselines and the proposed multimodal framework across training epochs.}
\label{fig:learning_curves_comparison}
\end{figure}

Figure~\ref{fig:learning_curves_comparison} illustrates the training and validation accuracy and loss trajectories for a representative cross-validation fold (Fold 1), providing insight into convergence behavior, optimization stability, and generalization.

The acoustic baseline exhibits the slowest convergence and the clearest signs of overfitting. Although the training accuracy continues to increase throughout the 400-epoch training process, the validation accuracy saturates at a substantially lower level and remains relatively fluctuating. Similarly, the validation loss decreases rapidly during the early stage, reaches a minimum at approximately epochs 30--50, and subsequently increases despite the continued decrease in training loss. This divergence between the training and validation trajectories indicates progressively deteriorating generalization during prolonged optimization. The slower convergence should be considered in the context of AMARU-Net being a task-specific architecture trained from scratch, without pretrained initialization.

In contrast, EfficientNetB0 with early two-frame visual encoding converges much more rapidly, exceeding 90\% accuracy within the first 10 epochs and reaching a peak validation accuracy of 95.64\%. This rapid convergence is partly supported by the ImageNet-pretrained initialization of the visual backbone. The validation loss also decreases substantially during the early optimization stage and subsequently remains relatively stable, although several transient fluctuations are observed, indicating some variation in validation behavior during training.

The proposed multimodal framework combines rapid convergence with comparatively stable optimization. It reaches a peak validation accuracy of 96.91\% at epoch 101, while the validation loss decreases smoothly and stabilizes at approximately 0.12. Unlike the acoustic baseline, the multimodal model does not exhibit a progressive increase in validation loss, while its validation trajectory is also smoother than that of the visual baseline. The rapid initial convergence is partly supported by the pretrained visual presentation, whereas the more stable validation behavior is consistent with the complementary acoustic--visual information integrated through the hierarchical

\noindent BMCF architecture.

The learning curves in the representative fold show that the multimodal framework converges rapidly while maintaining a relatively stable validation trajectory throughout training. Although the convergence behavior is influenced by both initialization and architecture, the smoother validation trajectory of the multimodal model suggests that complementary acoustic and visual cues can reduce reliance on modality-specific information and support more stable optimization.

\subsubsection{Comparison of multimodal fusion strategies}
\label{sec:multimodal_fusion_comparison}
Table~\ref{tab:multimodal_fusion_strategies} compares different multimodal fusion strategies on the held-out test folds using the same two-frame visual input (frames 0 and 49). Intermediate feature fusion consistently outperforms late probability averaging. For example, with PANNs CNN6 and EfficientNetB0, the accuracy increases from $94.96 \pm 0.58\%$ to $96.07 \pm 0.12\%$, corresponding to a gain of 1.11 percentage points. Similar improvements are observed with ConvNeXt-Tiny and Swin-Tiny, indicating that feature-level integration provides more effective multimodal interaction than probability-level averaging in the evaluated configurations.

\begin{table}[H]
\centering
\caption{Comparison of multimodal fusion strategies using the same two frames as input (frames 0 and 49) in terms of classification accuracy (\%, mean $\pm$ std) and computational cost on the held-out test folds.}
\label{tab:multimodal_fusion_strategies}
\footnotesize
\renewcommand{\arraystretch}{1.3}
\setlength{\tabcolsep}{5pt}

\resizebox{\textwidth}{!}{
\begin{tabular}{|l|l|l|l|c|c|c|}
\hline
\textbf{Fusion strategy}
& \textbf{Fusion method}
& \textbf{Acoustic model}
& \textbf{Visual model}
& \textbf{Accuracy}
& \textbf{Params (M)}
& \textbf{GFLOPs} \\
\hline

\multirow{4}{*}{Late fusion}
& \multirow{4}{*}{Probability averaging}
& \multirow{3}{*}{PANNs CNN6}
& EfficientNetB0
& $94.96 \pm 0.58$
& 8.583
& 5.951 \\
\cline{4-7}

& &
& ConvNeXt-Tiny
& $94.75 \pm 0.62$
& 32.397
& 14.085 \\
\cline{4-7}

& &
& Swin-Tiny
& $94.35 \pm 0.68$
& 32.096
& 11.085 \\
\cline{3-7}

& &
AMARU-Net
& EfficientNetB0
& $94.02 \pm 0.63$
& 4.252
& 1.042 \\
\hline

\multirow{4}{*}{Intermediate fusion}
& \multirow{4}{*}{Feature concatenation}
& \multirow{3}{*}{PANNs CNN6}
& EfficientNetB0
& $96.07 \pm 0.12$
& 9.504
& 5.953 \\
\cline{4-7}

& &
& ConvNeXt-Tiny
& $95.98 \pm 0.16$
& 33.056
& 14.059 \\
\cline{4-7}

& &
& Swin-Tiny
& $95.99 \pm 0.20$
& 32.755
& 11.087 \\
\cline{3-7}

& &
AMARU-Net
& EfficientNetB0
& $95.97 \pm 0.33$
& 3.835
& 1.002 \\
\hline

\textbf{Proposed HMMF-Net}
& \textbf{Hierarchical BMCF fusion}
& \textbf{AMARU-Net}
& \textbf{EfficientNetB0}
& \textbf{96.77 $\pm$ 0.27}
& \textbf{3.987}
& \textbf{1.002} \\
\hline

\end{tabular}
}
\end{table}

The use of AMARU-Net substantially reduces the computational cost while maintaining competitive multimodal performance. With EfficientNetB0 and feature concatenation, AMARU-Net achieves $95.97 \pm 0.33\%$ accuracy with only 3.835M parameters and 1.002 GFLOPs, compared with 9.504M parameters and 5.953 GFLOPs for the corresponding PANNs CNN6 configuration. This corresponds to reductions of approximately 59.6\% in parameters and 83.2\% in GFLOPs.

More importantly, HMMF-Net achieves $96.77 \pm 0.27\%$ accuracy, improving over conventional feature concatenation with the same AMARU-Net and EfficientNetB0 branches by 0.80 percentage points. This improvement requires only 0.152M additional parameters (3.96\%), while the computational cost remains unchanged at 1.002 GFLOPs. Compared with the strongest late-fusion configuration, HMMF-Net also provides a 1.81 pp accuracy improvement while using substantially fewer parameters and GFLOPs. These results indicate that hierarchical cross-modal interaction through BMCF provides additional predictive benefit beyond conventional feature concatenation while preserving the compact computational profile of the proposed framework.

\subsection{Edge deployment}

The practical feasibility of the proposed framework was further investigated through deployment on an NVIDIA Jetson Orin Nano embedded computing platform. The complete system was executed using TensorRT with FP32 precision, and the on-device evaluation was conducted across the five cross-validation folds using the corresponding held-out test sets. The evaluation was designed to examine both predictive performance after deployment and the computational efficiency of the multimodal processing pipeline on the edge device.

First, the predictive performance of the five fold-specific models was evaluated on their corresponding held-out test sets. Table~\ref{tab:test_set_performance} summarizes the resulting classification performance after deployment on the Jetson Orin Nano.

\begin{table}[H]
\centering
\caption{Predictive performance of the HMMF-Net on the corresponding held-out test sets after deployment on NVIDIA Jetson Orin Nano.}
\label{tab:test_set_performance}
\footnotesize
\renewcommand{\arraystretch}{1.2}
\setlength{\tabcolsep}{8pt}
\resizebox{0.65\textwidth}{!}{
\begin{tabular}{|c|c|c|c|}
\hline
\textbf{Accuracy} &
\textbf{Precision} &
\textbf{Recall} &
\textbf{F1-score} \\ \hline

$96.76 \pm 0.29$ &
$96.84 \pm 0.26$ &
$96.91 \pm 0.31$ &
$96.87 \pm 0.28$ \\ \hline

\end{tabular}
}
\end{table}

The deployed framework achieves an accuracy of $96.76 \pm 0.29\%$, precision of $96.84 \pm 0.26\%$, recall of $96.91 \pm 0.31\%$, and F1-score of $96.87 \pm 0.28\%$. These results are highly consistent with the corresponding offline results reported in Section~\ref{sec:performance_proposed}, where the framework achieved an accuracy of $96.77 \pm 0.27\%$ and an F1-score of $96.88 \pm 0.26\%$. The differences in accuracy and F1-score are only 0.01 percentage points, indicating that the deployment process does not result in a noticeable degradation in predictive performance. The deployed model has a size of 18.81 MB, indicating a relatively compact model footprint for embedded execution.

The computational cost of the deployed pipeline was then evaluated across the five test folds on the Jetson Orin Nano. For each fold, the corresponding test samples were organized into continuous audio-video sequences of approximately 10 minutes and loaded onto the device. The original samples remained 2-second audio-video segments, and their concatenation was used to provide a continuous input sequence for on-device processing. For each 2-second sample, GPU-based preprocessing included video frame extraction, audio extraction, resampling to 64 kHz, short-time Fourier transform (STFT), mel-spectrogram computation, normalization, and tensor conversion. The resulting multimodal tensors were then processed by the TensorRT inference engine on the GPU.

\begin{table}[H]
\centering
\caption{Processing time and model size of the HMMF-Net framework on NVIDIA Jetson Orin Nano.}
\label{tab:deployment_performance}
\footnotesize
\renewcommand{\arraystretch}{1.2}
\setlength{\tabcolsep}{8pt}
\resizebox{0.7\textwidth}{!}{
\begin{tabular}{|c|c|c|}
\hline
\textbf{Data preprocessing (ms)} &
\textbf{Inference (ms)} &
\textbf{Model size (MB)} \\ \hline

$7.78 \pm 0.27$ &
$11.54 \pm 0.36$ &
18.81 \\ \hline

\end{tabular}
}
\end{table}

As shown in Table~\ref{tab:deployment_performance}, the GPU-based preprocessing stage requires $7.78 \pm 0.27$ ms per 2-second audio-video sample, while TensorRT inference requires $11.54 \pm 0.36$ ms. The combined preprocessing and inference time is therefore approximately 19.32 ms per sample. This processing time is substantially shorter than the 2-second observation interval, corresponding to less than 1\% of the duration of an individual input sample. The result indicates that the embedded platform has sufficient computational throughput to process consecutive 2-second multimodal samples without the processing time becoming the limiting factor relative to the observation interval.

\begin{figure}[H]
\centering
\includegraphics[width=\textwidth]{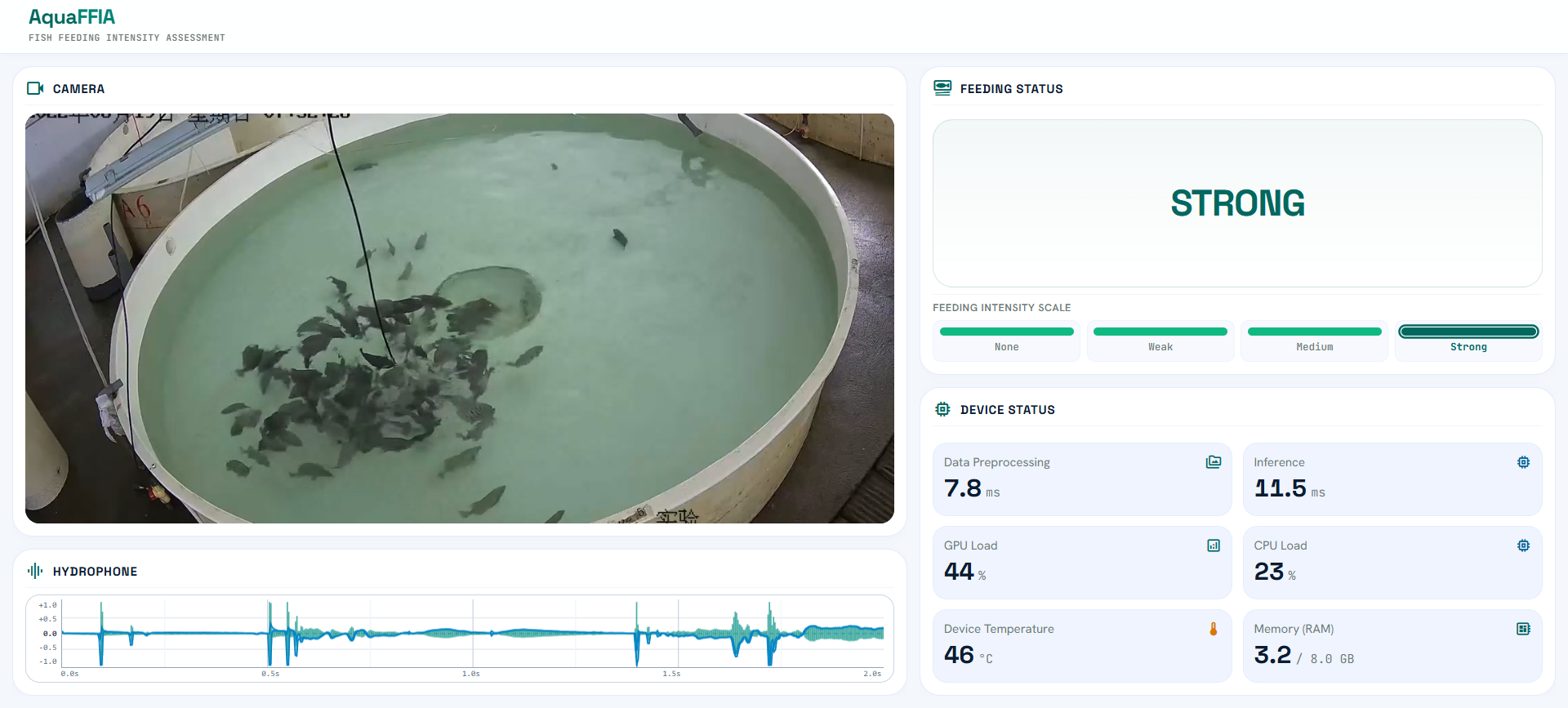}
\caption{User interface of the on-device fish feeding intensity assessment system deployed on NVIDIA Jetson Orin Nano.}
\label{fig:on_device_interface}
\end{figure}

For practical operation, a user interface was implemented on the Jetson Orin Nano, as illustrated in Figure~\ref{fig:on_device_interface}. The interface displays the synchronized video stream and acoustic waveform together with runtime information, including data preprocessing and inference time. It also presents the predicted feeding intensity category using a feeding intensity scale, with the current prediction highlighted in the interface. These components provide an integrated view of the multimodal input, prediction output, and runtime status during on-device operation.

\subsection{Comparison with existing FFIA methods}

Table~\ref{tab:methods_comparison} compares HMMF-Net with other multimodal AV-FFIA methods in terms of performance, computational complexity, evaluation protocols, and deployment.

\cite{Cui2025} introduced U-FFIA, a unified mixed-modality Transformer framework, achieving an accuracy of 90.70\% with 21.626M parameters and 15.488 GFLOPs.~\cite{Li2025} subsequently proposed ACAF-Net, which combines low-rank bilinear pooling with adaptive attention and achieved 95.40\% accuracy with 11.80M parameters. Both approaches employ comparatively large model architectures in terms of parameter count. U-FFIA additionally requires 15.488 GFLOPs, whereas the computational complexity of ACAF-Net was not reported.

More recent studies, including LDMF~\citep{Sha2026} and HMKD~\citep{Du2026}, achieved accuracies of 98.22\% and 97.47 $\pm$ 0.61\%, respectively. However, these results were obtained using filtered subsets of the AV-FFIA dataset containing 6,016 and 26,843 samples, respectively, whereas HMMF-Net uses the full dataset. Therefore, the accuracy values should be interpreted in the context of their different dataset scopes. In addition, LDMF and HMKD adopted holdout-based evaluation, while HMMF-Net was evaluated using five-fold cross-validation, with performance reported as the mean and standard deviation across the held-out test folds.

Among the methods evaluated on the full dataset, HMMF-Net achieves the highest accuracy, reaching $96.77 \pm 0.27\%$, which corresponds to improvements of 6.07 and 1.37 pp over U-FFIA and ACAF-Net, respectively. The proposed framework also has the smallest parameter count among all compared methods, with only 3.987M parameters. This represents reductions of approximately 81.56\%, 66.21\%, 43.85\%, and 37.01\% compared with U-FFIA, ACAF-Net, LDMF, and HMKD, respectively. In terms of computational complexity, HMMF-Net requires only 1.002 GFLOPs, corresponding to reductions of approximately 93.53\%, 85.26\%, and 91.99\% compared with U-FFIA, LDMF, and HMKD, respectively. These results indicate that competitive predictive performance can be achieved with substantially lower computational complexity.

The efficiency of the proposed framework is supported by its architectural design. Early two-frame visual encoding captures temporally distributed visual information without requiring computationally intensive 3D convolution, while AMARU-Net extracts compact spectro-temporal acoustic representations. The hierarchical BMCF module then enables bidirectional cross-modal interaction through learnable messenger tokens across multiple representation levels. These design choices allow the model to combine complementary acoustic and visual information while maintaining a small computational footprint. Furthermore, it was deployed on an NVIDIA Jetson Orin Nano, demonstrating its practical suitability for resource-constrained edge-based aquaculture monitoring.

\begin{table}[H]
\centering
\caption{Performance and computational complexity comparison of the proposed multimodal framework with existing methods on the AV-FFIA dataset.}
\label{tab:methods_comparison}
\footnotesize
\renewcommand{\arraystretch}{1.3}
\setlength{\tabcolsep}{8pt}
\resizebox{\textwidth}{!}{
\begin{tabular}{|l|c|c|c|c|c|c|}
\hline
\textbf{Method}
& \textbf{Accuracy (\%)}
& \textbf{Params (M)}
& \textbf{GFLOPs}
& \textbf{Dataset scope}
& \textbf{Evaluation protocol}
& \textbf{Deployment} \\
\hline

U-FFIA~\citep{Cui2025}
& 90.70
& 21.626
& 15.488
& Full dataset
& Holdout (8:1:1, 1 run)
& NR \\
\hline

ACAF-Net~\citep{Li2025}
& 95.40
& 11.80
& NR
& Full dataset
& Holdout (8:1:1, 5 runs)
& NR \\
\hline

LDMF~\citep{Sha2026}
& 98.22
& 7.10
& 6.80
& 6,016 samples (Filtered)
& Holdout (8:1:1, 1 run)
& NR \\
\hline

HMKD~\citep{Du2026}
& 97.47 $\pm$ 0.61
& 6.33
& 12.51
& 26,843 samples (Filtered)
& Holdout (8:1:1, 3 runs)
& NR \\
\hline

\textbf{Proposed framework}
& \textbf{96.77 $\pm$ 0.27}
& \textbf{3.987}
& \textbf{1.002}
& \textbf{Full dataset}
& \textbf{Cross-validation (5-fold)}
& \textbf{Jetson Orin Nano} \\
\hline

\end{tabular}
}
\vspace{0.5mm}

\begin{minipage}{\textwidth}
\scriptsize
\textit{Note:} NR denotes values not reported in the original study.
\end{minipage}

\end{table}

\section{Conclusion}\label{sec:conclusion}

This study proposed HMMF-Net, a lightweight multimodal framework for accurate and computationally efficient fish feeding intensity assessment. HMMF-Net integrates early multi-frame visual encoding with an adapted 2D EfficientNetB0 backbone, an adaptive multi-axis residual acoustic branch (AMARU), and hierarchical bidirectional messenger cross-modal fusion (BMCF). On the AV-FFIA benchmark, HMMF-Net achieved 96.77 $\pm$ 0.27\% accuracy on the held-out test folds and reduced the classification error by 28.38\% relative to the best-performing visual configuration. Deployment on an NVIDIA Jetson Orin Nano using TensorRT with FP32 precision maintained 96.76 $\pm$ 0.29\% accuracy with a processing time of 19.32~ms per 2-second sample and a model size of 18.81~MB, demonstrating its potential for practical edge-based FFIA.
These results demonstrate the effectiveness of HMMF-Net for accurate and efficient edge-based FFIA. However, the current evaluation was conducted on a single fish species within a controlled recirculating aquaculture facility, which may limit generalization to other species and more variable farming environments. In addition, the fixed two-frame temporal sampling may not fully capture prolonged or irregular feeding dynamics. Future work will therefore investigate cross-species generalization, validation in open-water sea cages under non-stationary acoustic and visual conditions, adaptive temporal frame selection, and closed-loop integration with automated feeding actuators for precision aquaculture management.

\section*{Funding:} This research received no funding.


\section*{Author contribution:}

\textbf{Phi-Hung Hoang}: Writing - review \& editing, Writing - original draft, Data curation, Visualization, Validation, Software, Methodology, Investigation. \textbf{Van-Manh Tran}: Writing - review \& editing, Writing - original draft, Data curation, Validation, Software, Investigation. \textbf{Nam-Thuan Trinh}: Writing - review \& editing, Writing - original draft, Data curation, Validation, Software, Investigation. \textbf{Thi-Thu-Hong Phan}: Conceptualization, Methodology, Supervision, Validation, Writing - original draft, Writing - review \& editing.

\section*{Declaration of interests:}
 
The authors declare that they have no known competing financial interests or personal relationships that could have appeared to influence the work reported in this paper.

\section*{Declaration of generative AI and AI-Assisted technologies in the writing process:}

During the preparation of this manuscript, Grammarly and generative large language models were employed to improve grammar and refine wording for clarity. The authors subsequently reviewed and revised the content as necessary and take full responsibility for the final version of the manuscript.

\bibliography{refs}

\begin{thebibliography}{39}
\expandafter\ifx\csname natexlab\endcsname\relax\def\natexlab#1{#1}\fi
\providecommand{\url}[1]{\texttt{#1}}
\providecommand{\href}[2]{#2}
\providecommand{\path}[1]{#1}
\providecommand{\DOIprefix}{doi:}
\providecommand{\ArXivprefix}{arXiv:}
\providecommand{\URLprefix}{URL: }
\providecommand{\Pubmedprefix}{pmid:}
\providecommand{\doi}[1]{\href{http://dx.doi.org/#1}{\path{#1}}}
\providecommand{\Pubmed}[1]{\href{pmid:#1}{\path{#1}}}
\providecommand{\bibinfo}[2]{#2}
\ifx\xfnm\relax \def\xfnm[#1]{\unskip,\space#1}\fi
\bibitem[{Alayrac et~al.(2022)Alayrac, Donahue, Luc, Miech, Barr, Hasson, Lenc, Mensch, Millican, Reynolds, Ring, Rutherford, Cabi, Han, Gong, Samangooei, Monteiro, Menick, Borgeaud, Brock, Nematzadeh, Sharifzadeh, Binkowski, Barreira, Vinyals, Zisserman and Simonyan}]{Alayrac2022}
\bibinfo{author}{Alayrac, J.B.}, \bibinfo{author}{Donahue, J.}, \bibinfo{author}{Luc, P.}, \bibinfo{author}{Miech, A.}, \bibinfo{author}{Barr, I.}, \bibinfo{author}{Hasson, Y.}, \bibinfo{author}{Lenc, K.}, \bibinfo{author}{Mensch, A.}, \bibinfo{author}{Millican, K.}, \bibinfo{author}{Reynolds, M.}, \bibinfo{author}{Ring, R.}, \bibinfo{author}{Rutherford, E.}, \bibinfo{author}{Cabi, S.}, \bibinfo{author}{Han, T.}, \bibinfo{author}{Gong, Z.}, \bibinfo{author}{Samangooei, S.}, \bibinfo{author}{Monteiro, M.}, \bibinfo{author}{Menick, J.}, \bibinfo{author}{Borgeaud, S.}, \bibinfo{author}{Brock, A.}, \bibinfo{author}{Nematzadeh, A.}, \bibinfo{author}{Sharifzadeh, S.}, \bibinfo{author}{Binkowski, M.}, \bibinfo{author}{Barreira, R.}, \bibinfo{author}{Vinyals, O.}, \bibinfo{author}{Zisserman, A.}, \bibinfo{author}{Simonyan, K.}, \bibinfo{year}{2022}.
\newblock \bibinfo{title}{Flamingo: a visual language model for few-shot learning}.
\newblock \href{http://arxiv.org/abs/2204.14198}{{\tt arXiv:2204.14198}}.
\bibitem[{Carreira and Zisserman(2018)}]{Carreira2018}
\bibinfo{author}{Carreira, J.}, \bibinfo{author}{Zisserman, A.}, \bibinfo{year}{2018}.
\newblock \bibinfo{title}{Quo vadis, action recognition? a new model and the kinetics dataset}.
\newblock \href{http://arxiv.org/abs/1705.07750}{{\tt arXiv:1705.07750}}.
\bibitem[{Cui et~al.(2025)Cui, Liu, Liu, Du, Chen, Lian, Li and Wang}]{Cui2025}
\bibinfo{author}{Cui, M.}, \bibinfo{author}{Liu, X.}, \bibinfo{author}{Liu, H.}, \bibinfo{author}{Du, Z.}, \bibinfo{author}{Chen, T.}, \bibinfo{author}{Lian, G.}, \bibinfo{author}{Li, D.}, \bibinfo{author}{Wang, W.}, \bibinfo{year}{2025}.
\newblock \bibinfo{title}{Multimodal fish feeding intensity assessment in aquaculture}.
\newblock \bibinfo{journal}{IEEE Transactions on Automation Science and Engineering} \bibinfo{volume}{22}, \bibinfo{pages}{9485--9497}.
\newblock \DOIprefix\doi{10.1109/TASE.2024.3507098}.
\bibitem[{Cui et~al.(2022)Cui, Liu, Zhao, Sun, Lian, Chen, Plumbley, Li and Wang}]{Cui2022}
\bibinfo{author}{Cui, M.}, \bibinfo{author}{Liu, X.}, \bibinfo{author}{Zhao, J.}, \bibinfo{author}{Sun, J.}, \bibinfo{author}{Lian, G.}, \bibinfo{author}{Chen, T.}, \bibinfo{author}{Plumbley, M.D.}, \bibinfo{author}{Li, D.}, \bibinfo{author}{Wang, W.}, \bibinfo{year}{2022}.
\newblock \bibinfo{title}{Fish feeding intensity assessment in aquaculture: A new audio dataset affia3k and a deep learning algorithm}, in: \bibinfo{booktitle}{2022 IEEE 32nd International Workshop on Machine Learning for Signal Processing (MLSP)}, pp. \bibinfo{pages}{1--6}.
\newblock \DOIprefix\doi{10.1109/MLSP55214.2022.9943405}.
\bibitem[{Dong et~al.(2024)Dong, Zhao, Wang, Cai, Pang and Liu}]{Dong2024}
\bibinfo{author}{Dong, Y.}, \bibinfo{author}{Zhao, S.}, \bibinfo{author}{Wang, Y.}, \bibinfo{author}{Cai, K.}, \bibinfo{author}{Pang, H.}, \bibinfo{author}{Liu, Y.}, \bibinfo{year}{2024}.
\newblock \bibinfo{title}{An integrated three-stream network model for discriminating fish feeding intensity using multi-feature analysis and deep learning}.
\newblock \bibinfo{journal}{PLOS ONE} \bibinfo{volume}{19}, \bibinfo{pages}{1--24}.
\newblock \DOIprefix\doi{https://doi.org/10.1371/journal.pone.0310356}.
\bibitem[{Dosovitskiy et~al.(2021)Dosovitskiy, Beyer, Kolesnikov, Weissenborn, Zhai, Unterthiner, Dehghani, Minderer, Heigold, Gelly, Uszkoreit and Houlsby}]{Dosovitskiy2021}
\bibinfo{author}{Dosovitskiy, A.}, \bibinfo{author}{Beyer, L.}, \bibinfo{author}{Kolesnikov, A.}, \bibinfo{author}{Weissenborn, D.}, \bibinfo{author}{Zhai, X.}, \bibinfo{author}{Unterthiner, T.}, \bibinfo{author}{Dehghani, M.}, \bibinfo{author}{Minderer, M.}, \bibinfo{author}{Heigold, G.}, \bibinfo{author}{Gelly, S.}, \bibinfo{author}{Uszkoreit, J.}, \bibinfo{author}{Houlsby, N.}, \bibinfo{year}{2021}.
\newblock \bibinfo{title}{An image is worth 16x16 words: Transformers for image recognition at scale}.
\newblock \href{http://arxiv.org/abs/2010.11929}{{\tt arXiv:2010.11929}}.
\bibitem[{Du et~al.(2023)Du, Cui, Wang, Liu, Xu, Bai, Sun, Wang, Wang and Li}]{Du2023}
\bibinfo{author}{Du, Z.}, \bibinfo{author}{Cui, M.}, \bibinfo{author}{Wang, Q.}, \bibinfo{author}{Liu, X.}, \bibinfo{author}{Xu, X.}, \bibinfo{author}{Bai, Z.}, \bibinfo{author}{Sun, C.}, \bibinfo{author}{Wang, B.}, \bibinfo{author}{Wang, S.}, \bibinfo{author}{Li, D.}, \bibinfo{year}{2023}.
\newblock \bibinfo{title}{Feeding intensity assessment of aquaculture fish using mel spectrogram and deep learning algorithms}.
\newblock \bibinfo{journal}{Aquacultural Engineering} \bibinfo{volume}{102}, \bibinfo{pages}{102345}.
\newblock \DOIprefix\doi{https://doi.org/10.1016/j.aquaeng.2023.102345}.
\bibitem[{Du et~al.(2026)Du, Tong, Hao, Zhou, Zhang, Tian and Zhou}]{Du2026}
\bibinfo{author}{Du, Z.}, \bibinfo{author}{Tong, X.}, \bibinfo{author}{Hao, Y.}, \bibinfo{author}{Zhou, X.}, \bibinfo{author}{Zhang, L.}, \bibinfo{author}{Tian, J.}, \bibinfo{author}{Zhou, C.}, \bibinfo{year}{2026}.
\newblock \bibinfo{title}{An uncertainty-aware heterogeneous multi-level knowledge distillation framework for lightweight audio-visual recognition of fish feeding intensity}.
\newblock \bibinfo{journal}{Information Processing in Agriculture} \DOIprefix\doi{https://doi.org/10.1016/j.inpa.2026.06.006}.
\bibitem[{Feichtenhofer(2020)}]{Feichtenhofer2020}
\bibinfo{author}{Feichtenhofer, C.}, \bibinfo{year}{2020}.
\newblock \bibinfo{title}{X3d: Expanding architectures for efficient video recognition}.
\newblock \href{http://arxiv.org/abs/2004.04730}{{\tt arXiv:2004.04730}}.
\bibitem[{Feng et~al.(2022)Feng, Yang, Liu, Zhao, Liu, Yan and Zhou}]{Feng2022}
\bibinfo{author}{Feng, S.}, \bibinfo{author}{Yang, X.}, \bibinfo{author}{Liu, Y.}, \bibinfo{author}{Zhao, Z.}, \bibinfo{author}{Liu, J.}, \bibinfo{author}{Yan, Y.}, \bibinfo{author}{Zhou, C.}, \bibinfo{year}{2022}.
\newblock \bibinfo{title}{Fish feeding intensity quantification using machine vision and a lightweight 3d resnet-glore network}.
\newblock \bibinfo{journal}{Aquacultural Engineering} \bibinfo{volume}{98}, \bibinfo{pages}{102244}.
\newblock \DOIprefix\doi{https://doi.org/10.1016/j.aquaeng.2022.102244}.
\bibitem[{Gu et~al.(2025)Gu, Zhao, Duan, Meng, Li and Zhao}]{Gu2025}
\bibinfo{author}{Gu, X.}, \bibinfo{author}{Zhao, S.}, \bibinfo{author}{Duan, Y.}, \bibinfo{author}{Meng, Y.}, \bibinfo{author}{Li, D.}, \bibinfo{author}{Zhao, R.}, \bibinfo{year}{2025}.
\newblock \bibinfo{title}{Mmfinet: A multimodal fusion network for accurate fish feeding intensity assessment in recirculating aquaculture systems}.
\newblock \bibinfo{journal}{Computers and Electronics in Agriculture} \bibinfo{volume}{232}, \bibinfo{pages}{110138}.
\newblock \DOIprefix\doi{https://doi.org/10.1016/j.compag.2025.110138}.
\bibitem[{He et~al.(2015)He, Zhang, Ren and Sun}]{He2015}
\bibinfo{author}{He, K.}, \bibinfo{author}{Zhang, X.}, \bibinfo{author}{Ren, S.}, \bibinfo{author}{Sun, J.}, \bibinfo{year}{2015}.
\newblock \bibinfo{title}{Deep residual learning for image recognition}.
\newblock \href{http://arxiv.org/abs/1512.03385}{{\tt arXiv:1512.03385}}.
\bibitem[{Howard et~al.(2017)Howard, Zhu, Chen, Kalenichenko, Wang, Weyand, Andreetto and Adam}]{Howard2017}
\bibinfo{author}{Howard, A.G.}, \bibinfo{author}{Zhu, M.}, \bibinfo{author}{Chen, B.}, \bibinfo{author}{Kalenichenko, D.}, \bibinfo{author}{Wang, W.}, \bibinfo{author}{Weyand, T.}, \bibinfo{author}{Andreetto, M.}, \bibinfo{author}{Adam, H.}, \bibinfo{year}{2017}.
\newblock \bibinfo{title}{Mobilenets: Efficient convolutional neural networks for mobile vision applications}.
\newblock \href{http://arxiv.org/abs/1704.04861}{{\tt arXiv:1704.04861}}.
\bibitem[{Huang et~al.(2018)Huang, Liu, van~der Maaten and Weinberger}]{Huang2018}
\bibinfo{author}{Huang, G.}, \bibinfo{author}{Liu, Z.}, \bibinfo{author}{van~der Maaten, L.}, \bibinfo{author}{Weinberger, K.Q.}, \bibinfo{year}{2018}.
\newblock \bibinfo{title}{Densely connected convolutional networks}.
\newblock \href{http://arxiv.org/abs/1608.06993}{{\tt arXiv:1608.06993}}.
\bibitem[{Huang et~al.(2025)Huang, Ma, Jin, Fan, Xie and Cai}]{Huang2025}
\bibinfo{author}{Huang, X.}, \bibinfo{author}{Ma, X.}, \bibinfo{author}{Jin, J.}, \bibinfo{author}{Fan, S.}, \bibinfo{author}{Xie, Y.}, \bibinfo{author}{Cai, W.}, \bibinfo{year}{2025}.
\newblock \bibinfo{title}{Assessment of feeding intensity of oreochromis niloticus (tilapia) based on improved vgg16 and voice spectrograms}.
\newblock \bibinfo{journal}{Aquaculture International} \bibinfo{volume}{33}, \bibinfo{pages}{246}.
\newblock \DOIprefix\doi{https://doi.org/10.1007/s10499-025-01913-0}.
\bibitem[{Kong et~al.(2020)Kong, Cao, Iqbal, Wang, Wang and Plumbley}]{Kong2020}
\bibinfo{author}{Kong, Q.}, \bibinfo{author}{Cao, Y.}, \bibinfo{author}{Iqbal, T.}, \bibinfo{author}{Wang, Y.}, \bibinfo{author}{Wang, W.}, \bibinfo{author}{Plumbley, M.D.}, \bibinfo{year}{2020}.
\newblock \bibinfo{title}{Panns: Large-scale pretrained audio neural networks for audio pattern recognition}.
\newblock \href{http://arxiv.org/abs/1912.10211}{{\tt arXiv:1912.10211}}.
\bibitem[{Li et~al.(2026)Li, Pan, Kong, Liu and Duan}]{Li2026}
\bibinfo{author}{Li, C.}, \bibinfo{author}{Pan, H.}, \bibinfo{author}{Kong, M.}, \bibinfo{author}{Liu, D.}, \bibinfo{author}{Duan, Q.}, \bibinfo{year}{2026}.
\newblock \bibinfo{title}{Fish feeding intensity recognition in aquaculture based on mfcc and an improved cnn}.
\newblock \bibinfo{journal}{Aquaculture International} \bibinfo{volume}{34}, \bibinfo{pages}{40}.
\newblock \DOIprefix\doi{https://doi.org/10.1007/s10499-026-02438-w}.
\bibitem[{Li et~al.(2020)Li, Wang, Wu, Miao, Du and Duan}]{Li2020}
\bibinfo{author}{Li, D.}, \bibinfo{author}{Wang, Z.}, \bibinfo{author}{Wu, S.}, \bibinfo{author}{Miao, Z.}, \bibinfo{author}{Du, L.}, \bibinfo{author}{Duan, Y.}, \bibinfo{year}{2020}.
\newblock \bibinfo{title}{Automatic recognition methods of fish feeding behavior in aquaculture: A review}.
\newblock \bibinfo{journal}{Aquaculture} \bibinfo{volume}{528}, \bibinfo{pages}{735508}.
\newblock \DOIprefix\doi{https://doi.org/10.1016/j.aquaculture.2020.735508}.
\bibitem[{Li et~al.(2025)Li, Wei, Ma and Wang}]{Li2025}
\bibinfo{author}{Li, J.}, \bibinfo{author}{Wei, Y.}, \bibinfo{author}{Ma, W.}, \bibinfo{author}{Wang, T.}, \bibinfo{year}{2025}.
\newblock \bibinfo{title}{Cross-modal complementarity learning for fish feeding intensity recognition via audio--visual fusion}.
\newblock \bibinfo{journal}{Animals} \bibinfo{volume}{15}, \bibinfo{pages}{2245}.
\newblock \DOIprefix\doi{https://doi.org/10.3390/ani15152245}.
\bibitem[{Liu et~al.(2024)Liu, Becerra, Bienvenido-Barcena, Yang, Zhao and Zhou}]{Liu2024}
\bibinfo{author}{Liu, J.}, \bibinfo{author}{Becerra, A.T.}, \bibinfo{author}{Bienvenido-Barcena, J.F.}, \bibinfo{author}{Yang, X.}, \bibinfo{author}{Zhao, Z.}, \bibinfo{author}{Zhou, C.}, \bibinfo{year}{2024}.
\newblock \bibinfo{title}{Cffi-vit: Enhanced vision transformer for the accurate classification of fish feeding intensity in aquaculture}.
\newblock \bibinfo{journal}{Journal of Marine Science and Engineering} \bibinfo{volume}{12}, \bibinfo{pages}{1132}.
\newblock \DOIprefix\doi{https://doi.org/10.3390/jmse12071132}.
\bibitem[{Liu et~al.(2021)Liu, Lin, Cao, Hu, Wei, Zhang, Lin and Guo}]{Liu2021}
\bibinfo{author}{Liu, Z.}, \bibinfo{author}{Lin, Y.}, \bibinfo{author}{Cao, Y.}, \bibinfo{author}{Hu, H.}, \bibinfo{author}{Wei, Y.}, \bibinfo{author}{Zhang, Z.}, \bibinfo{author}{Lin, S.}, \bibinfo{author}{Guo, B.}, \bibinfo{year}{2021}.
\newblock \bibinfo{title}{Swin transformer: Hierarchical vision transformer using shifted windows}.
\newblock \href{http://arxiv.org/abs/2103.14030}{{\tt arXiv:2103.14030}}.
\bibitem[{Liu et~al.(2022)Liu, Mao, Wu, Feichtenhofer, Darrell and Xie}]{Liu2022}
\bibinfo{author}{Liu, Z.}, \bibinfo{author}{Mao, H.}, \bibinfo{author}{Wu, C.Y.}, \bibinfo{author}{Feichtenhofer, C.}, \bibinfo{author}{Darrell, T.}, \bibinfo{author}{Xie, S.}, \bibinfo{year}{2022}.
\newblock \bibinfo{title}{A convnet for the 2020s}.
\newblock \href{http://arxiv.org/abs/2201.03545}{{\tt arXiv:2201.03545}}.
\bibitem[{Sandler et~al.(2019)Sandler, Howard, Zhu, Zhmoginov and Chen}]{Sandler2019}
\bibinfo{author}{Sandler, M.}, \bibinfo{author}{Howard, A.}, \bibinfo{author}{Zhu, M.}, \bibinfo{author}{Zhmoginov, A.}, \bibinfo{author}{Chen, L.C.}, \bibinfo{year}{2019}.
\newblock \bibinfo{title}{Mobilenetv2: Inverted residuals and linear bottlenecks}.
\newblock \href{http://arxiv.org/abs/1801.04381}{{\tt arXiv:1801.04381}}.
\bibitem[{Sha et~al.(2026)Sha, Wu, Zheng, Wu, Cheng, Chen, Zhou and Yu}]{Sha2026}
\bibinfo{author}{Sha, Y.}, \bibinfo{author}{Wu, J.}, \bibinfo{author}{Zheng, S.}, \bibinfo{author}{Wu, P.}, \bibinfo{author}{Cheng, S.}, \bibinfo{author}{Chen, M.}, \bibinfo{author}{Zhou, Y.}, \bibinfo{author}{Yu, H.}, \bibinfo{year}{2026}.
\newblock \bibinfo{title}{Dynamic audio--visual fusion for accurate fish feeding intensity recognition in recirculating aquaculture environments}.
\newblock \bibinfo{journal}{Aquaculture International} \bibinfo{volume}{34}, \bibinfo{pages}{77}.
\newblock \DOIprefix\doi{https://doi.org/10.1007/s10499-026-02487-1}.
\bibitem[{Tan and Le(2020)}]{Tan2020}
\bibinfo{author}{Tan, M.}, \bibinfo{author}{Le, Q.V.}, \bibinfo{year}{2020}.
\newblock \bibinfo{title}{Efficientnet: Rethinking model scaling for convolutional neural networks}.
\newblock \href{http://arxiv.org/abs/1905.11946}{{\tt arXiv:1905.11946}}.
\bibitem[{de~Verdal et~al.(2018)de~Verdal, Komen, Quillet, Chatain, Allal, Benzie and Vandeputte}]{Verdal2018}
\bibinfo{author}{de~Verdal, H.}, \bibinfo{author}{Komen, H.}, \bibinfo{author}{Quillet, E.}, \bibinfo{author}{Chatain, B.}, \bibinfo{author}{Allal, F.}, \bibinfo{author}{Benzie, J.A.}, \bibinfo{author}{Vandeputte, M.}, \bibinfo{year}{2018}.
\newblock \bibinfo{title}{Improving feed efficiency in fish using selective breeding: a review}.
\newblock \bibinfo{journal}{Reviews in Aquaculture} \bibinfo{volume}{10}, \bibinfo{pages}{833--851}.
\newblock \DOIprefix\doi{https://doi.org/10.1111/raq.12202}.
\bibitem[{Wang et~al.(2025)Wang, Wang, He, Han, Long, Wu and Yu}]{Wang2025}
\bibinfo{author}{Wang, H.}, \bibinfo{author}{Wang, M.}, \bibinfo{author}{He, Z.}, \bibinfo{author}{Han, A.}, \bibinfo{author}{Long, J.}, \bibinfo{author}{Wu, J.}, \bibinfo{author}{Yu, H.}, \bibinfo{year}{2025}.
\newblock \bibinfo{title}{Collaborative learning of multimodal fusion and audio-guided attention for fish feeding behavior analysis in aquafarm}.
\newblock \bibinfo{journal}{Aquaculture International} \bibinfo{volume}{33}, \bibinfo{pages}{463}.
\newblock \DOIprefix\doi{https://doi.org/10.1007/s10499-025-02150-1}.
\bibitem[{Wang et~al.(2026a)Wang, Xu, Lu, Chen, Zhang and Zhai}]{Wang2026b}
\bibinfo{author}{Wang, H.}, \bibinfo{author}{Xu, H.}, \bibinfo{author}{Lu, C.}, \bibinfo{author}{Chen, X.}, \bibinfo{author}{Zhang, Z.}, \bibinfo{author}{Zhai, Z.}, \bibinfo{year}{2026}a.
\newblock \bibinfo{title}{Feedingmonitor: An improved tnt-based method for evaluating the fish feeding intensity in the recirculating aquaculture system}.
\newblock \bibinfo{journal}{Computers and Electronics in Agriculture} \bibinfo{volume}{240}, \bibinfo{pages}{111233}.
\newblock \DOIprefix\doi{https://doi.org/10.1016/j.compag.2025.111233}.
\bibitem[{Wang et~al.(2026b)Wang, Wen, Yang, Mei, Yang and Li}]{Wang2026a}
\bibinfo{author}{Wang, Z.}, \bibinfo{author}{Wen, S.}, \bibinfo{author}{Yang, L.}, \bibinfo{author}{Mei, Y.}, \bibinfo{author}{Yang, Q.}, \bibinfo{author}{Li, Y.}, \bibinfo{year}{2026}b.
\newblock \bibinfo{title}{A dual-stream spatiotemporal fusion method for fish school feeding intensity identification}.
\newblock \bibinfo{journal}{Aquaculture International} \bibinfo{volume}{34}, \bibinfo{pages}{33}.
\newblock \DOIprefix\doi{https://doi.org/10.1007/s10499-025-02425-7}.
\bibitem[{Wu et~al.(2024)Wu, Wang, Shi, Wang, Qian and Jiang}]{Wu2024}
\bibinfo{author}{Wu, Y.}, \bibinfo{author}{Wang, X.}, \bibinfo{author}{Shi, Y.}, \bibinfo{author}{Wang, Y.}, \bibinfo{author}{Qian, D.}, \bibinfo{author}{Jiang, Y.}, \bibinfo{year}{2024}.
\newblock \bibinfo{title}{Fish feeding intensity assessment method using deep learning-based analysis of feeding splashes}.
\newblock \bibinfo{journal}{Computers and Electronics in Agriculture} \bibinfo{volume}{221}, \bibinfo{pages}{108995}.
\newblock \DOIprefix\doi{https://doi.org/10.1016/j.compag.2024.108995}.
\bibitem[{Xiao et~al.(2025)Xiao, Huang, Zhang, Bi, You, He and Guan}]{Xiao2025}
\bibinfo{author}{Xiao, Y.}, \bibinfo{author}{Huang, L.}, \bibinfo{author}{Zhang, S.}, \bibinfo{author}{Bi, C.}, \bibinfo{author}{You, X.}, \bibinfo{author}{He, S.}, \bibinfo{author}{Guan, J.}, \bibinfo{year}{2025}.
\newblock \bibinfo{title}{Feeding behavior quantification and recognition for intelligent fish farming application: A review}.
\newblock \bibinfo{journal}{Applied Animal Behaviour Science} \bibinfo{volume}{285}, \bibinfo{pages}{106588}.
\newblock \DOIprefix\doi{https://doi.org/10.1016/j.applanim.2025.106588}.
\bibitem[{Xie et~al.(2018)Xie, Sun, Huang, Tu and Murphy}]{Xie2018}
\bibinfo{author}{Xie, S.}, \bibinfo{author}{Sun, C.}, \bibinfo{author}{Huang, J.}, \bibinfo{author}{Tu, Z.}, \bibinfo{author}{Murphy, K.}, \bibinfo{year}{2018}.
\newblock \bibinfo{title}{Rethinking spatiotemporal feature learning: Speed-accuracy trade-offs in video classification}.
\newblock \href{http://arxiv.org/abs/1712.04851}{{\tt arXiv:1712.04851}}.
\bibitem[{Yang et~al.(2024)Yang, Yu, Zhang, Zhang, Tu and Gu}]{Yang2024}
\bibinfo{author}{Yang, Y.}, \bibinfo{author}{Yu, H.}, \bibinfo{author}{Zhang, X.}, \bibinfo{author}{Zhang, P.}, \bibinfo{author}{Tu, W.}, \bibinfo{author}{Gu, L.}, \bibinfo{year}{2024}.
\newblock \bibinfo{title}{Fish behavior recognition based on an audio-visual multimodal interactive fusion network}.
\newblock \bibinfo{journal}{Aquacultural Engineering} \bibinfo{volume}{107}, \bibinfo{pages}{102471}.
\newblock \DOIprefix\doi{https://doi.org/10.1016/j.aquaeng.2024.102471}.
\bibitem[{Zhang et~al.(2022)Zhang, Wang, Li, Liu, Zhang and Duan}]{Zhang2022}
\bibinfo{author}{Zhang, L.}, \bibinfo{author}{Wang, J.}, \bibinfo{author}{Li, B.}, \bibinfo{author}{Liu, Y.}, \bibinfo{author}{Zhang, H.}, \bibinfo{author}{Duan, Q.}, \bibinfo{year}{2022}.
\newblock \bibinfo{title}{A mobilenetv2-senet-based method for identifying fish school feeding behavior}.
\newblock \bibinfo{journal}{Aquacultural Engineering} \bibinfo{volume}{99}, \bibinfo{pages}{102288}.
\newblock \DOIprefix\doi{https://doi.org/10.1016/j.aquaeng.2022.102288}.
\bibitem[{Zhang et~al.(2025a)Zhang, Wang and Li}]{Zhang2025c}
\bibinfo{author}{Zhang, L.}, \bibinfo{author}{Wang, L.}, \bibinfo{author}{Li, S.}, \bibinfo{year}{2025}a.
\newblock \bibinfo{title}{A fish school feeding intensity assessment method based on efficientnetb0-fpsanet}.
\newblock \bibinfo{journal}{Aquaculture International} \bibinfo{volume}{33}, \bibinfo{pages}{688}.
\newblock \DOIprefix\doi{https://doi.org/10.1007/s10499-025-02337-6}.
\bibitem[{Zhang et~al.(2026)Zhang, Yao, Zhao, Li, Chen and Wang}]{Zhang2026a}
\bibinfo{author}{Zhang, S.}, \bibinfo{author}{Yao, M.}, \bibinfo{author}{Zhao, J.}, \bibinfo{author}{Li, D.}, \bibinfo{author}{Chen, Y.}, \bibinfo{author}{Wang, H.}, \bibinfo{year}{2026}.
\newblock \bibinfo{title}{Progressive multimodal interaction network for reliable quantification of fish feeding intensity in aquaculture}.
\newblock \bibinfo{journal}{Aquaculture International} \bibinfo{volume}{34}, \bibinfo{pages}{224}.
\newblock \DOIprefix\doi{https://doi.org/10.1007/s10499-026-02624-w}.
\bibitem[{Zhang et~al.(2025b)Zhang, Zou, Hu and Li}]{Zhang2025a}
\bibinfo{author}{Zhang, Z.}, \bibinfo{author}{Zou, B.}, \bibinfo{author}{Hu, Q.}, \bibinfo{author}{Li, W.}, \bibinfo{year}{2025}b.
\newblock \bibinfo{title}{Multimodal knowledge distillation framework for fish feeding behaviour recognition in industrial aquaculture}.
\newblock \bibinfo{journal}{Biosystems Engineering} \bibinfo{volume}{255}, \bibinfo{pages}{104170}.
\newblock \DOIprefix\doi{https://doi.org/10.1016/j.biosystemseng.2025.104170}.
\bibitem[{Zhou et~al.(2019)Zhou, Xu, Chen, Zhang, Sun, Yang and Wang}]{Zhou2019}
\bibinfo{author}{Zhou, C.}, \bibinfo{author}{Xu, D.}, \bibinfo{author}{Chen, L.}, \bibinfo{author}{Zhang, S.}, \bibinfo{author}{Sun, C.}, \bibinfo{author}{Yang, X.}, \bibinfo{author}{Wang, Y.}, \bibinfo{year}{2019}.
\newblock \bibinfo{title}{Evaluation of fish feeding intensity in aquaculture using a convolutional neural network and machine vision}.
\newblock \bibinfo{journal}{Aquaculture} \bibinfo{volume}{507}, \bibinfo{pages}{457--465}.
\newblock \DOIprefix\doi{https://doi.org/10.1016/j.aquaculture.2019.04.056}.
\bibitem[{Zhou et~al.(2018)Zhou, Xu, Lin, Sun and Yang}]{Zhou2018}
\bibinfo{author}{Zhou, C.}, \bibinfo{author}{Xu, D.}, \bibinfo{author}{Lin, K.}, \bibinfo{author}{Sun, C.}, \bibinfo{author}{Yang, X.}, \bibinfo{year}{2018}.
\newblock \bibinfo{title}{Intelligent feeding control methods in aquaculture with an emphasis on fish: a review}.
\newblock \bibinfo{journal}{Reviews in Aquaculture} \bibinfo{volume}{10}, \bibinfo{pages}{975--993}.
\newblock \DOIprefix\doi{https://doi.org/10.1111/raq.12218}.

\end{thebibliography}
\bibliographystyle{elsarticle-harv} 

\end{document}